\documentclass[lettersize,journal]{IEEEtran}

\usepackage{multicol}
\usepackage{soul}
\usepackage{epsfig}
\usepackage{cite}
\usepackage{booktabs}
\usepackage{graphicx}
\usepackage{url}
\usepackage{stfloats}
\usepackage{amsmath,accents}
\usepackage{mathrsfs}
\usepackage{hyperref}
\usepackage{amsfonts}   
\usepackage{amsopn}     
\usepackage{psfrag}
\usepackage[tight]{subfigure}
\usepackage{mdwtab}     
\usepackage{mdwlist}
\usepackage[normalem]{ulem}
\usepackage{acronym}
\usepackage{url}
\usepackage{accents}
\usepackage[utf8]{inputenc}
\usepackage{enumerate}
\usepackage{color}
\usepackage{amsthm}
\usepackage{epsfig}
\usepackage{cite}
\usepackage{graphicx}
\graphicspath{{figureEPS/}}
\usepackage{subfigure}
\usepackage{url}
\usepackage{stfloats}
\usepackage{amsopn}    
\usepackage{psfrag}
\usepackage{xcolor}
\usepackage{subfigure}
\usepackage{mdwtab} 
\usepackage{epstopdf}
\usepackage{marginnote}
\usepackage{algorithm}
\usepackage{algorithmicx} 
\usepackage{algpseudocode} 

\newtheorem{theorem}{\bf Theorem}

\newtheorem{definition}{\bf Definition}
\newtheorem{prop}{\bf Proposition}

\acrodef{DOF}[\textsc{DOF}]{degrees of freedom}
\acrodef{COM}[\textsc{COM}]{center of mass}
\acrodef{DMP}[\textsc{DMP}]{\emph{Dynamic Movement Primitive}}
\acrodef{SSME}[\textsc{SSME}]{\emph{Switching System with Multiple Equilibria}}
\acrodef{RRT}[\textsc{RRT}]{Rapidly Exploring Random Trees}
\acrodef{MPC}[\textsc{MPC}]{Model Predictive Control}
\acrodef{HZD}[\textsc{HZD}]{Hybrid Zero Dynamics}
\acrodef{ZMP}[\textsc{ZMP}]{Zero Moment Point}
\acrodef{SOS}[\textsc{SoS}]{sums-of-squares}
\acrodef{BOA}[\textsc{BoA}]{basin of attraction}
\acrodef{LIP}[\textsc{LIP}]{Linear Inverted Pendulum}
\acrodef{SLIP}[\textsc{SLIP}]{Spring Loaded Inverted Pendulum}
\acrodef{SWC}[\textsc{SWC}]{\emph{Safe Walking Corridor}}

\ifCLASSINFOpdf

\else

\fi

\begin{document}
\pagestyle{plain}   

\title{Reactive Gait Composition with Stability: Dynamic Walking amidst Static and Moving Obstacles}

\author{Kunal~Sanjay~Narkhede, Mohamad~Shafiee~Motahar, Sushant~Veer, and Ioannis~Poulakakis
\thanks{K. S. Narkhede and I. Poulakakis are with the Department of Mechanical Engineering, University of Delaware, Newark, DE 19716, USA;  e-mail: {\tt\small \{kunalnk, poulakas\}@udel.edu.} M. S. Motahar is with ThermoFisher Scientific, Hillsboro, OR 97124, USA; e-mail: {\tt\small motahar@udel.edu}. S. Veer is with NVIDIA Research, Santa Clara, CA 95051, USA; e-mail: {\tt\small sveer@nvidia.com}. Both M. S. Motahar and S. Veer were with the Department of Mechanical Engineering, University of Delaware, Newark, DE 19716, USA, when this work was performed.} 
\thanks{This work is supported by NSF grants NRI-1327614 and IIS-1350721.}
}

\maketitle

\thispagestyle{plain}  
\pagestyle{plain}

\begin{abstract} 
This paper presents a modular approach to motion planning with provable stability guarantees for robots that move through changing environments via periodic locomotion behaviors. We focus on dynamic walkers as a paradigm for such systems, although the tools developed in this paper can be used to support general compositional approaches to robot motion planning with Dynamic Movement Primitives (DMPs). Our approach ensures \emph{a priori} that the suggested plan can be stably executed. This is achieved by formulating the planning process as a Switching System with Multiple Equilibria (SSME) and proving that the system's evolution remains within explicitly characterized trapping regions in the state space under suitable constraints on the frequency of switching among the DMPs. These conditions effectively encapsulate the low-level stability limitations in a form that can be easily communicated to the planner to guarantee that the suggested plan is compatible with the robot's dynamics. Furthermore, we show how the available primitives can be safely composed online in a receding horizon manner to enable the robot to react to moving obstacles. The proposed framework is applied on 3D bipedal walking models under common modeling assumptions, and offers a modular approach towards stably integrating readily available low-level locomotion control and high-level planning methods.
\end{abstract}
 
\section{Introduction}

The ability to navigate in changing environments is central to developing autonomous robots. Many approaches aim at enabling robots to ``translate'' a navigation task to a sequence of \emph{motion} (or, \emph{movement}) \emph{primitives}, the execution of which transfers a robot to a desired location, avoiding obstacles on the way. These approaches lend themselves naturally to a hierarchical formulation, wherein, at the high level, the requisite primitive sequence is constructed, and then realized at the low level via suitable feedback control laws. One important advantage of this hierarchy is that the low-level control and high-level planning modules can be treated separately. This benefit, however, comes with the challenge of ensuring \emph{safe} integration of the control and planning modules; here, the term ``safe'' incorporates both collision-free navigation and stable platform operation.

\subsection{Central themes and relation to prior work}

\begin{figure*}[t]
\centering
\subfigure[]{\includegraphics[]{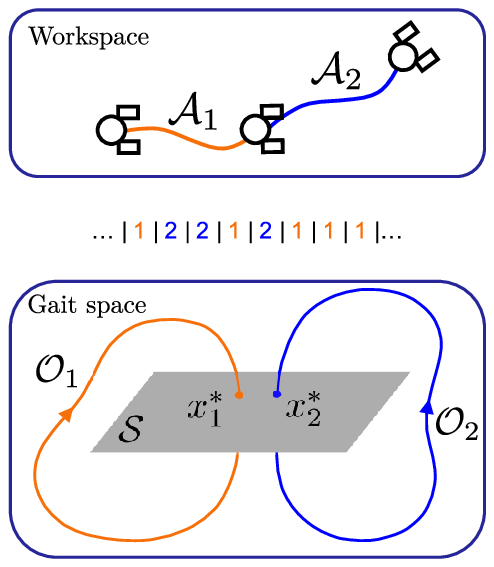} \label{fig:HILO_limit_cycles} 
}\quad
\subfigure[]{\includegraphics[]{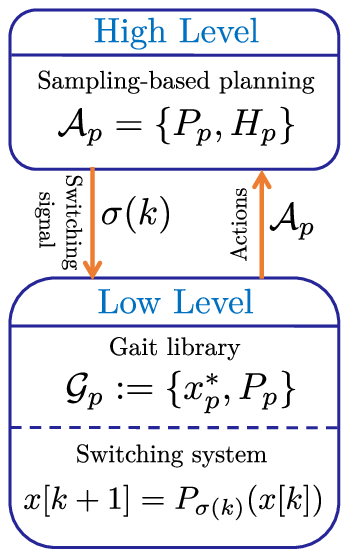} \label{fig:HILO_structure}
}\quad
\subfigure[]{\includegraphics[]{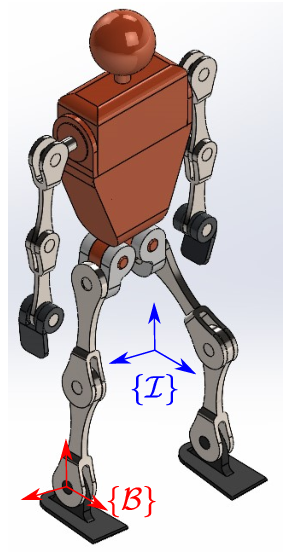} \label{fig:3Dbiped_actions}
}\quad
\subfigure[]{\includegraphics[]{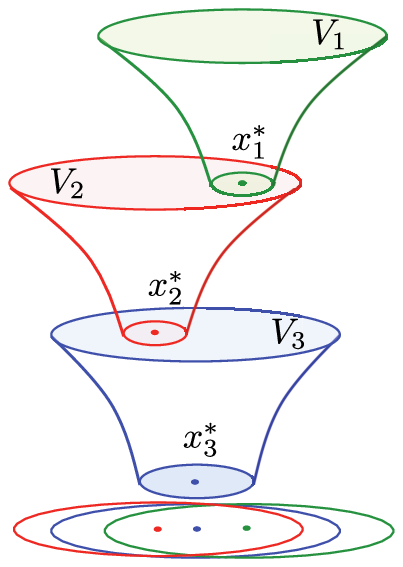} \label{fig:funnels}
}\quad
\caption{A modular approach to planning with limit-cycle gaits. \textbf{(a)} Switching between limit-cycle locomotion gaits. The limit cycles $\mathcal{O}_1$ and $\mathcal{O}_2$ are represented as fixed points $x^*_1$ and $x^*_2$ of Poincar\'e maps $P_1$ and $P_2$ defined on $\mathcal{S}$. Execution of the limit cycles results in displacements $\mathcal{A}_1$ and $\mathcal{A}_2$ in the system's workspace. \textbf{(b)} The overall problem is decomposed into low-level gait generation and high-level planning subproblems. At the low level, a gait controller is employed to extract a library $\mathbb{G} = \{ \mathcal{G}_p ~|~ p \in \mathcal{P} \}$ of limit-cycle gait primitives $\mathcal{G}_p = \{x^*_p, P_p \}$; each gait primitive generates an action $\mathcal{A} = \{P_p, H_p\}$ thus forming collection $\mathbb{A} = \{ \mathcal{A}_p ~|~ p \in \mathcal{P} \}$ that represents displacements available for planning. At the high level, an algorithm suggests a plan that achieves the desired transfer. The execution of the suggested plan is captured by a discrete-time switching system with multiple equilibria as in \eqref{eq:switched_system}. \textbf{(c)} A generic model of a 3D bipedal robot with $\{\mathcal{I}\}$ and $\{\mathcal{B}\}$ are the inertia and body fixed frames respectively. \textbf{(d)} Practically stable primitive composition through switching among Lyapunov functions $V_p$ associated with each gait $\mathcal{G}_p$.}
\label{fig:overview}
\end{figure*}

This paper proposes and rigorously analyzes, a \emph{modular} framework (Fig.~\ref{fig:overview}), within which established planning and control design methods can be integrated in a \emph{provably} stable way. We focus on robots that move through their environment via dynamic rhythmic interactions, adopting dynamically stable walking bipeds~\cite{mcgeer1990passive, Collins2005Science, Hobbelen2007} as an example. However, the tools developed in this paper can, in principle, be used to safely plan motions for aerial~\cite{Ramezani2017Science} and underwater~\cite{Georgiades2009OE} vehicles that use oscillating appendages---e.g., flapping wings or flippers---for locomotion.

At the core of our approach is the notion of a \ac{DMP}~\cite{ijspeert2013dynamical}, which in the case of dynamic walkers takes the form of a stride-to-stride map with a locally asymptotically stable fixed point representing a desired periodic locomotion behavior (Fig.~\ref{fig:HILO_limit_cycles}). Thus, a \ac{DMP} captures the \emph{local dynamics} around a nominal locomotion behavior at the stride level. Execution of each \ac{DMP} results in a displacement in the robot's workspace; these displacements form \emph{actions} available for planning (Fig.~\ref{fig:HILO_limit_cycles}). To derive conditions under which the planner's suggestions are compatible with the platform's stability limitations, this paper formulates the planning process as a \ac{SSME} in discrete time (Fig.~\ref{fig:HILO_structure}), and proves that its solution can be confined in an \emph{explicitly characterized} safe region provided that a bound on the frequency of \ac{DMP} switching is imposed. This bound effectively \emph{compresses} the information relevant to stable switching of \ac{DMP}s to a single number, which can be easily communicated to the planner. Besides stability, these results enable the effective approximation of the DMPs in a form that reduces computational effort, allowing online locomotion planning in the presence of moving obstacles.

This paper extends our preliminary work~\cite{motahar2016composing, Veer2017IROS} in three directions. First, it characterizes the dynamic properties of 3D bipeds that enable the application of the proposed approach. Second, it applies the notion of practical stability~\cite[p. 121]{la1961stability} and recent theoretical developments in \ac{SSME}s~\cite{Veer2020TAC} to derive explicit conditions for stable reactive planning with dynamic walking gaits.
Third, it demonstrates how the \emph{a priori} known stability conditions can be exploited to formulate a \emph{primitive-based} sequential \ac{MPC} scheme for online planning that allows dynamically walking bipeds to react to moving obstacles. 

\subsection{Related work}
\label{sec:related_work}

This section surveys related work and highlights connections and contributions to these approaches. 

\subsubsection{Locomotion as a limit cycle}
Dynamically stable walking is characterized by alternating phases of fall-and-capture pendular motions~\cite{mcgeer1990passive, Collins2005Science}. Stability in this case is often evaluated by treating the gait as a whole~\cite{Hobbelen2007}, and can be captured by the stability properties of isolated periodic solutions---i.e., \emph{limit cycles}---in the system's state space~\cite{westervelt2007feedback}. 
A significant volume of research on dynamically stable bipeds revolves around low-level feedback control design methods for  walking or running motions; the book~\cite{Sharbafi2017book} contains several approaches. An example is the \ac{HZD} method~\cite{westervelt2007feedback} and its extensions~\cite{Poulakakis2009ICRA, ames2014rapidly, hamed2014event, Sadeghian2017ICRA, Hereid2020ICRA}; see~\cite{AmesPoulakakis2017BLL} for an overview. Typically, however, control design methods for dynamically stable walking do not consider high-level planning objectives, such as avoiding unsafe regions on the walking surface or in the robot's workspace. Notable exceptions include~\cite{manchester2014ICRA}, which uses an energy-based planner to output suitable sequences of virtual constraints that generate feasible walking on known uneven terrain, and~\cite{nguyen2015safety, Nguyen2020Algo} where one- or two-step nominally periodic gaits are safely ``deformed'' to comply with precise foot positioning specifications. Most relevant to our work are~\cite{Gregg2010ICRA, gregg-planning2012} which use a library of limit-cycle walking gaits to generate collision-free paths in spaces cluttered by static obstacles.  

In this paper, we interpret limit cycles as task-simplifying locomotion abstractions~\cite{Hubicki2015ICRA} that generate suitable actions available for high-level planning. Formally, these abstractions take the form of pairs of  stride-to-stride maps together with locally asymptotically stable fixed points as in~\cite{Gregg2010ICRA, gregg-planning2012}, and thus can be considered as discrete-time versions of \ac{DMP}s~\cite{ijspeert2013dynamical}.

\subsubsection{Switched systems and motion planning}
Planning motions via \ac{DMP}s can be formulated as a switched dynamical system. The earliest work that makes this connection explicit in the context of dynamically walking bipeds is~\cite{Gregg2010ICRA, gregg-planning2012}, which also discusses the existence of a bound on the frequency of switching that ensures stability of the resulting walking sequence. However, this bound is estimated on the basis of computationally intensive simulations. The relation between switching frequency and stability is prevalent in the theory of switched systems, and can be captured by the notion of \emph{dwell time}, which was introduced in~\cite{hespanha1999stability} for switching among dynamical systems that \emph{share} a common equilibrium; refer to~\cite{liberzon2003switching} for a thorough study. Contrary to that work, switched systems formulations of motion planning problems are characterized by switching among systems with \emph{distinct} equilibria. An early study of such systems can be found in~\cite{basar2010}, which established boundedness of solutions under a fixed dwell-time bound. This bound was adapted to switched discrete-time systems in our previous work~\cite{motahar2016composing}, where it was applied to bipedal robot navigation amidst static obstacles. Beyond obstacle avoidance, switching\footnote{Related to gait switching is gait interpolation, which has been used in~\cite{da2016first, Gong2019ACC} to track a speed profile, in~\cite{Nguyen2018IJRR} to walk on stochastic terrain, and in~\cite{Sreenath2021Arxiv} to plan motions in height-constraint environments.} has been used to realize robust walking on uneven terrain with varying slope~\cite{Saglam2014RSS} or height~\cite{Park2013TRO, manchester2014ICRA, Buss2016ICRA}, and to achieve gait adaptation under persistent external forcing~\cite{veer2017supervisory, Veer2019ICRA, Chand2022RAL} and model uncertainty~\cite{Chand2020ICRA}. Switching has also been used to create hopping motions with speed adaptation in monopods~\cite{Bhounsule2018ACC, Bhounsule2019ICRA}, and to study gait transitions in quadrupeds~\cite{QuCao2015, Cao2016ALR, Ames2021IROS}.

This paper takes advantage of recent theoretical tools for \ac{SSME}s~\cite{Veer2019ICRA, Veer2019ACC, Veer2020TAC} to contribute to switching approaches for motion planning an explicit set characterization for \emph{practically stable} \ac{DMP} composition in a model predictive fashion. This way, changes in the robot’s workspace during plan execution, such as moving obstacles, can be taken into account. These tools are relevant to other robots, which---like legged robots--move via periodic interactions; e.g., aerial robots with flapping wings~\cite{Dorothy2016SCL, Ramezani2017Science}, and underwater robots with paddles~\cite{Georgiades2009OE}.

\subsubsection{Compositional approaches to planning}
The \ac{DMP}s defined in this paper correspond to attractive ``landscapes'' in the state space, focused at asymptotically stable equilibrium behaviors. In our discrete-time setting~\cite{motahar2016composing}, certificates for asymptotically stable equilibria can be obtained by discrete-time Lyapunov functions~\cite[Chapter 13]{haddad2008nonlinear}, which can be idealized as ``funnels'' attracting the (discrete-time) flow of the system towards their center. This idealization was made explicit\footnote{As stated in~\cite{burridge1999sequential}, the notion of a funnel was first proposed as a metaphor for robust behaviors in~\cite{Mason1985ICRA} and sequential composition was introduced in~\cite{Lozano1984IJRR} to combine controllers via pre-image backchaining from the overall task goal in the context of fine-motion planning.} in~\cite{burridge1999sequential}, where funnels were sequentially composed to generate dexterous ``batting'' maneuvers in a robot arm. Taking advantage of breakthrough developments in the automatic verification of Lyapunov functions using \ac{SOS} programming~\cite{parrilo2000structured} and in sample-based randomized motion planning algorithms~\cite{lavalle-RRT}, LQR trees have been introduced in~\cite{tedrake2010} as a computationally tractable approach to feedback motion planning for nonlinear systems. However, LQR trees cannot address situations in which the environment is only partially known prior to runtime; online implementation of the algorithm to handle changing environments is computationally infeasible~\cite{majumdar2016funnel}. To compose plans under uncertainty at runtime,~\cite{majumdar2016funnel} introduced a funnel library and applied the method to a small airplane performing obstacle avoidance tasks.        

Infinite-horizon Lyapunov funnels, such as those employed in~\cite{burridge1999sequential} and in our prior work~\cite{motahar2016composing}, represent convenient ways to describe positively invariant sets; thus, composition of Lyapunov funnels is related to motion planning with positively invariant set trees~\cite{DiCairano2016CDC, DiCairano2017CCTA}. These algorithms construct a tree of local state feedback control laws, each stabilizing a randomly sampled equilibrium, and use the corresponding positively invariant sets to decide when to safely transition from one controller to another. As in LQR trees, invariant set trees are extended backwards from the goal until they contain the initial condition, and thus they are not directly suited to receding-horizon problems~\cite{Barbosa2020CDC}. 

To achieve reactivity to changes in the environment, recent methods integrate a dedicated high-level component in the planning and control hierarchy. In~\cite{Shen2021ICRA}, this component relies on behavior trees to achieve efficient re-planning in changing environments. Another example of a hierarchical planning and control architecture can be found in~\cite{Vasilopoulos2022ICRA}, wherein a deliberative layer suggests a sequence of parameterized actions for a given task, a reactive layer converts these actions to target velocity commands, and a gait layer realizes those commands on the robot. Finally,~\cite{Zhao2021arxiv} proposes a reactive high-level planner based on Linear Temporal Logic; the suggested plans are then translated to robot motions using a reduced-order model ensuring safety within the context of that model.

\subsubsection{Reduced-order bipedal locomotion models}
Given the high-dimensional, nonlinear dynamics of bipedal walkers, several planning methods rely on \emph{reduced-order models}~\cite{Wieber2016ModelingAC, Wensing2017BLL} to resolve complexity. Prominent among these models is the \ac{LIP}~\cite{Kajita2001IROS}. However, establishing a formal connection between a reduced-order model selected \emph{a priori} and the full-order dynamics of the system remains largely an open problem. An early result in that direction can be found in~\cite{poulakakis2009spring} where a \ac{SLIP} is embedded as the \ac{HZD} of an asymmetric hopper so that solutions of the (pre-selected) reduced-order model are also solutions of the closed-loop full dynamics. However, selecting the target reduced-order model in advance and establishing the geometric constructions required in~\cite{poulakakis2009spring} can be overly restrictive~\cite{Poulakakis2010ICRA}. A less strict form of correspondence was recently proposed in~\cite{Wensing2021RAL} where an approximate simulation relation between a \ac{LIP} and a 30 \ac{DOF} model of the Valkyrie humanoid was established. 

An alternative approach---and the one we adopt in this paper---is to refrain from selecting in advance the reduced-order model used for planning. Instead, we start with the full-order system and impose virtual constraints to reduce it to a lower dimensional subsystem, which is not specified \emph{a priori} and can be used to define reduced-order \ac{DMP}s. Combining this reduction approach with the explicit characterization of solution trapping regions is the key to efficient computation via approximate planning actions.

\subsubsection*{Notation} 
$\mathbb{R}$ and $\mathbb{Z}$ represent real and integer numbers, while $\mathbb{R}+$ and $\mathbb{Z}+$ denote their non-negative counterparts.
The index $k\in\mathbb{Z}_+$ represents discrete time. Norms are denoted by $\|\cdot\|$. If $\mathcal{S}\subseteq \mathbb{R}^n$, $\accentset{\circ}{\mathcal{S}}$ denotes the interior of $\mathcal{S}$. We use $\mathsf{SO}(n)$ for the special orthogonal group of dimension $n$ and $\mathbb{S}^1$ to denote the unit circle. A continuous function $\alpha : \mathbb{R}_+ \to \mathbb{R}_+$ is of class $\mathcal{K}_\infty$ if it is strictly increasing, $\alpha(0)=0$ and $\lim_{s\to\infty} \alpha(s)=\infty$~\cite[p. 162]{haddad2008nonlinear}.

\section{Overview: Planning with DMPs}
\label{sec:overview}

This section provides an overview of our approach---see also Fig.~\ref{fig:overview}---and serves as a guide for the rest of the paper; technical details are left to subsequent sections.

\subsection{Problem formulation: Planning with limit cycles} 
\label{subsec:augmented_stride}

We consider a class of locomotion systems evolving in spaces that can be decomposed into a global position and heading part $\mathsf{G}$ and a state part $\mathcal{X}$. It is assumed that $\mathcal{X}$ contains distinguished periodic orbits---i.e., \emph{limit cycles}---corresponding to gaits of interest. Suppose that $\mathcal{O}$ is a limit cycle corresponding to one such gait (see Fig.~\ref{fig:HILO_limit_cycles}), and let $\mathcal{S} \subset \mathcal{X}$ be a surface transversal to $\mathcal{O}$. Then, $\mathcal{O}$ can be represented by a fixed point $x^* \in \mathcal{S}$ of the discrete-time system 
\begin{equation}\label{eq:P_map}
x_{k+1} = P(x_k)
\end{equation}
where $P : \mathcal{S} \to \mathcal{S}$ is the corresponding Poincar\'e map and $P(x^*)=x^*$. Intuitively, \eqref{eq:P_map} captures the discrete evolution of successive crossings through the surface $\mathcal{S}$. As the state $x$ evolves on $\mathcal{S}$ according to \eqref{eq:P_map}, the pose---that is, the location and heading---of the system in $\mathsf{G}$ is also updated in a discrete fashion. Given a coordinate representation $\mathsf{g} \in \mathsf{G}$, this discrete update can be captured by the \emph{displacement} map $H : \mathcal{S} \to \mathsf{G}$ so that
\begin{equation}\label{eq:H_map}
\mathsf{g}_{k+1} = \mathsf{g}_k + H(x_k) \enspace.  
\end{equation}
Combining \eqref{eq:P_map} and \eqref{eq:H_map}, we define the (augmented) \emph{stride map} $\hat{P} : \mathsf{G} \times \mathcal{S} \to \mathsf{G} \times \mathcal{S}$ by 
\begin{equation}\label{eq:Phat_map}
\hat{P} (\mathsf{g}, x) = 
\begin{bmatrix}
\mathsf{g} + H(x) \\
P(x)
\end{bmatrix}
\end{equation}
that captures the complete stride-to-stride evolution. 

Owing to the structure of~\eqref{eq:Phat_map}, planning can be naturally decomposed in a low-level gait generation problem governed by the state $x$ and its evolution on $\mathcal{S}$ according to \eqref{eq:P_map}, and a high-level planning problem in $\mathsf{G}$ captured by the displacement map $H$ (see Figs.~\ref{fig:HILO_limit_cycles}-\ref{fig:HILO_structure}). Consider now a system for which a finite number of limit-cycle gaits $\mathcal{O}_p$, $p\in\mathcal{P}$, is available. The dynamics around each limit cycle is represented by a pair 
\begin{equation}\label{eq:gaits}
\mathcal{G}_p = \{P_p, x^*_p\}, \qquad p \in \mathcal{P}
\end{equation}
where $P_p : \mathcal{S} \to \mathcal{S}$ is the corresponding Poincar\'e map and $P_p(x^*_p) = x^*_p$. We require that, for each $p \in \mathcal{P}$, $x^*_p$ is a locally exponentially stable fixed point of the corresponding map\footnote{For systems with impulse effects such as the ones arising in legged locomotion,~\cite[Corollary 1]{veer2019poincare} establishes that $\mathcal{O}_p$ is a locally exponentially stable limit cycle if, and only if, $x^*_p$ is a locally exponentially stable equilibrium of the corresponding map $P_p$. More details on this can be found in~\cite{veer2019poincare, Veer2019CDC}; we will not delve deeper into this matter here, focusing on the stride-to-stride evolution of the system.} $P_p$. Thus, we adopt a dynamical systems perspective to planning robot motions, according to which primitive movements are represented by (discrete-time) \emph{attractor landscapes} $\mathcal{G}_p$, as in~\cite{ijspeert2013dynamical}. We refer to these \ac{DMP}s as \emph{gait primitives}, and to the collection $\mathbb{G} = \{\mathcal{G}_p~|~p \in \mathcal{P} \}$ as the \emph{gait library}.   

Suppose now that a gait library $\mathbb{G}$ is available. The execution of each primitive in $\mathbb{G}$ results in displacement of the robot in $\mathsf{G}$ according to \eqref{eq:H_map}. Thus, for each $p \in \mathcal{P}$, augmenting the gait primitives $\mathcal{G}_p$ in \eqref{eq:gaits} with the corresponding displacements $\delta \mathsf{g} = H_p(x)$ provides \emph{actions} 
\begin{equation}\label{eq:actions}
\mathcal{A}_p = \{ H_p, P_p \} , \qquad p \in \mathcal{P}
\end{equation}
available for planning; see Figs.~\ref{fig:HILO_limit_cycles}--\ref{fig:HILO_structure}. Note that each action \eqref{eq:actions} represents a \emph{prediction}. That is, if $x$ is the state at the beginning of a stride, $\mathcal{A}_p$ predicts the displacement $H_p(x)$ and the state update $P_p(x)$ caused if the gait primitive $\mathcal{G}_p$ is engaged in the forthcoming stride. The predicted state $P_p(x)$ is then used to evaluate the displacements available to the planner at the next stride and so on.

This paper focuses on navigation problems, wherein a robot is tasked with reaching a target location in its workspace $\mathcal{W}$ that contains both static and moving obstacles. To achieve this, a planning algorithm equipped with a collection of actions $\mathbb{A} = \{\mathcal{A}_p~|~ p \in \mathcal{P} \}$ is used to find a sequence $\sigma : \mathbb{Z_+} \to \mathcal{P}$ that maps the stride number $k \in \mathbb{Z_+}$ to the index 
\begin{equation}\label{eq:switching}
p = \sigma(k)
\end{equation}
of the action $\mathcal{A}_p \in \mathbb{A}$ required at that stride. The suggested sequence of actions is then realized by concatenating the corresponding gait primitives \eqref{eq:gaits} according to \eqref{eq:switching}.

\subsection{Challenges and contributions} 
\label{subsec:challenges}

Before continuing, it is important to identify the main challenges of the approach described above.

\subsubsection{Symmetry properties and model structure}
Our approach exploits the ``triangular'' (cf. \cite[p. 43]{Isidori}) structure of the augmented stride map \eqref{eq:Phat_map} to separate gait generation from navigation. Implicit in \eqref{eq:Phat_map} is the assumption that the stride-to-stride evolution of the system's state $x \in \mathcal{S}$ and the net change $\delta \mathsf{g} = \mathsf{g}_{k+1} - \mathsf{g}_k$ of the pose of the system in $\mathsf{G}$ given by \eqref{eq:P_map} and \eqref{eq:H_map} respectively, do \emph{not} depend on the system's pose $\mathsf{g} \in \mathsf{G}$ at the beginning of the stride. In Section~\ref{sec:model} we show that under mild assumptions, a large class of 3D bipedal walking models possesses augmented stride maps that have the structure of \eqref{eq:Phat_map}.

\subsubsection{Practically stable path planning}  
The realization of the sequence of actions $\mathcal{A}_{\sigma(k)} \in \mathbb{A}$ according to \eqref{eq:switching} requires the implementation of the corresponding gait primitives $\mathcal{G}_{\sigma(k)} \in \mathbb{G}$. However, this sequence of gaits may result in loss of stability, \emph{even when} each gait primitive is itself stable. Defining what stability means in our setting is challenging; for, switching due to planning commands causes the system to be in an ``unrelenting'' transient phase, never converging to any of the underlying equilibrium behaviors. Beyond analysis, it is also important to be able to \emph{communicate} stability constraints to the planner so that the suggested plans are compatible with stable operation. These issues are addressed in Section~\ref{sec:safety}, where the set-based notion of \emph{practical stability} is adopted to identify a class of switching signals $\sigma$ that \emph{provably} guarantee stable operation as the system switches among the primitives in $\mathbb{G}$. These signals are characterized by a lower bound on the dwell time between switches. This way, the information relevant to stability is ``compressed'' to a single number or pair of numbers, which can be easily communicated to the planner.

\subsubsection{Reactive path planning and computational tractability} 
%
While feasible offline, the need to resort to numerical integration to evaluate the maps $P_p$ and $H_p$ challenges the implementation of optimization-based path planning approaches in real time. To enable computationally efficient implementation, Section~\ref{sec:tractability} below provides a method for extracting a collection $\tilde{\mathcal{A}}_p = \{ \tilde{H}_p, \tilde{P}_p \} , ~p \in \mathcal{P}$, of \emph{approximate actions}, represented as closed-form estimates of the maps $H_p$ and $P_p$ in \eqref{eq:actions}. This way, the planner has the ability to (approximately) predict the future stride-to-stride evolution of the system, and use this information to adjust its actions online while being consistent with practically stable gait switching without adversely affecting computational time. 
This is achieved in a primitive-based \ac{MPC} fashion in Section~\ref{sec:path-planning}.

\section{3D Bipedal Walking Models: Main Properties}
\label{sec:model}

This section identifies a general class of 3D walking robot models and feedback control laws that result in the ``triangular'' structure \eqref{eq:Phat_map} of the corresponding stride-to-stride map. Our discussion here is \emph{not} restricted to a particular robot model or  controller structure; rather, the focus is on the properties that enable the decomposition of the planning problem to task-relevant and gait-design components, as described in Section~\ref{subsec:augmented_stride}.  

\subsection{Dynamics: Symmetry properties}
 \label{subsec:symmetries}

Bipedal robot models can be represented as tree structures of rigid bodies; see Fig.~\ref{fig:3Dbiped_actions}. We will assume that each joint is revolute and allows a single \ac{DOF} rotation; note that this assumption does not entail significant loss of generality since multi-\ac{DOF} joints---e.g., ball joints---can be represented by multiple single-\ac{DOF} revolute joints with links of zero length in between~\cite{spong2005controlled}. Let $\{ \mathcal{I} \}$ be an inertia frame with its $Z$-axis aligned with the direction of gravity and the $X$-axis and $Y$-axis forming a plane that represents the ground surface. It is assumed here that the ground is flat and not deformable. Let $\mathrm{p}=(X,Y,Z) \in \mathbb{R}^3$ and $R \in \mathsf{SO}(3)$ be the position and orientation with respect to $\{ \mathcal{I} \}$ of a body-fixed frame $\{ \mathcal{B} \}$ attached at a reference (base) link. If  $q_\mathrm{r} \in \mathcal{Q}_\mathrm{r}$ includes the relative angles of the rest of the links that determine the shape of the multi-body chain, the configuration of the floating-base model can be captured by $(\mathrm{p}, R, q_\mathrm{r}) \in \mathbb{R}^3 \times \mathsf{SO}(3) \times \mathcal{Q}_\mathrm{r}$. Choosing a minimal representation of $\mathsf{SO}(3)$ by the yaw $q_1$, pitch $q_2$, and roll $q_3$ angles, we use $q = \begin{bmatrix} q_1 & q_2 & q_3 & q^\mathsf{T}_\mathrm{r} \end{bmatrix}^\mathsf{T}$ to describe the angular configuration of the model in a subset $\mathcal{Q}$ of $\mathsf{SO}(3) \times \mathcal{Q}_\mathrm{r}$ that contains physically reasonable configurations. 

Bipedal walking is composed by alternating single and double support phases.  In single support---left or right---we assume that the foot in contact with the ground remains stationary. Then, attaching $\{ \mathcal{B} \}$ at the support foot implies $\dot{\mathrm{p}} = 0$. Without loss of generality, we can take $Z=0$ and identify $\mathrm{p} = (X, Y, 0)$ with its non-zero components; thus, with a slight abuse of notation, $\mathrm{p}=(X,Y) \in \mathbb{R}^2$. During the single support phase, neither the kinetic $\mathcal{K}$ nor the potential $\mathcal{V}$ energy depend on the location $\mathrm{p}$ with respect to $\{ \mathcal{I}\}$. Hence, if $\mathcal{L}(q, \dot{q}) = \mathcal{K}(q, \dot{q}) - \mathcal{V}(q)$ is the corresponding Lagrangian, the equations of motion can be written as
\begin{equation}\label{eq:lagrangian}
\frac{d}{dt} \frac{\partial \mathcal{L}(q, \dot{q})}{\partial \dot{q}} -  \frac{\partial \mathcal{L}(q, \dot{q})}{\partial q}  = \mathcal{F}(q, u)
\end{equation}
where $\mathcal{F}(q, u) \!=\! B(q) u$ are the generalized forces due to the motor torques $u \in \mathcal{U}$.

The continuous-time evolution of single support is interrupted when the swing leg contacts the ground. This event occurs when the vertical distance of the tip of the swing leg from the ground reduces to zero. Due to the flat ground assumption, this distance does not depend on the position $\mathrm{p} \in \mathbb{R}^2$ of the biped relative to $\{ \mathcal{I} \}$. Hence, if $h^{\rm v}_{\rm sw} : \mathcal{Q} \to \mathbb{R}$ is the map that associates to a configuration $q \in \mathcal{Q}$ the vertical distance $h^{\rm v}_{\rm sw}(q)$ of the tip of the swing leg from the ground, contact can be captured by the switching surface
\begin{equation}\label{eq:impact_surface}
\mathcal{S} = \left\{ (q, \dot{q}) \in T\mathcal{Q} ~|~ h^{\rm v}_{\rm sw}(q) = 0, ~\frac{d}{dt} \left( h^{\rm v}_{\rm sw}(q) \right) < 0  \right\} \enspace.
\end{equation}
Crossing $\mathcal{S}$ triggers the double support phase, which is assumed to be instantaneous, and, as a result, can be modeled via a reset map $\Delta$ defined on $\mathcal{S}$ by its components
\begin{equation}\label{eq:impact_map}
q^+ = \Delta_q (q^-) ~~\text{and}~~ \dot{q}^+ = \Delta_{\dot{q}} (q^-, \dot{q}^-)
\end{equation}
where $(q^-, \dot{q}^-) \in \mathcal{S}$ are the angles and velocities prior to impact and $(q^+, \dot{q}^+)$ are the corresponding values after impact. Due to the flat, non-deformable ground, \eqref{eq:impact_map} does not depend on the location $\mathrm{p}$ with respect to $\{ \mathcal{I} \}$; see~\cite{3D-robotica2012, spong2005controlled} for details. 

For motion planning, we define the augmented state
\begin{equation}\label{eq:x_hat}
\hat{x} = 
\begin{bmatrix}
\mathrm{p}^\mathsf{T} & q_1 & x^\mathsf{T} 
\end{bmatrix}^\mathsf{T} \in \hat{\mathcal{X}}
\end{equation}
where $\hat{\mathcal{X}} = \mathbb{R}^2 \times T\mathcal{Q}$ and $x = \begin{bmatrix} q_2 & q_3 & q^\mathsf{T}_\mathrm{r} & \dot{q}^\mathsf{T} \end{bmatrix}^\mathsf{T}$. In this setting, the dynamics of the system can be written as
\begin{equation} \label{eq:state_space}
\dot{\hat{x}} = \hat{f} (\hat{x}) + \hat{g}(\hat{x}) u 
\end{equation}
where the first two components of the vectors fields $\hat{f}$ and $\hat{g}$ are zero by the fact that $\dot{\mathrm{p}}=0$ and the rest are computed by \eqref{eq:lagrangian}. The augmented switching surface $\hat{\mathcal{S}} = \{ \hat{x} \in \hat{\mathcal{X}} ~|~ (q, \dot{q}) \in \mathcal{S} \}$ and the augmented reset map $\hat{x}^+ = \hat{\Delta}(\hat{x}^-)$ can be analogously defined. Collecting all terms, the dynamics of the model can be written as a hybrid system
\begin{eqnarray}
\label{eq:hybridcontrolsystem}
\mathscr{HC} = \left\{ \hat{\mathcal{X}} , \mathcal{U} , \hat{\mathcal{S}}, (\hat{f},\hat{g}), \hat{\Delta} \right\} \enspace.
\end{eqnarray}
 
We are now ready to state properties of \eqref{eq:hybridcontrolsystem} that lead to the triangular structure of the augmented stride map \eqref{eq:Phat_map}. First, note that the domain $\hat{\mathcal{X}}$ in \eqref{eq:hybridcontrolsystem} is decomposed in two parts: (i) the ``task-relevant'' (planning) part $\mathsf{G}$ parametrized by the location $\mathrm{p}=(X, Y)$ of the robot and its heading $q_1$, and (ii) the ``gait'' part $\mathcal{X}$ parametrized by the states $x$ in \eqref{eq:x_hat}; that is, $\hat{\mathcal{X}} = \mathsf{G} \times \mathcal{X}$ as  in Section~\ref{subsec:augmented_stride}. Next, we examine the effect on \eqref{eq:hybridcontrolsystem} of translations along the $(X, Y)$-axes and rotations about the $Z$-axis of $\{\mathcal{I}\}$. Consider an element $\mathsf{g}=[\mathrm{r}^\mathsf{T} ~ \psi]^\mathsf{T}$ of $\mathsf{G}$ corresponding to a translation by $\mathrm{r} \in \mathbb{R}^2$ and a rotation by $\psi \in \mathbb{S}^1$, and let $\Psi: \mathsf{G} \times \hat{\mathcal{X}} \to \hat{\mathcal{X}}$ be the map 
\begin{equation}\label{eq:Psi}
\Psi(\mathsf{g}, \hat{x})= 
\begin{bmatrix}
(\mathrm{p} + \mathrm{r})^\mathsf{T} & q_1 + \psi & x^\mathsf{T}
\end{bmatrix}^\mathsf{T}
=\Psi_\mathsf{g} (\hat{x}) \enspace.
\end{equation}
The following proposition summarizes the properties of the dynamics \eqref{eq:hybridcontrolsystem} that are of interest to motion planning. 
\begin{prop}\label{prop:symmetry}
Consider the dynamics $\mathscr{HC}$ \eqref{eq:hybridcontrolsystem} and the map \eqref{eq:Psi}. Then, the following properties are true:
\begin{enumerate}[(i)]
\item \label{P_f} If $\hat{x}(t) \!=\! \hat{\varphi}(t, \hat{x}(0))$ is the zero-input ($u \equiv 0$) solution of \eqref{eq:state_space} with initial condition $\hat{x}(0) \in \hat{\mathcal{X}}$, then
\begin{equation}\nonumber
\Psi_\mathsf{g}( \hat{\varphi}(t, \hat{x}(0)) ) = \hat{\varphi}(t, \Psi_\mathsf{g}(\hat{x}(0))) ~~\text{for all}~~ \mathsf{g} \in \mathsf{G}.
\end{equation}
%
\item \label{P_S} If $\hat{x} \in \hat{\mathcal{S}}$, then $\Psi_\mathsf{g} (\hat{x}) \in \hat{\mathcal{S}} ~~\text{for all}~~ \mathsf{g} \in \mathsf{G}$. 
\item \label{P_Delta} If $\hat{x} \in \hat{\mathcal{S}}$, then $\Psi_\mathsf{g}( \hat{\Delta}(\hat{x}) ) \!=\! \hat{\Delta}(\Psi_\mathsf{g}(\hat{x})) ~\text{for all}~ \mathsf{g} \in \mathsf{G}$. 
\end{enumerate}
\end{prop}
A proof of Proposition~\ref{prop:symmetry} can be found in Appendix~\ref{app:sec:model}. Note that Parts (\ref{P_f}) and (\ref{P_Delta}) imply that the zero-input solution of \eqref{eq:state_space} and the map $\hat{\Delta}$ are equivariant under \eqref{eq:Psi}, and (\ref{P_S}) implies that $\hat{\mathcal{S}}$ is invariant under \eqref{eq:Psi}.

\subsection{Control: Symmetry-preserving feedback laws}
 \label{subsec:feedback}

We will be concerned with locomotion control laws that preserve the symmetry properties listed in Proposition~\ref{prop:symmetry}. To be specific, consider a feedback law $\Gamma : \hat{\mathcal{X}} \to \mathcal{U}$ that prescribes the actuator inputs during the single support phase by the rule $u = \Gamma(\hat{x})$. Let $\hat{x}(t) = \hat{\varphi}^\mathrm{cl}(t, \hat{x}(0))$ be  the solution with initial condition $\hat{x}(0) \in \hat{\mathcal{X}}$ of \eqref{eq:state_space} in closed loop with $\Gamma$, i.e., 
\begin{equation}\label{eq:state_space_cl}
\dot{\hat{x}} = \hat{f}^\mathrm{cl}(\hat{x})
\end{equation}
where $\hat{f}^\mathrm{cl}(\hat{x}) = \hat{f} (\hat{x}) + \hat{g}(\hat{x}) \Gamma(\hat{x})$. We then require that $\Gamma$ is designed so that $\hat{\varphi}^\mathrm{cl}$ satisfies the following property 
\begin{equation}\label{eq:sol_equivariance_cl}
\Psi_\mathsf{g}( \hat{\varphi}^\mathrm{cl}(t, \hat{x}(0)) ) = \hat{\varphi}^\mathrm{cl}(t, \Psi_\mathsf{g}(\hat{x}(0))) ~~ \text{for all}~~ \mathsf{g} \in \mathsf{G}
\end{equation}
which essentially extends the equivariance property of  Proposition~\ref{prop:symmetry}(\ref{P_f}) to the closed-loop system. It is emphasized here that while the properties listed in Proposition~\ref{prop:symmetry} are inherent to the class of bipedal robot models described in Section~\ref{subsec:symmetries}, property \eqref{eq:sol_equivariance_cl} hinges upon the design of the feedback controller $\Gamma$. This, in turn, depends on the actuation structure $\mathcal{F}(q, u) = B(q) u$ in \eqref{eq:lagrangian}, and thus, the design of a controller $\Gamma$ to satisfy \eqref{eq:sol_equivariance_cl} must be grounded to the  morphology of the robot; as an example, Section~\ref{sec:tractability} below examines a common underactuated configuration, in which \eqref{eq:sol_equivariance_cl} is naturally satisfied.

\subsection{The augmented stride map}
 \label{subsec:stride_map}

A full stride consists of right and left single support phases, each having the form of \eqref{eq:hybridcontrolsystem}. Due to the nontrivial hip width, the equations for right and left support differ, and will be distinguished by the indices $\mathrm{R}$ and $\mathrm{L}$, respectively. If $\Gamma_\mathrm{R}$ and $\Gamma_\mathrm{L}$ are feedback control laws designed so that property \eqref{eq:sol_equivariance_cl} is satisfied, the resulting closed-loop hybrid system is composed by the right and left support phases $\mathscr{H}_\mathrm{R} \!=\! \{ \hat{\mathcal{X}}_\mathrm{R} , \hat{\mathcal{S}}_\mathrm{R}, \hat{f}^\mathrm{cl}_\mathrm{R}, \hat{\Delta}_\mathrm{R} \}$ and $\mathscr{H}_\mathrm{L} \!=\! \{ \hat{\mathcal{X}}_\mathrm{L} , \hat{\mathcal{S}}_\mathrm{L}, \hat{f}^\mathrm{cl}_\mathrm{L}, \hat{\Delta}_\mathrm{L} \}$respectively, where $\hat{f}^\mathrm{cl}_\mathrm{R}$ and $\hat{f}^\mathrm{cl}_\mathrm{L}$ represent the corresponding closed-loop vector fields \eqref{eq:state_space_cl}.

To derive the augmented stride map, suppose that  $\hat{x}^-_\mathrm{R} \in \hat{\mathcal{S}}_\mathrm{R}$ is a state that results in a complete stride and let $\hat{x}^+_\mathrm{L} = \hat{\Delta}_\mathrm{R}(\hat{x}^-_\mathrm{R})$ be the initial condition for the ensuing left support phase and $\hat{x}_\mathrm{L}(t) = \hat{\varphi}^\mathrm{cl}_\mathrm{L}(t, \hat{x}^+_\mathrm{L})$ be the corresponding flow associated with the closed-loop dynamics $\hat{f}^\mathrm{cl}_\mathrm{L}$. If $\hat{T}_\mathrm{L}(\hat{x}^+_\mathrm{L})$ is the time-to-impact, $\hat{P}_\mathrm{LR} : \hat{\mathcal{S}}_\mathrm{R} \to \hat{\mathcal{S}}_\mathrm{L}$ defined by $\hat{P}_\mathrm{LR}(\hat{x}^-_\mathrm{R}) = \hat{\varphi}^\mathrm{cl}_\mathrm{L}(\hat{T}_\mathrm{L}(\hat{\Delta}_\mathrm{R}(\hat{x}^-_\mathrm{R})), \hat{\Delta}_\mathrm{R}(\hat{x}^-_\mathrm{R}))$ is the map taking the state prior to left support to the state prior to right support. The map $\hat{P}_\mathrm{RL} : \hat{\mathcal{S}}_\mathrm{L} \to \hat{\mathcal{S}}_\mathrm{R}$ can be defined analogously, and the (full) stride map $\hat{P} : \hat{\mathcal{S}}_\mathrm{R} \to \hat{\mathcal{S}}_\mathrm{R}$ is\footnote{A map $\hat{P}_\mathrm{L}$ can be  defined similarly. This map is diffeomorphic to $\hat{P}$, thus the choice between $\hat{P}$ or $\hat{P}_\mathrm{L}$ is arbitrary; in what follows, we use $\hat{P}$ by \eqref{eq:P_hat_def}.}
\begin{equation}\label{eq:P_hat_def}
\hat{P} = \hat{P}_\mathrm{RL} \circ \hat{P}_\mathrm{LR} \enspace.
\end{equation}

A direct consequence of Proposition~\ref{prop:symmetry}  and property  \eqref{eq:sol_equivariance_cl} is that \eqref{eq:P_hat_def} is equivariant under the action  $\Psi_\mathsf{g}$ defined by \eqref{eq:Psi}.
\begin{prop}\label{prop:symmetry_P}
Consider the augmented stride map $\hat{P}$ defined by \eqref{eq:P_hat_def} and the map \eqref{eq:Psi}. Then, for all $\mathsf{g} \in \mathsf{G}$, we have
\begin{equation} \label{eq:P_hat_equivariance}
\Psi_\mathsf{g} \circ \hat{P} = \hat{P} \circ \Psi_\mathsf{g} \enspace .
\end{equation}
\end{prop}
\noindent A proof of Proposition~\ref{prop:symmetry_P} can be found in Appendix~\ref{app:sec:model}. Intuitively, \eqref{eq:P_hat_equivariance} implies that rotating with respect to $Z$ and translating along $(X,Y)$ after completing a stride results in the same displacement as first rotating and translating the robot and then taking a stride. 

Property \eqref{eq:P_hat_equivariance} results in the ``triangular'' structure \eqref{eq:Phat_map} of $\hat{P}$. Indeed, let $\mathsf{g} = [\mathrm{p}^\mathsf{T}~q_1]^\mathsf{T} \in \mathsf{G}$ and  $x \in \mathcal{S}_\mathrm{R}$, where $\mathcal{S}_\mathrm{R}$ is the switching surface \eqref{eq:impact_surface} for the right support phase. Then, for $\hat{x} = [\mathsf{g}^\mathsf{T} ~ x^\mathsf{T}]^\mathsf{T} \in \hat{\mathcal{S}}_\mathrm{R} = \mathsf{G} \times \mathcal{S}_\mathrm{R}$, we have
\begin{equation}\nonumber
\hat{P} \left( \begin{bmatrix} \mathsf{g} \\ x \end{bmatrix} \right) =
\hat{P} \left( \Psi_\mathsf{g} \left( \begin{bmatrix} 0 \\ x \end{bmatrix} \right) \right) =
\Psi_\mathsf{g} \left( \hat{P} \left( \begin{bmatrix} 0 \\ x \end{bmatrix} \right) \right) 
\end{equation}
resulting in the augmented stride map \eqref{eq:Phat_map}, where $H : \mathcal{S}_\mathrm{R} \to \mathsf{G}$ is defined by the first three components of $\hat{P}(0, x)$ and $P: \mathcal{S}_\mathrm{R} \to \mathcal{S}_\mathrm{R}$ includes the rest of the components of $\hat{P}(0, x)$. As was discussed in Section~\ref{subsec:augmented_stride}, the triangular structure \eqref{eq:Phat_map} of $\hat{P}$ leads to a natural decomposition of the problem to a low-level gait generation component governed by the stride-by-stride update of the state $x$ according to $P$, and a high-level planning problem in $\mathsf{G}$ captured by $H$; see Fig.~\ref{fig:overview}.

\section{Practically Stable Gait Composition}
\label{sec:safety}

Suppose now that a collection of gait primitives $\mathcal{G}_p = \{P_p, x^*_p\}$ and corresponding actions $\mathcal{A}_p = \{H_p,~P_p\}$ are available to a planner. To keep the discussion general, we defer the details on how the gait primitives $\mathcal{G}_p$ and the corresponding actions $\mathcal{A}_p$ are designed to the following section. The task of the planner is to suggest a sequence $\sigma: \mathbb{Z}_+ \to \mathcal{P}$ mapping the current stride number $k$ to the index of the gait primitive $\sigma(k) \in \mathcal{P}$ that must be engaged at that stride so that a higher-level objective is achieved. Implementing the suggested sequence of actions naturally gives rise to a \emph{switching system with multiple equilibria}
\begin{equation}\label{eq:switched_system}
x_{k+1} = P_{\sigma(k)}(x_k) \enspace.
\end{equation}
This system differs from classical switching systems---such as those studied in~\cite{liberzon2003switching} and in references therein---in a fundamental way: the maps $P_p$ do \emph{not} share the same fixed point; that is, $x^*_p \neq x^*_q$ when $p \neq q$. In this section, we discuss a theoretical result on the stability properties of systems like \eqref{eq:switched_system} that is important in motion planning.

\subsection{Preliminaries and definitions}
\label{subsec:stability_def}

As was mentioned in Section~\ref{subsec:challenges}, defining stability in switching systems with multiple equilibria can be challenging; for, continuing switching in response to varying operation conditions causes the system to shift to different equilibria, \emph{never} converging to any one of them. To certify stability in \eqref{eq:switched_system}, we will adopt the notion of \emph{practical stability}, which is defined as follows.
\begin{definition}[Adapted from \cite{Zhai2003CDC}]\label{def:practical-stability}
Let $\Omega$ and $\Omega_0$ be given sets with $\Omega$ being closed and bounded and $\Omega_0 \subset \Omega$. The switched system \eqref{eq:switched_system} is \emph{practically stable} with respect to $\Omega_0$ and $\Omega$, if $x_0 \in \Omega_0$ implies $x_k \in \Omega$ for all $k\in\mathbb{Z}_+$.
\end{definition}

In what follows, we will provide \emph{explicitly computable} characterizations of the sets $\Omega_0$ and $\Omega$ for \eqref{eq:switched_system} under suitable conditions on the individual gait primitives in $\mathbb{G}$ and on the switching signal $\sigma$. We begin with the requirement that each gait primitive $\mathcal{G}_p \in \mathbb{G}$ is locally exponentially stable. Safety certificates of this form can be obtained through suitable Lyapunov functions~\cite[Chapter 13]{haddad2008nonlinear} defined as follows.
\begin{definition}\label{def:exp-Lyap}
Consider the discrete-time system \eqref{eq:P_map} defined on $\mathcal{S}$. Let $\mathcal{D}$ be an open subset of $\mathcal{S}$ with $x^* \in \mathcal{D}$. A continuous function $V: \mathcal{D} \rightarrow \mathbb{R}$ is an exponential Lyapunov function, if, for all $x \in \mathcal{D}$,
\begin{equation}\label{eq:V_1}
\underline{\alpha} (\| x - x^* \|) \leq V (x) \leq \overline{\alpha}(\| x - x^*\|)
\end{equation}
\begin{equation}\label{eq:V_2}
V(P (x)) \leq \lambda \cdot V (x)
\end{equation}
where $\underline{\alpha}$, $\overline{\alpha}$ are class-$\mathcal{K}_\infty$ functions and $0 <  \lambda < 1$.
\end{definition}

The second requirement we need is that switching must be sufficiently slow. A convenient way to express this condition is based on the notion of dwell time, which is defined below.
\begin{definition}\label{def:dwell-time}
Consider a switching signal $\sigma : \mathbb{Z_+} \to \mathcal{P}$.
\begin{enumerate}[(i)]
\item \label{def:dwell-time-fixed} Let $\{k_1, k_2, ... \}$ be the switching times of $\sigma$. Then, $\sigma$ has \emph{dwell time} $N_{\rm d} > 0$ if, for any $k_i \in \{k_1, k_2, ... \}$,  
\begin{equation}\label{eq:fixed-dwell-time-def}
\sigma(k)=\sigma(k_i) ~~\text{for all}~~ k \in [k_i, k_i+ N_{\rm d}) \enspace.
\end{equation}
%
\item \label{def:dwell-time-avg} Let $N_\sigma(\overline{k}, \underline{k}) \in\mathbb{Z}_+$ be the number of switches over the discrete-time interval $[\underline{k},\overline{k})$. Then, $\sigma$ has \emph{average dwell time} $N_{\rm a}>0$ if, for any $\overline{k}>\underline{k}\geq 0$, 
\begin{equation} \label{eq:avg-dwell-time-def}
N_\sigma(\overline{k}, \underline{k}) \leq N_0 + \frac{k-\underline{k}}{N_{\rm a}} ~~\text{for all}~~ k \in [\underline{k}, \overline{k}) 
\end{equation}
where $N_0\!>\!0$ is a finite constant called the \emph{chatter bound}.
\end{enumerate}
\end{definition}
 
According to Definition~\ref{def:dwell-time}, the dwell time $N_\mathrm{d}>0$ of a switching signal is simply the minimum number of strides between two successive switches; equivalently, $k_{i+1} - k_i \geq N_\mathrm{d}$ for every pair of successive switching instants. The average dwell-time constraint, on the other hand, relaxes\footnote{Note that the family of signals that satisfy the constraint \eqref{eq:avg-dwell-time-def} for $N_0=1$ and for some $N_\mathrm{a} >0$ is precisely the family of signals with (fixed) dwell time $N_\mathrm{d} = N_\mathrm{a}$. Thus, the family of signals with dwell time $N_\mathrm{d}>0$ is included in the family of signals that satisfy \eqref{eq:avg-dwell-time-def} with $N_0 \geq 1$ and $N_\mathrm{a} = N_\mathrm{d}$.} this requirement, thus allowing more frequent switching over some period, as long as this is compensated by slower switching so that the constraint \eqref{eq:avg-dwell-time-def} is satisfied. Note that the notion of dwell time compresses the information regarding the frequency of switching to a \emph{single} number $N_\mathrm{d}$ for the fixed dwell time case, or to the  \emph{pair} $(N_0, N_\mathrm{a})$ for the average dwell time. This information can be conveniently communicated to the planning algorithm, which can now devise plans that respect the stability limitations imposed by the low-level gait switching system \eqref{eq:switched_system} so that stability in the sense of Definition~\ref{def:practical-stability} is guaranteed.

\subsection{Stable switching: Set constructions and explicit bounds}
\label{subsec:sets}

\begin{figure}[t]
\begin{centering}
\includegraphics[width=0.9\columnwidth]{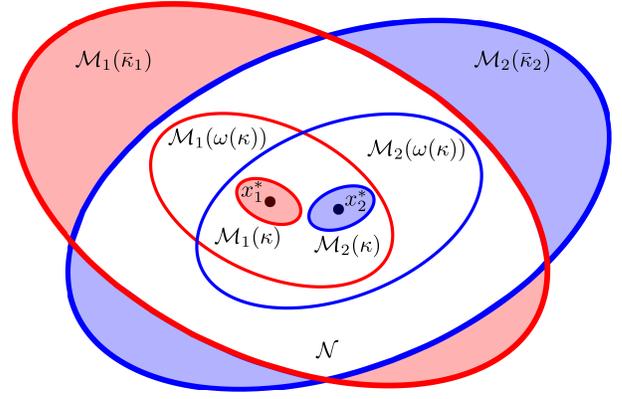} 
\par\end{centering}
\caption{Illustration of the sets associated with Theorem~\ref{thm:dwell-time} for $\mathcal{P}=\{1, 2\}$.}
\vspace{-0.2in}
\label{fig:set-construction-local} 
\end{figure}

We begin with a description of the set construction of Fig.~\ref{fig:set-construction-local}, which enables stating and proving our key theoretical result. We  assume that, for each gait $\mathcal{G}_p = \{ P_p, x^*_p \}$, a Lyapunov function $V_p$ is available that satisfies Definition~\ref{def:exp-Lyap} over an open set $\mathcal{D}_p \subset \mathcal{S}_p$ with appropriate $\underline{\alpha}_p, \overline{\alpha}_p \in \mathcal{K}_\infty$ and $\lambda_p \in (0, 1)$. 

Due to the fact that the fixed points $x^*_p$ of \eqref{eq:switched_system} are distinct, it must be ensured that switching among them is well defined. To provide a feasibility condition for switching, let $\overline{\kappa}_p > 0$ be such that the sublevel set $\mathcal{M}_p(\overline{\kappa}_p)$ of $V_p$ satisfies
\begin{equation} \nonumber 
\mathcal{M}_p(\overline{\kappa}_p) = \{x \in \mathbb{R}^n ~|~ V_p(x) \leq \overline{\kappa}_p \} \subset \mathcal{D}_p \enspace.
\end{equation}
It can be recognized that each $\mathcal{M}_p(\overline{\kappa}_p)$ represents a compact inner approximation of the \ac{BOA} associated with the fixed point $x^*_p$ of $P_p$; such approximations can be verified computationally via \ac{SOS} programming~\cite{parrilo2000structured, majumdar2013robust}; see Appendix~\ref{app:sec:implementation}. For notational convenience, we define
\begin{equation}\label{eq:feasibility-1}
\mathcal{N} = \bigcap_{p\in\mathcal{P}} \accentset{\circ}{\mathcal{M}}_p (\bar{\kappa}_p)
\end{equation}
which is an open subset of $\mathbb{R}^n$; see Fig.~\ref{fig:set-construction-local} for an illustration. We then require the following \emph{feasibility} condition
\begin{equation}\label{eq:feasibility-2}
x^*_p \in \mathcal{N}~~~~\text{for all}~~~~ p\in\mathcal{P}
\end{equation}
which ensures that switching in \eqref{eq:switched_system} is well defined. Note that all the set constructions that follow take place in $\mathcal{N}$.   

Next, choose a $\kappa > 0$ and define the $\kappa$-sublevel sets $\mathcal{M}_p(\kappa)=\{x\in\mathbb{R}^n~|~V_p(x)\leq \kappa\}$ as in Fig.~\ref{fig:set-construction-local}. Let 
\begin{equation}\nonumber 
\mathcal{M}(\kappa)=\bigcup_{p\in\mathcal{P}} \mathcal{M}_p(\kappa)
\end{equation}
be the union of these sublevel sets over $\mathcal{P}$. Define
\begin{equation}\label{eq:omega-def}
\omega(\kappa)=\max_{p\in\mathcal{P}}\max_{x\in\mathcal{M}(\kappa)} V_p(x) \enspace.
\end{equation}
The definition of $\omega(\kappa)$ by  \eqref{eq:omega-def} implies that $V_p(x) \leq \omega({\kappa})$ for all $x \in \mathcal{M}(\kappa)$ and for all $p \in \mathcal{P}$. Thus, as illustrated in Fig.~\ref{fig:set-construction-local}, $\omega({\kappa})$ effectively enlarges the sets $\mathcal{M}_p(\kappa)$ to the sets $\mathcal{M}_p(\omega(\kappa))$, so that 
\begin{equation}\label{eq:N_sub_Munder}
\mathcal{M}(\kappa) 
\subset \bigcap_{p\in\mathcal{P}} \mathcal{M}_p(\omega(\kappa)) \enspace.
\end{equation}

To bound ``energy" gain due to switching, let\footnote{The exclusion of $\accentset{\circ}{\mathcal{M}}_p(\kappa)$ in the supremum in \eqref{eq:mu-def} is to prevent the possibility of the denominator becoming $0$, since $x_p^* \in \accentset{\circ}{\mathcal{M}}_p(\kappa)$ and $V_p(x_p^*)=0$.}
\begin{equation}\label{eq:mu-def}
\mu(\kappa):=\max_{p,r\in\mathcal{P}} \sup_{x\in \mathcal{N}\setminus \accentset{\circ}{\mathcal{M}}_p(\kappa)} \frac{V_r(x)}{V_p(x)} 
\end{equation}
so that, for any pair $p,r\in\mathcal{P}$ of subsystems, switching from $p$ to $r$ satisfies 
\begin{align}\label{eq:mu-bound-V}
V_r(x) \leq \mu(\kappa)V_p(x) ~~~\text{for all}~~~ x \in \mathcal{N}\setminus \accentset{\circ}{\mathcal{M}}_p(\kappa).
\end{align}
Note that the interchangeability of the indices $p$ and $r$ implies $\mu(\kappa)\geq 1$. Indeed, as long as $x \in \mathcal{N}\setminus \accentset{\circ}{\mathcal{M}}_r(\kappa)$, we can also write $V_p(x) \leq \mu(\kappa) V_r(x)$ . Thus, when $x \in \mathcal{N} \setminus (\accentset{\circ}{\mathcal{M}}_p(\kappa) \cup \accentset{\circ}{\mathcal{M}}_r(\kappa))$, we have $V_p(x) \leq \mu(\kappa)^2 V_p(x)$, from which it follows that $\mu(\kappa) \geq 1$ since $V_p$ is positive for all $x \in \mathcal{N} \setminus (\accentset{\circ}{\mathcal{M}}_p(\kappa) \cup \accentset{\circ}{\mathcal{M}}_r(\kappa))$.

We are now ready to state the main result of this section; refer to Fig.~\ref{fig:set-construction-local} for the associated set constructions. Intuitively, this result establishes conditions for the exponential decay between successive switches to dominate over the potentially expansive effect of switching so that the state of the switching system \eqref{eq:switched_system} remains bounded.

\begin{theorem}\label{thm:dwell-time}
Consider \eqref{eq:switched_system} where $\sigma: \mathbb{Z}_+ \to \mathcal{P}$ is a switching signal and $\mathcal{P}$ is a finite index set. Assume that for each $p\in\mathcal{P}$, $x^*_p$ is a locally exponentially stable equilibrium of $P_p$ certified by a Lyapunov function $V_p$ satisfying Definition~\ref{def:exp-Lyap} with $\lambda_p \in (0,1)$, and let $\lambda=\max_{p\in\mathcal{P}}\lambda_p$. Suppose that the feasibility condition \eqref{eq:feasibility-1}-\eqref{eq:feasibility-2} is satisfied, and define $\omega(\kappa)$ by \eqref{eq:omega-def} and $\mu(\kappa)$ by \eqref{eq:mu-def}.  
\begin{enumerate}[(i)]
\item Assume that $\kappa>0$ can be chosen so that 
\begin{equation}\label{eq:feasible-0-inp-fixed}
\bigcup_{p\in\mathcal{P}} \mathcal{M}_p (\omega(\kappa)) \subset \mathcal{N} \enspace.
\end{equation}
Then, for any $\sigma$ that satisfies \eqref{eq:fixed-dwell-time-def} in Definition~\ref{def:dwell-time} with
\begin{equation}\label{eq:fixed-dwell}
N_{\rm d}\geq \overline{N}_{\rm d} 
\end{equation}
where
\begin{equation}\nonumber
\overline{N}_{\rm d}= \frac{\ln{\mu(\kappa)}}{\ln{(1/\lambda)}}
\end{equation}
the switched system \eqref{eq:switched_system} is practically stable with respect to the compact sets $\Omega_0$ and $\Omega$ defined by 
\begin{equation} \nonumber
\Omega_0=\bigcap_{p\in\mathcal{P}} \mathcal{M}_p(\omega(\kappa)) ~~\mbox{and}~~ \Omega= \bigcup_{p\in\mathcal{P}} \mathcal{M}_p(\omega(\kappa)) \enspace. 
\end{equation}
%
\item Assume that $\kappa>0$ and $\overline{N}_0 \geq 1$ can be chosen so that 
\begin{equation}\label{eq:feasible-0-inp}
\bigcup_{p\in\mathcal{P}} \mathcal{M}_p (\mu(\kappa)^{\overline{N}_0}\omega(\kappa)) \subset \mathcal{N} \enspace.
\end{equation}
Then, for any $\sigma$ that satisfies \eqref{eq:avg-dwell-time-def} in Definition~\ref{def:dwell-time} with  
\begin{equation}\label{eq:avg-dwell}
N_0 \leq \overline{N}_0 \text{~~and~~} N_{\rm a}\geq \overline{N}_{\rm a} 
\end{equation}
where
\begin{equation} \nonumber
\overline{N}_{\rm a}= \frac{\ln{\mu(\kappa)}}{\ln{(1/\lambda)}} 
\end{equation}
the switched system \eqref{eq:switched_system} is practically stable with respect to the compact sets $\Omega_0$ and $\Omega$ defined by 
\begin{equation} \nonumber
\label{eq:sets-avg-dwell}
\Omega_0=\bigcap_{p\in\mathcal{P}} \mathcal{M}_p(\omega(\kappa)) ~~\mbox{and}~~ \Omega = \bigcup_{p\in\mathcal{P}} \mathcal{M}_p(\mu(\kappa)^{N_0}\omega(\kappa)) \enspace. 
\end{equation}
\end{enumerate}
\end{theorem}

A proof of this result can be found in Appendix~\ref{sec:proof}. Essentially, Theorem~\ref{thm:dwell-time} identifies switching signals that ensure practical stability of \eqref{eq:switched_system} with respect to sets $\Omega_0$ and $\Omega$ that are explicitly described. Furthermore, it furnishes closed-form expressions for bounds on the (fixed) dwell time \eqref{eq:fixed-dwell} and on the average dwell time \eqref{eq:avg-dwell}, which can be easily communicated to the planner to restrict its choices to sequences of primitives that are compatible with the system's stability constraints. Note that for unrestricted switching---i.e., when the planner is allowed to switch at every stride---it is desirable to design the gait library so that $N_{\rm d}=1$.

\section{Computationally Tractable Planning Actions}
\label{sec:tractability}

\begin{figure}[t]
\begin{centering}
\includegraphics[width=0.5\columnwidth]{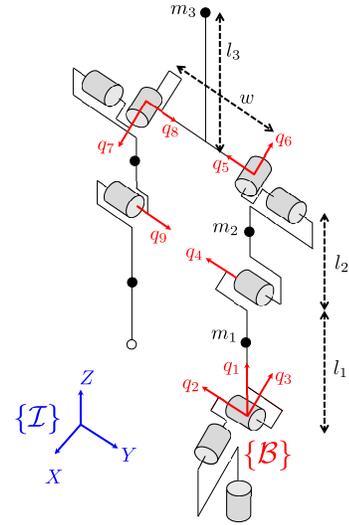} 
\par\end{centering}
\vspace{-0.1in}
\caption{A 3D biped with a choice of generalized coordinates; $\{\mathcal{I}\}$ is the inertial frame and $\{\mathcal{B}\}$ the body-fixed frame attached at the support foot.}
\vspace{-0.1in}
\label{fig:model} 
\end{figure}

To illustrate the approach, we consider here a fairly common instantiation of the class of models described in Section~\ref{sec:model}; see Fig.~\ref{fig:model}. In this model, the legs are identical and consist of two links coupled via an 1 DOF revolute knee joint. Each leg is connected to the torso via a 2 DOF revolute hip joint allowing motion in the sagittal and frontal planes, and terminates at a foot that is articulated by a 3 DOF (ball) ankle joint. The multi-DOF hip and ankle joints can be represented as multiple single-DOF revolute joints connected with links of zero length; see Fig.~\ref{fig:model}. We assume that the mass of each foot is small compared to the rest of the links and can be considered negligible\footnote{This assumption combined with the assumption that the robot's feet neither slip nor rotate when in contact with the ground lead to a model that is effectively equivalent to point-foot walking models~\cite{Chevallereau2010IROS, 3D-robotica2012}. Thus, the terms ``foot'' and ``ankle'' will both be used to denote the leg end.}. Thus, in the single support phase, the model features a total of 9 DOFs described by the corresponding yaw $q_1$, pitch $q_2$, and roll $q_3$ angles and by the joint angles $q_\mathrm{r} = (q_4, ..., q_9)$, as in Fig.~\ref{fig:model}. For simplicity, and to keep the focus on gait composition rather than controller design, we adopt here a typical actuation structure according to which the yaw and pitch angles of the ankle joint of the support leg are unactuated while the rest of the DOFs are actuated. This is consistent with other similar models as in~\cite{3D-robotica2012, motahar2016composing, Veer2017IROS}. As a result, the actuation structure $\mathcal{F}(q,u)$ in \eqref{eq:lagrangian} does not depend on the yaw angle $q_1$ and property \eqref{eq:sol_equivariance_cl} is  satisfied when the controller does not\footnote{This structure of the feedback controller is \emph{not} restrictive; it is, in fact, satisfied by many control design methods for 3D bipedal walking, including zero dynamics controllers~\cite{3D-robotica2012, Chevallereau2010IROS, hamed2014event}, geometric reduction controllers~\cite{gregg2009reduction, sinnet20093d}, passivity-based controllers~\cite{Spong2007RAM} and optimization-based  controllers~\cite{Nguyen2015RSS}; see also~\cite{grizzle2014models} and references therein.} rely on the feedback of the location $\mathrm{p}$ and orientation $q_1$. 

\vspace{+0.1in}

\subsection{Reduced-order planning actions via hybrid zero dynamics}
\label{subsec:reduction-feedback}

To harness the computational advantages of dimensional reduction, we will use the \ac{HZD} method to design walking controllers. In our setting, to ensure that the control law does not break the symmetry of the open-loop system so that the condition \eqref{eq:sol_equivariance_cl} is satisfied, we choose output functions that do not depend on $\mathrm{p}$ and $q_1$; i.e.
\begin{equation}\label{eq:output_function}
y = h(q_2, q_3, q_\mathrm{r}) \enspace.
\end{equation}
Differentiating \eqref{eq:output_function} twice along the model dynamics results in
\begin{equation}\nonumber
\ddot{y} = \alpha(x) + \zeta(q_2, q_3, q_\mathrm{r}) u
\end{equation}
where $\alpha$ and $\zeta$ do not depend on $\mathrm{p}$ and $q_1$. Assuming that the matrix $\zeta(q_2, q_3, q_\mathrm{r})$ is invertible, the controller
\begin{equation}\label{eq:fdb-general}
u = \Gamma(x) = - \zeta^{-1}(q_2, q_3, q_\mathrm{r}) \cdot \alpha(x)
\end{equation}
renders the surface
\begin{equation}\label{eq:Z_reduced}
\mathcal{Z} \!=\! \left\{ x \in \mathcal{X} \big|~  h(q_2, q_3, q_\mathrm{r}) \!=\! 0,~ \frac{d}{dt}\left( h(q_2, q_3, q_\mathrm{r}) \right) \!=\! 0 \right\}
\end{equation}
forward invariant\footnote{Forward invariance implies that if $x(t_0) \in \mathcal{Z}$ for some $t_0 \geq 0$, then the $x$-part of the closed-loop solution satisfies $x(t) \in \mathcal{Z}$ for all $t \geq t_0$.} under the the $x$-part of the solution of the closed-loop dynamics.

To obtain a library of gait primitives, we design a collection of controllers indexed by $p \in \mathcal{P}$ by employing the aforementioned procedure in each support phase using outputs
\begin{equation}\label{eq:leg_output_RL}
y_\mathrm{R} = h_{\mathrm{R},p}(q_2, q_3, q_\mathrm{r}) ~~\text{and}~~  y_\mathrm{L} = h_{\mathrm{L},p}(q_2, q_3, q_\mathrm{r}) 
\end{equation}
for the right and left support phases, respectively. This way, a pair $\Gamma_p = \{\Gamma_{\mathrm{R}, p}, \Gamma_{\mathrm{L}, p}\}$ of control laws is derived for each $p \in \mathcal{P}$. As above, the procedure results in a collection of forward-invariant, lower-dimensional surfaces $\mathcal{Z}_{\mathrm{R},p}$ and $\mathcal{Z}_{\mathrm{L},p}$.

\begin{figure}[t]
\vskip +5pt    
    \centerline{
    \subfigure[]{ 
    \includegraphics[width=0.4\columnwidth]{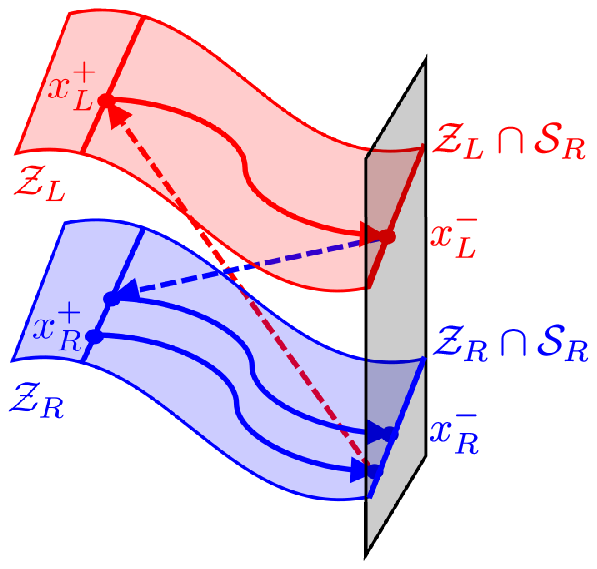}
    \label{fig:zero-dynamics}}
    \subfigure[]{  
    \includegraphics[width=0.4\columnwidth]{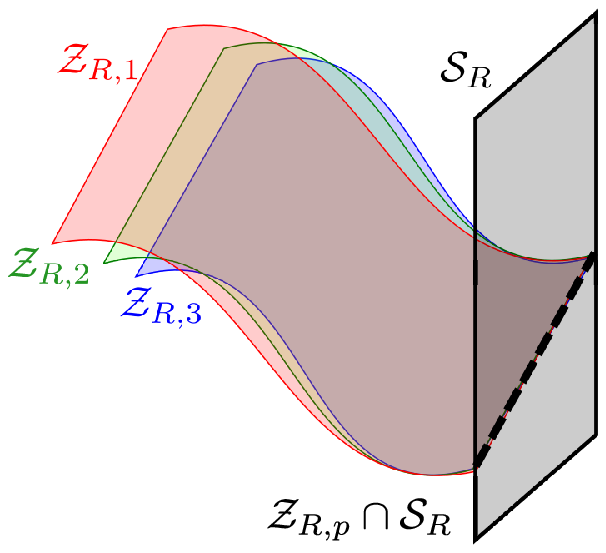}
    \label{fig:multi-surface}}
    } 
\vskip -5pt
\caption{Properties C.\ref{O1} and C.\ref{O2} for achieving dimensional reduction.}
\end{figure}

To fully take advantage of dimensional reduction when switching among the controllers $\Gamma_p$, we will require that the outputs \eqref{eq:leg_output_RL} are designed so that the corresponding zero dynamics surfaces $\mathcal{Z}_{\mathrm{R},p}$ and $\mathcal{Z}_{\mathrm{L},p}$ satisfy the conditions:
\begin{enumerate}[{{C}.1)}]
\item \label{O1} for each $p \in \mathcal{P}$, we have\footnote{Note that $\mathcal{S}_\mathrm{R}$ and $\mathcal{S}_\mathrm{L}$ correspond to the ground surface $\mathcal{S}$ under the right and left support phase coordinates, respectively.}
\begin{equation}\nonumber
\Delta_\mathrm{L} \left( \mathcal{S}_\mathrm{L}  \cap \mathcal{Z}_{\mathrm{L},p}  \right) \subset \mathcal{Z}_{\mathrm{R},p} ~~\text{and}~~ \Delta_\mathrm{R} \left( \mathcal{S}_\mathrm{R}  \cap \mathcal{Z}_{\mathrm{R},p} \right) \subset \mathcal{Z}_{\mathrm{L},p}
\end{equation}
%
\item \label{O2} for all $p, r\in\mathcal{P}$, we have\footnote{Condition~C.\ref{O2} is stated in terms of the right support phase zero dynamics surfaces $\mathcal{Z}_{\mathrm{R},p}$ due to our earlier choice to work with stride maps \eqref{eq:P_hat_def} mapping right leg pre-impact states to right leg pre-impact states.} $\mathcal{S}_\mathrm{R} \cap \mathcal{Z}_{\mathrm{R},p} =  \mathcal{S}_\mathrm{R} \cap \mathcal{Z}_{\mathrm{R},r}$. 
\end{enumerate}
Condition~C.\ref{O1} results in a well-defined \ac{HZD} for each  $\Gamma_p$; see Fig.~\ref{fig:zero-dynamics} and~\cite{3D-TRO2009, 3D-robotica2012} for details.
Condition~C.\ref{O2}, on the other hand, requires that all $\mathcal{Z}_{\mathrm{R},p}$ have a \emph{common} intersection with the switching surface $\mathcal{S}_\mathrm{R}$, as shown in Fig.~\ref{fig:multi-surface}. To avoid clutter, we drop the index $\mathrm{R}$ and denote this common intersection by 
\begin{equation} \nonumber
\mathcal{S} \cap \mathcal{Z} = \mathcal{S}_\mathrm{R} \cap \mathcal{Z}_{\mathrm{R},p}~~\text{for all}~~ p \in \mathcal{P} \enspace.
\end{equation}
Thus, condition~C.\ref{O2} ensures that switching among controllers $\Gamma_p$ does \emph{not} excite dynamics outside of $\mathcal{S} \cap \mathcal{Z}$, and effectively extends dimensionality reduction across the controllers. Note that conditions C.\ref{O1} and C.\ref{O2} can be easily and systematically satisfied using suitable polynomial functions to design the outputs \eqref{eq:leg_output_RL}; the details are provided in Appendix~\ref{app:HZD-controller}. 

The end result of the aforementioned control design is a collection of well-defined \emph{reduced-order} augmented stride maps $\hat{P}^\mathrm{red}_p : \mathsf{G} \times (\mathcal{S} \cap \mathcal{Z}) \to  \mathsf{G} \times (\mathcal{S} \cap \mathcal{Z})$ given by
\begin{equation} \nonumber 
\hat{P}^\mathrm{red}_p(\mathsf{g},z) = 
\begin{bmatrix} 
\mathsf{g}+ H_p|_{\mathcal{S} \cap \mathcal{Z}}(z) \\ 
P_p|_{\mathcal{S} \cap \mathcal{Z}}(z)   
\end{bmatrix} 
\end{equation}
where $z$ is a set of coordinates for the 2-dimensional surface $\mathcal{S} \cap \mathcal{Z}$, and $P_p|_{\mathcal{S} \cap \mathcal{Z}}$ and $H_p|_{\mathcal{S} \cap \mathcal{Z}}$ are the restrictions of the corresponding gait and displacement components of the augmented stride map \eqref{eq:Phat_map} on $\mathcal{S} \cap \mathcal{Z}$. As in~\cite{3D-TRO2009}, it can be shown that  a valid set of coordinates on $\mathcal{S} \cap \mathcal{Z}$ is $z = (\dot{q}_1,\dot{\theta})$, where $\dot{\theta}$ is the rate of the angle $\theta$ of the line connecting the point of contact of the support leg with the corresponding hip joint; see also Appendix~\ref{app:HZD-controller}. 

The objective of the aforementioned control laws  $\Gamma_p$ is to generate suitable limit-cycle walking gaits, described by (locally) exponentially stable fixed points $z^*_p \in \mathcal{S} \cap \mathcal{Z}$ of the maps $P_p|_{\mathcal{S} \cap \mathcal{Z}}$. This way, a library of \emph{reduced-order} gait primitives represented by pairs $\mathcal{G}_p|_{\mathcal{S} \cap \mathcal{Z}} = \{P_p|_{\mathcal{S} \cap \mathcal{Z}},~z^*_p\}$ is obtained, which in turn gives rise to a collection of \emph{reduced-order} planning actions
\begin{equation}\label{eq:reduced-actions}
\mathcal{A}_p|_{\mathcal{S} \cap \mathcal{Z}} = \left\{H_p|_{\mathcal{S} \cap \mathcal{Z}}, P_p|_{\mathcal{S} \cap \mathcal{Z}} \right\} , \qquad p \in \mathcal{P} \enspace.
\end{equation}
The benefit is that the reduced-order actions are defined on the 2-dimensional surface $\mathcal{S} \cap \mathcal{Z}$ as opposed to their full-order counterparts \eqref{eq:actions} that are defined on the 17-dimensional $\mathcal{S}$.

\begin{figure*}[b!]
\vskip -10pt    
    \centerline{
    \hspace{-5pt}
    \subfigure[]{ 
    \includegraphics[]{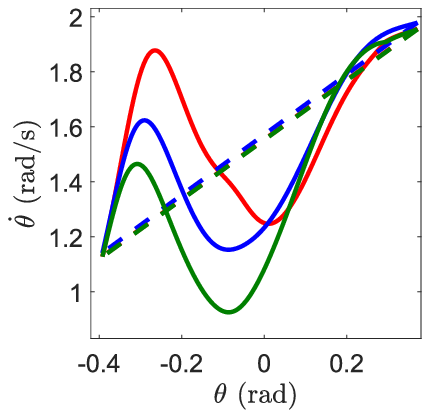} 
    \label{fig:gaits}}
    \hspace{-10pt}
    \subfigure[]{  
    \includegraphics[]{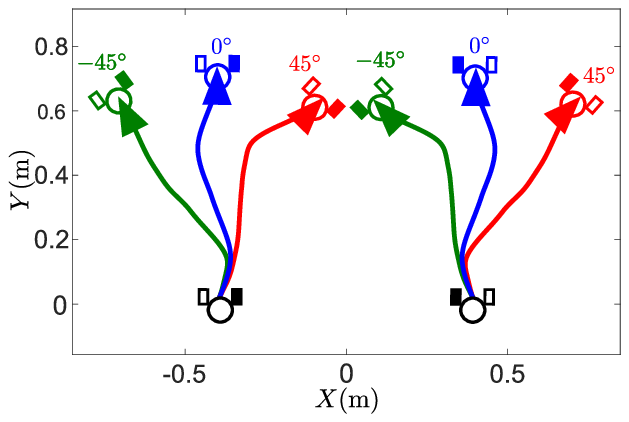}
    \label{fig:actions}}
    \hspace{-10pt}
    \subfigure[]{  
    \includegraphics[]{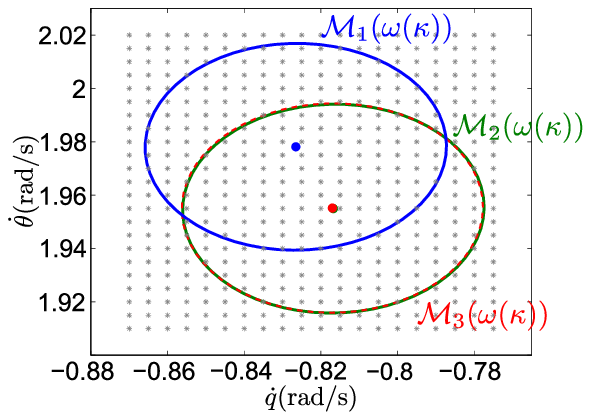}
    \label{fig:mesh}}
    } 
\vskip -5PT
\caption{(a) A family of limit cycles corresponding to 3D walking reduced-order gait primitives $\mathcal{G}_p = \{P_p|_{\mathcal{S} \cap \mathcal{Z}}, z^*_p\}$, $p \in \{1,2,3\}$. (b) Displacements  $\delta \mathsf{g} = H_p|_{\mathcal{S} \cap \mathcal{Z}}(z^*_p)$ caused by the limit cycles in Fig.~\ref{fig:gaits}. The solid triangles represent the predictions of approximate reduced-order planning actions $\tilde{H}_p|_{\mathcal{S} \cap \mathcal{Z}}$. (c) The sets $\mathcal{M}_p(\omega(\kappa))$ corresponding to the primitives $\mathcal{G}_p$ of Fig.~\ref{fig:gaits}; here $\mathcal{M}_2(\omega(\kappa))$ and $\mathcal{M}_3(\omega(\kappa))$ almost coincide. The intersection and union of these sets give $\Omega_0$ and $\Omega$, respectively, ensuring practically stable switching. Grey stars denote the mesh used to extract approximate planning actions.}
\label{fig:gait_actions}
\end{figure*}

\subsection{Approximate reduced-order planning actions}
\label{subsec:reduction-approximation}

Evaluating the reduced actions \eqref{eq:reduced-actions} in real-time is still challenging due to the required numerical integration, which, despite the reduced dimension, is computationally expensive. To address this issue, we will approximate the maps in \eqref{eq:reduced-actions} with polynomial functions, thus obtaining a collection of \emph{approximate reduced-order planning actions}
\begin{equation}\label{eq:approximate-reduced-actions}
\tilde{\mathcal{A}}_p|_{\mathcal{S} \cap \mathcal{Z}} = \left\{ \tilde{H}_p|_{\mathcal{S} \cap \mathcal{Z}}, \tilde{P}_p|_{\mathcal{S} \cap \mathcal{Z}} \right\} , \qquad p \in \mathcal{P} \enspace.
\end{equation}
Being polynomials, these approximate actions can be computed efficiently, providing sufficiently accurate predictions of the state $\tilde{P}_p|_{\mathcal{S} \cap \mathcal{Z}}(z)$ of the robot at the next step as well as the corresponding displacement $\tilde{H}_p|_{\mathcal{S} \cap \mathcal{Z}}(z)$.

Obtaining $\tilde{\mathcal{A}}_p|_{\mathcal{S} \cap \mathcal{Z}}$ can be greatly facilitated by Theorem~\ref{thm:dwell-time}. Consider the reduced-order switching system 
\begin{equation}\label{eq:switch-system-zd}
z_{k+1} = P_{\sigma(k)}|_{\mathcal{S} \cap \mathcal{Z}}(z_k)
\end{equation}
and apply Theorem~\ref{thm:dwell-time} to obtain an explicit characterization of the set $\Omega \subset \mathcal{S} \cap \mathcal{Z}$ within which the state of \eqref{eq:switch-system-zd} is trapped; this is achieved by the procedure detailed in Appendix~\ref{app:sec:implementation}. Given $\Omega$, introduce an appropriately dense mesh
\begin{equation}\label{eq:mesh-set}
\mathcal{B}_W(\bar{z}^*)=\{z\in\mathcal{S}\cap\mathcal{Z}~:~\|z-\bar{z}^*\|_\infty \leq W \}
\end{equation}
where $\bar{z}^*$ be the centroid of the fixed points $z_p^*$ and $W>0$ is chosen so that $\mathcal{B}_W$ covers the (bounded) set $\Omega$. Then, using least square minimization, polynomial functions
\begin{equation}\label{eq:polys}
\tilde{P}_p|_{\mathcal{S} \cap \mathcal{Z}}(z) = \sum^{M_P}_{j=0} a^P_{p,j} z^j ~~\text{and}~~
\tilde{H}_p|_{\mathcal{S} \cap \mathcal{Z}}(z) = \sum^{M_H}_{j=0} a^H_{p,j} z^j
\end{equation}
can be fitted to the values $P_p|_{\mathcal{S} \cap \mathcal{Z}}(z)$ and $H_p|_{\mathcal{S} \cap \mathcal{Z}}(z)$ at each node of the mesh, resulting in the approximate actions \eqref{eq:approximate-reduced-actions}. It should be emphasized here that the use of approximate reduced-order actions does \emph{not} affect the stability properties of \eqref{eq:switch-system-zd}; it only affects planning accuracy by introducing a small error; see the example below and Fig.~\ref{fig:actions}. 

\subsection{Example: Constructing a library of planning actions}
\label{subsec:example-actions}

For concreteness, we consider a gait library $\mathbb{G}$ composed by three primitives $\mathbb{G} = \{\mathcal{G}_1, \mathcal{G}_2, \mathcal{G}_3\}$, corresponding to straight-line walking and turning at $-45^\circ$ and $+45^\circ$, respectively. The design of the underlying controllers $\{\Gamma_1, \Gamma_2, \Gamma_3\}$ follows the procedure described above and in Appendix~\ref{app:HZD-controller}, and creates a 2-dimensional common intersection surface $\mathcal{S} \cap \mathcal{Z}$ on which the gait primitives in $\mathbb{G}$ can be represented in a reduced-order form $\mathcal{G}_p|_{\mathcal{S} \cap \mathcal{Z}} = \{P_p|_{\mathcal{S} \cap \mathcal{Z}},~z^*_p\}$, $p \in \{1,2,3\}$. This way, a collection of (exact) reduced-order planning actions \eqref{eq:reduced-actions} is extracted. Figure~\ref{fig:gaits} presents the limit cycles associated with the gaits above and Fig.~\ref{fig:actions} shows the corresponding displacements.

To compute approximate reduced-order planning actions \eqref{eq:approximate-reduced-actions} from the actions computed above, we first apply Theorem~\ref{thm:dwell-time} on the corresponding reduced-order switching system \eqref{eq:switch-system-zd}. To do this, we follow the procedure outlined in Appendix~\ref{app:sec:implementation} with the translated reduced-order dynamics $\bar{z}_{k+1} = \rho_p|_{\mathcal{S} \cap \mathcal{Z}}(\bar{z}_k)$, where $\bar{z} = z - z^*_p$ and $\rho_p|_{\mathcal{S} \cap \mathcal{Z}}(\bar{z}) = P_p|_{\mathcal{S} \cap \mathcal{Z}}(\bar{z} + z^*_p)-z^*_p$. Applying \ac{SOS} programming, we find that implication \eqref{eq:BoA-condition-approx} in Appendix~\ref{app:sec:implementation} is satisfied with $\lambda = \lambda_1=\lambda_2=\lambda_3=0.12$ and $\overline{\kappa}_1 = 0.11$, $\overline{\kappa}_2=0.15$ and $\overline{\kappa}_3=0.08$. Next, selecting $\kappa=0.0002$ and applying \eqref{eq:omega-bound}-\eqref{eq:mu-bound} in Appendix~\ref{app:sec:implementation} we obtain $\omega(\kappa)=0.0016$ and $\mu(\kappa)=8.08$, resulting in $N_\mathrm{d}=1$ and in the sets $\mathcal{M}_p(\omega(\kappa))$, $p \in \{1,2,3\}$ shown in Fig.~\ref{fig:mesh}. It is clear that the inclusion \eqref{eq:feasible-0-inp-fixed} is satisfied. Hence, Theorem~\ref{thm:dwell-time}\emph{(i)} implies that the gait switching system \eqref{eq:switch-system-zd} is practically stable with respect to the sets $\Omega_0$ and $\Omega$, which are explicitly computed through the intersection and union of the sets $\mathcal{M}_p(\omega(\kappa))$, $p \in \{1,2,3\}$, respectively. Given the trapping set $\Omega$, we approximate the maps $P_p|_{\mathcal{S} \cap \mathcal{Z}}$ and $H_p|_{\mathcal{S} \cap \mathcal{Z}}$, by fitting polynomials \eqref{eq:polys} on an evenly spaced mesh of $400$ nodes over the set \eqref{eq:mesh-set} with $W=0.05$; see Fig.~\ref{fig:mesh}. As can be seen from the solid triangles in Fig.~\ref{fig:actions}, the predictions of the approximate displacement maps are almost indistinguishable from the actual ones.
More details regarding the polynomial approximations \eqref{eq:polys} in this example can be found in Appendix~\ref{app:Polynomial_approximations}; see also~\cite{videolink} for a video of the resulting motions.

\section{Dynamic Walking in Changing Environments}
\label{sec:path-planning}

This section describes a primitive-based sequential \ac{MPC} approach for online planning with limit-cycle walking motions in changing environments. 
The proposed \ac{MPC} scheme enables the robot to compute in real time a suitable primitive sequence in the presence of static and moving obstacles; the motion of the latter is assumed to be known. The sequence of primitives suggested by the \ac{MPC} combines locomotion stability---in the sense of practically stable state evolution of the gait switching system as per Theorem~\ref{thm:dwell-time}---with collision-free paths.

\subsection{Practically stable primitive-based predictive planning}
\label{subsec:MPC}

We consider a class of reach-avoid navigation problems, wherein a dynamic walker is tasked with reaching a desired (goal) location in an obstacle-cluttered workspace $\mathcal{W}$. We will be concerned with both static and moving obstacles. Let $\mathcal{W}^\mathrm{s} \subset \mathcal{W}$ denote the static part of $\mathcal{W}$; moving obstacles are not included in $\mathcal{W}^\mathrm{s}$. Instead of treating each obstacle in $\mathcal{W}^\mathrm{s}$ separately, we work with the free space $\mathcal{W} \setminus \mathcal{W}^\mathrm{s}$ and extract a \ac{SWC} that contains the initial and goal locations and lies entirely in $\mathcal{W} \setminus \mathcal{W}^\mathrm{s}$. This allows us to focus on the free space in the vicinity of the robot, decreasing the number of constraints required for collision checking and reducing computation time~\cite{Nark2022RAL}. The resulting \ac{SWC} is then used to provide constraints to a sequence of \emph{primitive-based} \ac{MPC} programs, which compose the actions in $\tilde{\mathbb{A}}$ to drive the robot to the goal while keeping its stride-to-stride evolution bounded within the \ac{SWC}. Finally, moving obstacles are included through additional constraints, thus ensuring reactive composition without re-planning the corridor.

\subsubsection{Extracting safe walking corridors}
Given $\mathcal{W}^\mathrm{s}$ and initial and goal locations $\mathrm {p}_{\rm in}, \mathrm{p}_{\rm g} \in \mathcal{W} \setminus \mathcal{W}^\mathrm{s}$, a \ac{SWC} is constructed by computing an ordered collection of pairwise intersecting polytopes in $\mathcal{W} \setminus \mathcal{W}^\mathrm{s}$; the union of these polytopes provides a \ac{SWC}. Assuming that the obstacles in $\mathcal{W}^\mathrm{s}$ are convex, this computation can be performed efficiently by the method in~\cite{Nark2022RAL}. The end result is a sequence of polytopes $\mathcal{H}_i \subset \mathcal{W} \setminus \mathcal{W}^\mathrm{s}$, $i=1,...,M$, with $\mathrm{p}_{\rm in} \in \mathcal{H}_1$, $\mathrm{p}_{\rm g} \in \mathcal{H}_{M}$, and a collection of waypoints ${\rm w}_{i}$ so that
\begin{align} \nonumber
    {\rm w}_i &\in \mathcal{H}_{i} \bigcap \mathcal{H}_{i+1} ~~\text{for}~~ i=1,..., M-1 \enspace. 
\end{align}
Details can be found in~\cite{Nark2022RAL}; we only mention here that the method's computational time  scales well with the number of obstacles and that the resulting \ac{SWC}s capture more free space in $\mathcal{W} \setminus \mathcal{W}^\mathrm{s}$ than other similar methods. 

\subsubsection{Safe action composition via sequential MPC}
Here, the term ``safe'' indicates practically stable state evolution of the underlying reduced-order gait switching system \eqref{eq:switch-system-zd} combined with collision-free robot displacements.
Hitherto, a \ac{SWC} $\{ \mathcal{H}_{i}~|~i = 1,..., M \}$ and a collection of waypoints $\{ \mathrm{w}_i~|~i = 1,...,M-1 \}$ computed as above are assumed to be given. It is also assumed that a library $\mathbb{A}$ of reduced-order actions that leads to a practically stable \ac{SSME} \eqref{eq:switch-system-zd} with respect to sets $\Omega_0$ and $\Omega$ is available; these actions can be computed as in Section~\ref{subsec:example-actions} using Theorem~\ref{thm:dwell-time}. To simplify the formulation, we will assume that practical stability is ensured with dwell time constraint $\overline{N}_\mathrm{d}=1$. Having this condition is advantageous as it offers the highest degree of flexibility when combining actions, allowing switching to a new gait at each stride without jeopardizing locomotion stability.

\begin{figure*}[t!]
\vskip +5pt 
\centering
\includegraphics[width =0.95\linewidth]{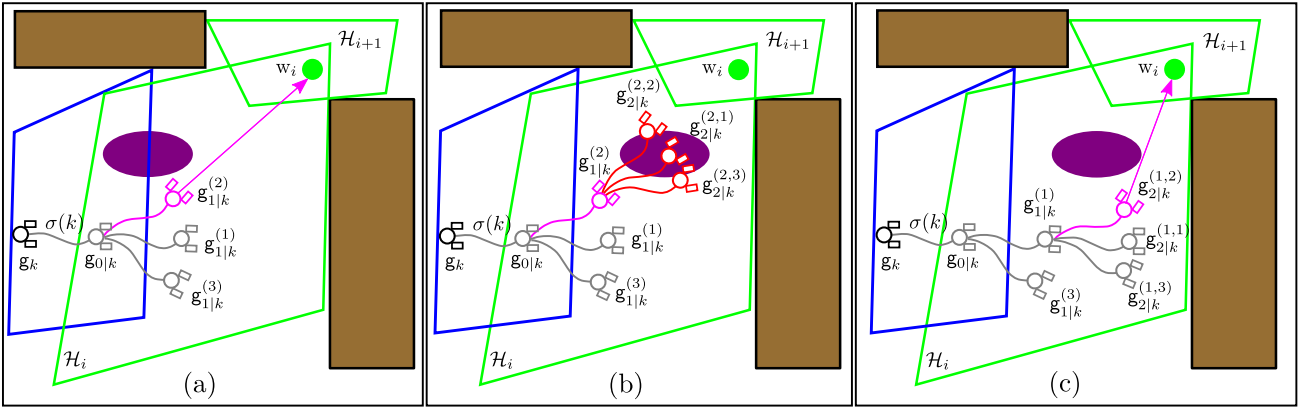}
\caption{A schematic representation of the inner workings of \ac{MPC}$(i)$ in the presence of a moving obstacle (ellipse in magenta) for horizon $N=2$. 
(a) The \ac{MPC} begins by predicting the pose $\mathsf{g}_{0|k}$ at stride $k+1$ using feedback of the pose $\mathsf{g}_k$ at the beginning of the current stride $k$ and the primitive index $\sigma(k)$ that is engaged at that stride and is available from the previous solution step. Since $\mathsf{g}_{0|k} \in \mathcal{H}_i$, the polytopes $\mathcal{H}_i$ and $\mathcal{H}_{i+1}$ of the \ac{SWC} are activated (green). The algorithm expands the primitive tree by applying the available actions and ensuring that its predictions are within $\mathcal{H}_i \cup \mathcal{H}_{i+1}$. It then selects the node obtained by primitive $\{2\}$ (magenta), the pose $\mathsf{g}^{(2)}_{1|k}$ of which is closest to the waypoint $\mathrm{w}_i$.
(b) Expanding $\mathsf{g}^{(2)}_{1|k}$ leads to predictions which collide with the moving obstacle. 
(c) All the nodes that lead to collisions together with their parent node are pruned. The algorithm backtracks to the previous stage and selects the node $\mathsf{g}^{(1)}_{1|k}$, which is the second closest to $\mathrm{w}_i$, for further expansion. The process continues until the intersection $\mathcal{H}_i \cap \mathcal{H}_{i+1}$ is reached. 
} 
\label{fig:mpc_algorithm}
\vskip -5pt
\end{figure*}

Equipped with a \ac{SWC} and a library of actions $\mathbb{A}$, we now proceed with formulating a corresponding sequence $\{\text{MPC}(i)~|~i=1,...,M\}$ of primitive-based \ac{MPC} programs that safely drives the robot to the goal. We begin by extracting a collection of approximate reduced-order actions $\tilde{\mathbb{A}}$ from $\mathbb{A}$ offline as in Section~\ref{subsec:example-actions}. 
The \emph{predictive} nature of the actions in $\tilde{\mathbb{A}}$ is then exploited for online planning as follows.
Given the robot's pose $\mathsf{g}_k = [\mathrm{p}^\mathsf{T}_k ~ \psi_k]^\mathsf{T}$, (reduced) state $z_k$, and primitive index $\sigma(k)$ at the beginning of the $k$-th stride, applying the corresponding action $\mathcal{A}_{\sigma(k)}$ provides a \emph{prediction} of the pose and state of the robot at the beginning of the $(k+1)$ stride. This prediction is taken as initial condition for the \ac{MPC}, and is denoted by $\mathsf{g}_{0|k}$, and $z_{0|k}$; see Fig.~\ref{fig:mpc_algorithm}. Subsequent application of actions in $\tilde{\mathbb{A}}$ results in the corresponding predictions $\mathsf{g}_{\ell|k}=[\mathrm{p}^\mathsf{T}_{\ell|k} ~ \psi_{\ell|k}]^\mathsf{T}$ and $z_{\ell|k}$ for the $\ell=1,...,N$ strides that constitute the horizon $N$ of the \ac{MPC}. 

With this notation, the objective of \ac{MPC}$(i)$ is to minimize the distance from the $i$-th waypoint $\mathrm{w}_i$ while keeping the predicted robot positions $\mathrm{p}_{\ell|k}$ within $\mathcal{H}_i$. To take moving obstacles into account, we assume that the area occupied by the $\nu$-th moving obstacle can be captured by an ellipse, the position $\mathrm{p}_{\nu,\ell|k}^{\rm obs}$ and orientation $\mathsf{E}_{\nu,\ell|k}$ of which are known over the horizon\footnote{It is assumed that the motion of the moving obstacles is known or can be predicted using a Kalman filter~\cite{Mora_2019_RAL}. Only moving obstacles that are within a distance of $5\mathrm{m}$ from the robot's predicted position are taken into account.} Then, for each moving obstacle $\nu=1,...,n_\mathrm{d}$, the following constraint is incorporated in the \ac{MPC} programs 
\begin{equation}\nonumber
    F_{\nu,\ell|k} = (\mathrm{p}_{\ell|k} - \mathrm{p}_{\nu,\ell|k}^{\mathrm{obs}})^\mathsf{T}\mathsf{E}_{\nu,\ell|k}(\mathrm{p}_{\ell|k} - \mathrm{p}_{\nu,\ell|k}^{\mathrm{obs}}) > 1 \enspace.
\end{equation}

To summarize, we have the following sequence of \ac{MPC} programs that drives the robot to the last polytope $\mathcal{H}_M$: 
\vskip -5pt
\noindent\rule{\columnwidth}{0.5pt}
\ac{MPC}$(i)$, $i = 1, 2,..., M-1$: 
\vskip -5pt
\noindent\rule{\columnwidth}{0.5pt}
\vskip -12pt
\begin{subequations}\label{eq:optim_detail}
\begin{IEEEeqnarray}{s'rCl'rCl'rCl}
\label{eq:MPC_index}
$\underset{ \{p_0, ..., p_{N-1} \} \in \mathcal{P}}{\text{minimize}}$ & \IEEEeqnarraymulticol{9}{l} {\displaystyle \sum^{N}_{\ell = 1} \| {\rm w}_{i} - \mathrm{p}_{\ell|k} \|^2}
\\ 
\label{eq:MPC_initial_pose}
subject to & \mathsf{g}_{0|k} &=& \mathsf{g}_k + \tilde{H}_{\sigma(k)}|_{\mathcal{S} \cap \mathcal{Z}}(z_k)
\\
\label{eq:MPC_initial_state}
& z_{0|k} &=& \tilde{P}_{\sigma(k)}|_{\mathcal{S} \cap \mathcal{Z}}(z_k)
\\
\label{eq:MPC_prediction_pose}
& \mathsf{g}_{\ell+1|k} &=& \mathsf{g}_{\ell|k} + \tilde{H}_{p_\ell}|_{\mathcal{S} \cap \mathcal{Z}}(z_{\ell|k}) 
\\
\label{eq:MPC_prediction_state} 
& z_{\ell+1|k} &=& \tilde{P}_{p_\ell}|_{\mathcal{S} \cap \mathcal{Z}}(z_{\ell|k}) 
\\
\label{eq:MPC_obst_static}
& \mathrm{p}_{\ell|k} &\in& \mathcal{H}_{i}\bigcup\mathcal{H}_{i+1} 
\\
\label{eq:MPC_obst_dynamic}
& F_{\nu,\ell|k} & > & 1, ~~ \nu=1,...,n_{\rm d} \enspace 
\end{IEEEeqnarray}
\end{subequations}
\vskip -10pt
\noindent\rule{\columnwidth}{0.5pt}
where we note that the constraint \eqref{eq:MPC_obst_static} was added in place of the constraint $\mathrm{p}_{\ell|k} \in \mathcal{H}_i$ to facilitate transitions between subsequent polytopes when their intersection is small. Once the robot is in $\mathcal{H}_M$, a slightly modified \ac{MPC}$(M)$ is engaged to drive it to the goal; in \ac{MPC}$(M)$, the waypoint in \eqref{eq:MPC_index} is replaced with the goal location $\mathrm{p}_\mathrm{g}$ and the constraint \eqref{eq:MPC_obst_static} is replaced by $\mathrm{p}_{\ell|k} \in \mathcal{H}_{M-1} \bigcup \mathcal{H}_{M}$. Owing to the polynomial nature of the approximate actions, the predictions \eqref{eq:MPC_prediction_pose}-\eqref{eq:MPC_prediction_state} are computed very fast. Solving MPC$(i)$ returns the sequence of indices $\{p^*_0,...,p^*_{N-1} \}$ that minimizes \eqref{eq:MPC_index}. We then set $\sigma(k+1)=p^*_0$ and proceed with solving MPC$(i)$ at the next stride based on the new initial condition. The process is repeated  until $\mathrm{p}_k \in \mathcal{H}_{i+1}$ for some $k$, where we switch to MPC$(i+1)$ and continue until the goal. 

The \ac{MPC} programs in \eqref{eq:optim_detail} rely on the approximate reduced-order planning actions, $\tilde{\mathbb{A}}$, rather than the exact ones, $\mathbb{A}$. While this choice impacts prediction accuracy to some extent,\footnote{As discussed in~\cite{Veer2017IROS}, its impact is minimal when compared with the predictions by the exact reduced-order actions $\mathbb{A}$; see also Fig.~\ref{fig:actions}.} it does not compromise the practical stability properties of the underlying reduced-order switching system \eqref{eq:switch-system-zd}. This is because the approximate maps derived in \eqref{eq:polys} and used in \eqref{eq:optim_detail} to make the requisite predictions do not excite dynamics outside $\mathcal{S} \cap \mathcal{Z}$. Hence, if $\overline{N}_\mathrm{d}=1$ as we assume in this section, the \ac{MPC} programs \eqref{eq:optim_detail} have the flexibility to decide to switch to a new primitive at every stride, while the evolution of the system \eqref{eq:switch-system-zd} remains practically stable with respect to the sets $\Omega_0$ and $\Omega$ associated with the exact action library $\mathbb{A}$. If, on the other hand, $\overline{N}_\mathrm{d}>1$ or an average dwell time constraint is needed, the formulation of the \ac{MPC} \eqref{eq:optim_detail} should be modified to incorporate the corresponding constraint. We will refrain from delving deeper into this matter here as it would introduce additional complexity without it being necessary for the examples considered.

Finally, due to the discrete nature of the actions, solving each \ac{MPC} corresponds to a tree search problem. In \eqref{eq:optim_detail}, at each stage of the \ac{MPC} all nodes of the tree are expanded using the available primitives; when all nodes are feasible, this expansion results in exponential time complexity $|\mathcal{P}|^N$, where $|\mathcal{P}|$ is the number of gait primitives. To avoid the rapidly increasing computational cost, we will instead employ a modified best first search algorithm to construct the tree online and solve for obstacle-free primitive sequences. The algorithm begins with expanding the primitive tree from $\mathsf{g}_{0|k}$ using the available actions $\tilde{\mathbb{A}}$, as shown in Fig.~\ref{fig:mpc_algorithm}. The expanded nodes that lead to collisions are pruned, and out of the remaining (feasible) ones, the best node---that is, the one closest to the waypoint of \ac{MPC}$(i)$ (or to the goal for \ac{MPC}$(M)$)---is selected for further expansion of the tree. This way only one node is expanded at each stage, resulting in linear time complexity $|\mathcal{P}| N$ (if all nodes are feasible) at the cost of finding a suboptimal primitive sequence. Note that if at some expansion stage, all nodes lead to collisions, the algorithm backtracks to the previous stage, prunes the best node and selects the second best for expansion. The process is repeated until the algorithm finds a feasible primitive sequence, from which the first one is executed.

\begin{figure*}[b!]
\centering
\vskip +5pt
\subfigure[]{\includegraphics[width=0.30\textwidth]{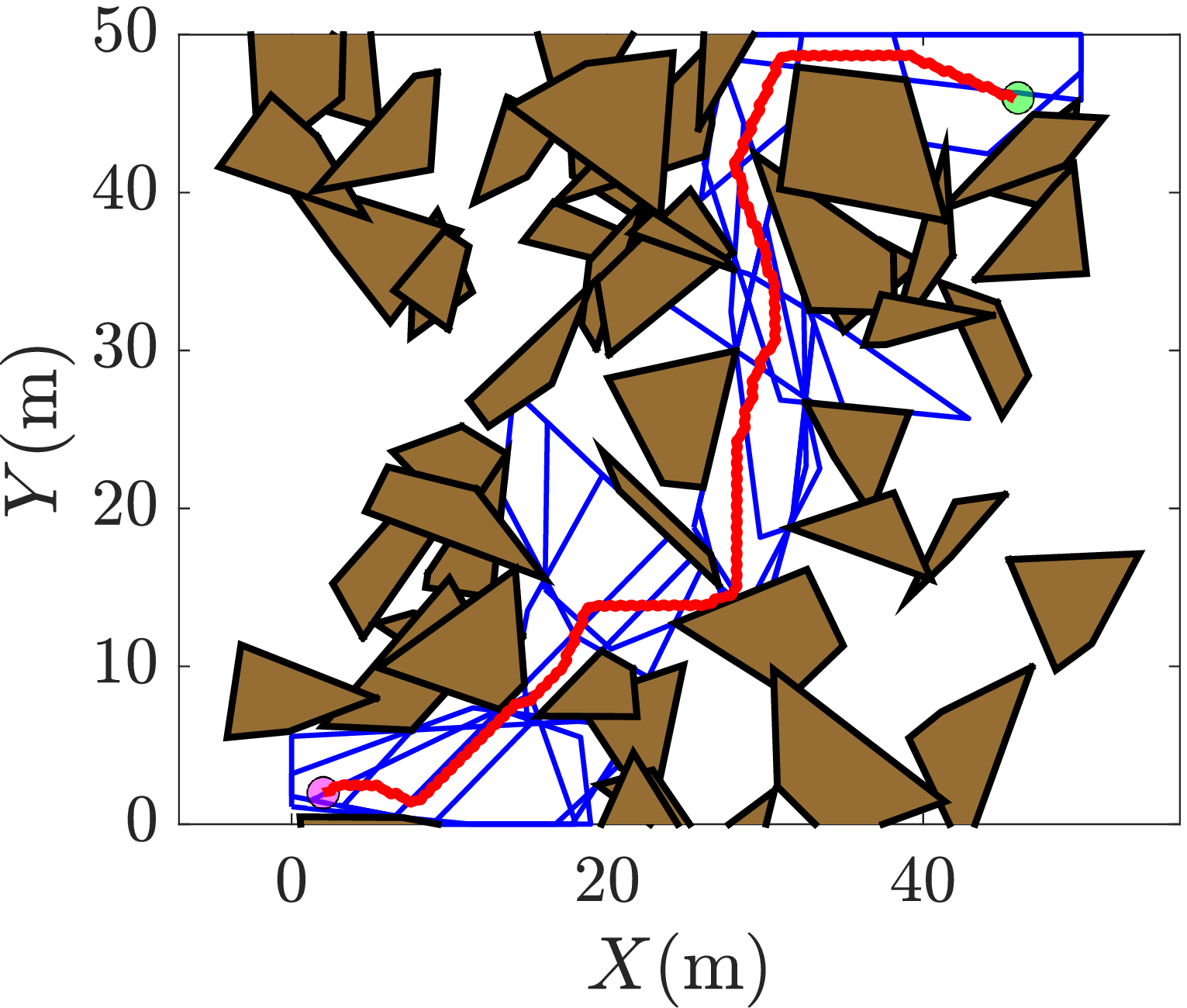} \label{fig:MPC_polygonal_osbtacles}}\quad
\subfigure[]{\includegraphics[width=0.32\textwidth]{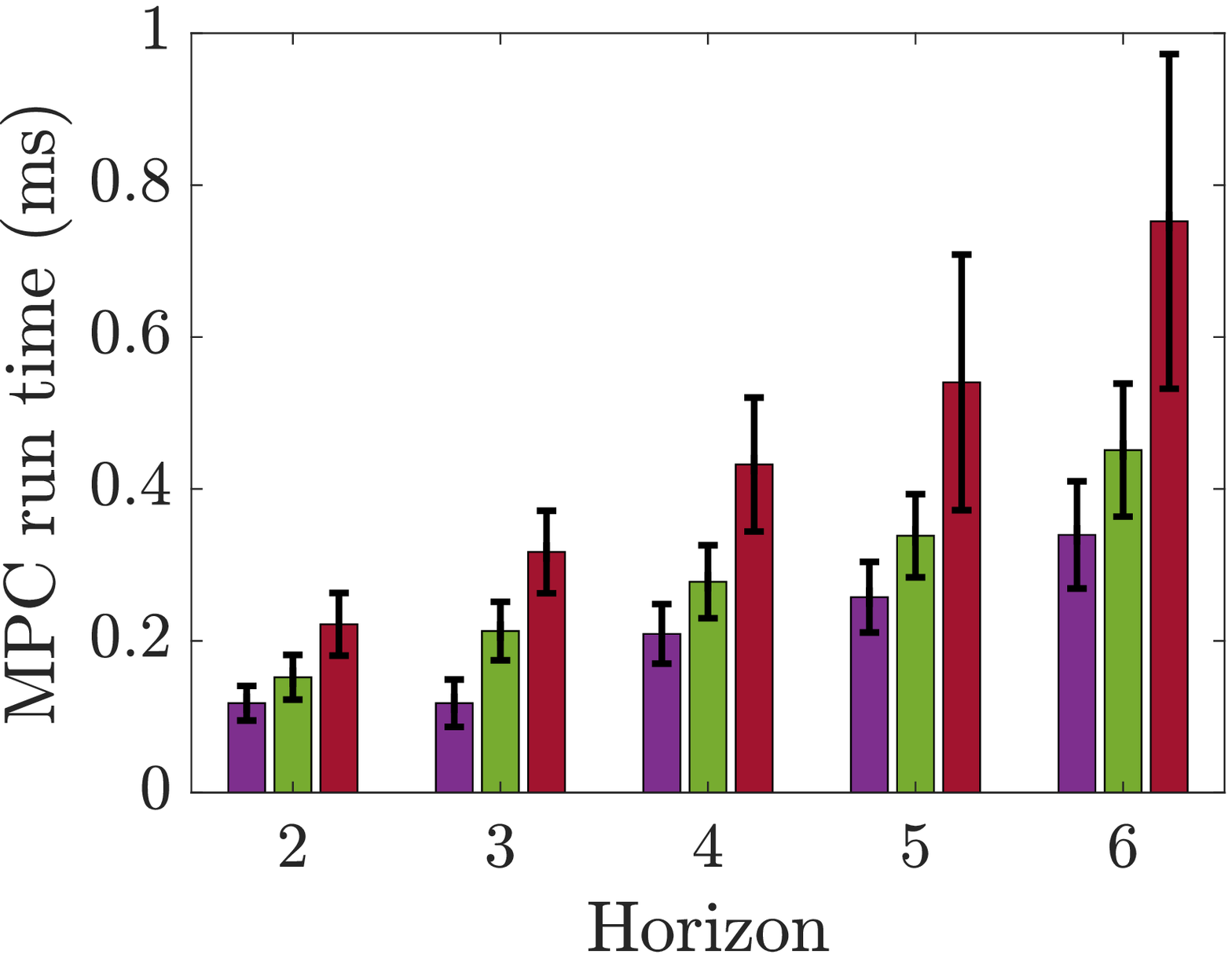} \label{fig:MPC_runtime}}\quad
\subfigure[]{\includegraphics[width=0.32\textwidth]{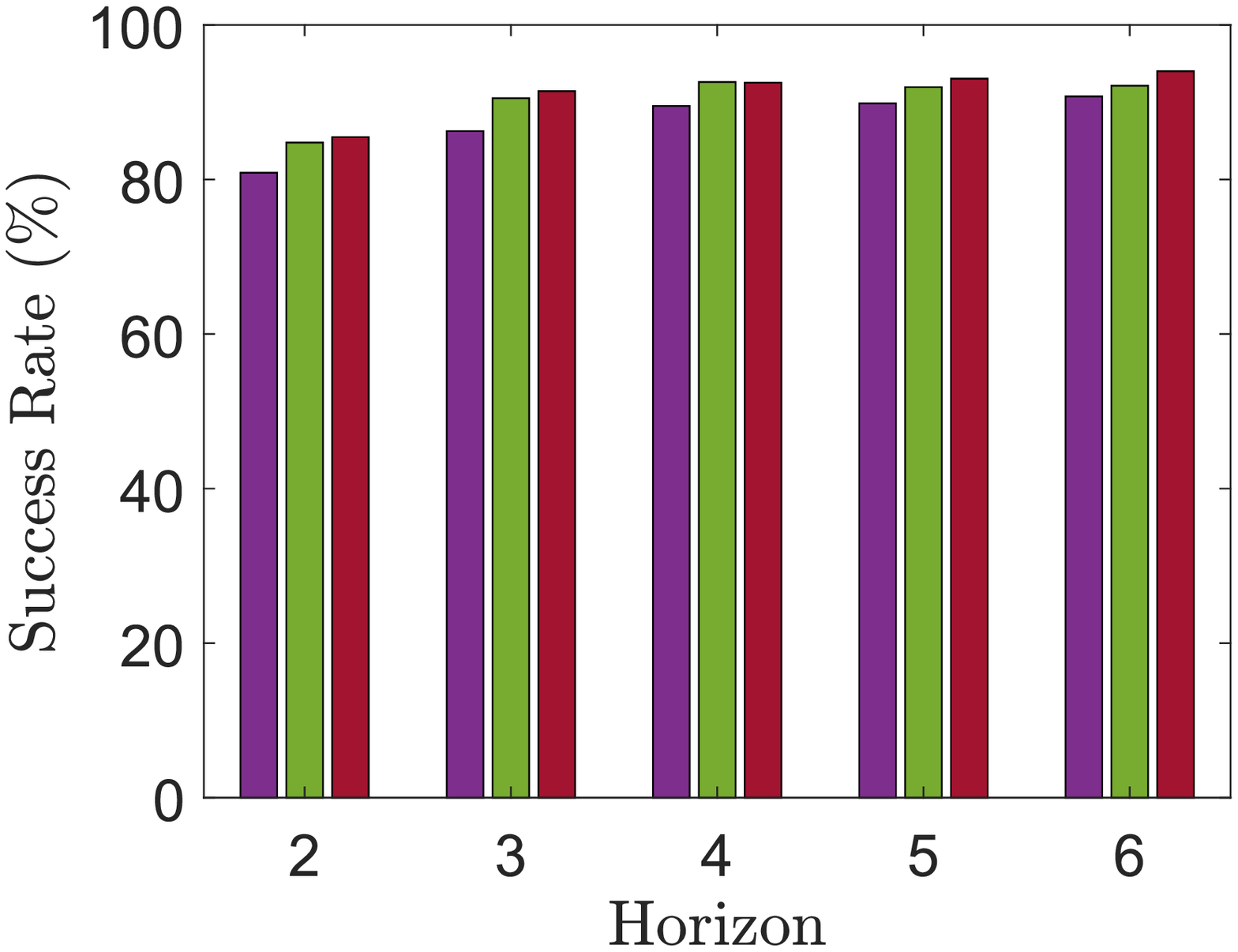} \label{fig:MPC_success_rates}}
\vskip -8pt
\caption{ (a) A sample environment with polygonal obstacles (brown) together with a \ac{SWC} (blue polytopes) and a suggested \ac{COM} path (red). (b) Average \ac{MPC} run times (we do not include the computation of the \ac{SWC}s, which are reported in~\cite{Nark2022RAL}) per iteration for horizons $N = 2,3,4,5,6$ with $3$ (purple), $5$ (green), $9$ (red) primitives. (b) Corresponding success rates.}
\label{fig:MPC_computations}
\vskip -10pt
\end{figure*}

\subsection{Performance Evaluation}
\label{subsec:MPC-simulations}

We evaluate the performance of the proposed MPC scheme in simulations with the 3D bipedal robot model of Section~\ref{sec:tractability}. The dynamics of the robot was integrated in MATLAB using $\mathtt{ode45}$ with event detection to detect swing leg touchdown for domain switching. All simulations were performed on an Intel PC with $\mathrm{i}7{\text -}9750\mathrm{H}$ processor ($2.60$ GHz) and $16\mathrm{GB}$ RAM.

\subsubsection{Algorithm performance}
To assess the efficacy of the primitive-based sequential \ac{MPC} algorithm described above, we provide numerical evidence on run times and success rates over a large number of randomly generated, highly cluttered environments. We generate these environment by randomly placing rectangular obstacles of varying size and orientation as well as general polytopic obstacles in a confined $50 {\rm m} \times 50 {\rm m}$ space keeping the position of the initial and goal point fixed. We consider the cases of 30, 40, 50, and 60 obstacles with the occupancy ratio kept at approximately 40\%; see Fig.~\ref{fig:MPC_polygonal_osbtacles} for an example environment with polytopic obstacles. This way, a total of 600 environments are constructed by generating 150 environments for each obstacle population. For each environment, we use our previous work~\cite{Nark2022RAL} to generate the \ac{SWC} connecting the starting and goal locations, which are kept the same across all the environments. 

The sequential primitive-based \ac{MPC} algorithm is then applied to generate collision-free primitive sequences for each environment. The algorithm runs until the robot is within a disc of radius $1\mathrm{m}$ around the goal. To evaluate the effect of the number of primitives $|\mathcal{P}|$ used on the success rates and computational time of the algorithm, we consider gait libraries with $3$, $5$, and $9$ primitives; these correspond to straight walking ($0^\circ$ change in orientation) and turning gaits with nominal orientation changes $\{\pm 45^\circ\}$ for the $3$ primitives, $\{\pm 45^\circ, \pm 30^\circ \}$ for the $5$ and $\{ \pm 45^\circ, \pm 30^\circ, \pm 20^\circ, \pm 10^\circ \}$ for the $9$ primitives. All primitives are computed as in Section~\ref{subsec:example-actions} so that practically stable switching with respect to sets $\Omega_0$ and $\Omega$ is guaranteed with dwell time bound $\overline{N}_{\rm d}=1$; the sets $\Omega_0$ and $\Omega$ are similar to those in Fig.~\ref{fig:mesh}.  

Our results over the 600 environments are summarized in Fig.~\ref{fig:MPC_computations}. It can be seen from Fig.~\ref{fig:MPC_runtime} that for the same number of primitives, the computational time grows almost linearly with the horizon of the \ac{MPC}. A similar observation can be made when the horizon is kept constant and the number of primitives increases. This is consistent with the linear \emph{lower} bound $|\mathcal{P}| N$ on the complexity of the best first search, indicating that the algorithm did not have to resort to frequent backtracking. Note that in all cases, the \ac{MPC} run times remain below $1 {\rm ms}$, which reflects the benefit of using discrete search over a finite library of primitives computed offline. Finally, Fig.~\ref{fig:MPC_success_rates} shows the corresponding success rates, which, for horizons larger than $N=2$, exceed $90\%$. 

\begin{figure*}[t]
\centering
\includegraphics[width=0.97\textwidth]{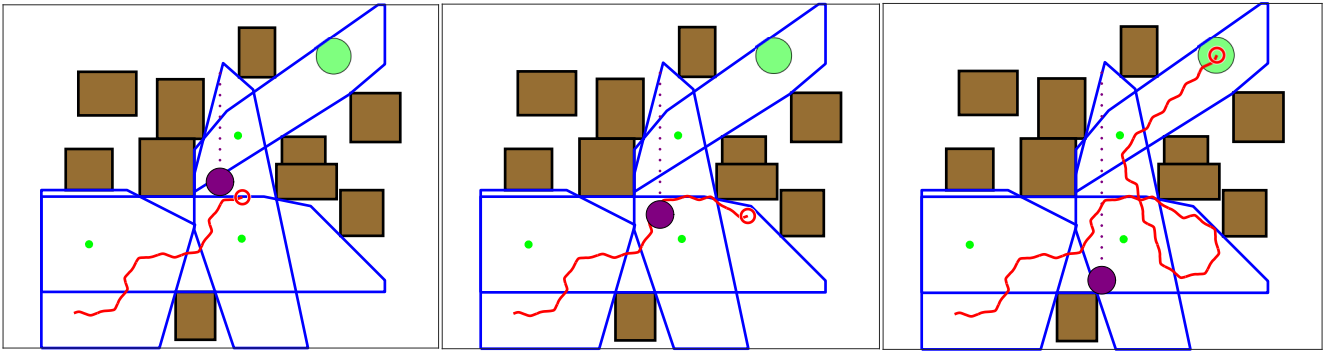} \label{fig:dyn_obs}
\caption{Snapshots of a walking robot (red circle with radius $0.2\mathrm{m}$) avoiding a moving obstacle (purple circle) in a space cluttered with static obstacles (brown rectangles). The \ac{SWC} is a sequence of pairwise intersecting polytopes (blue) with waypoints (green dots). See~\cite{videolink} for a video of the resulting motions. }
\label{fig:mpc}
\end{figure*}

\begin{figure}[t!]
\centering
\vskip -5pt
\subfigure[]{\includegraphics[width=0.23\textwidth]{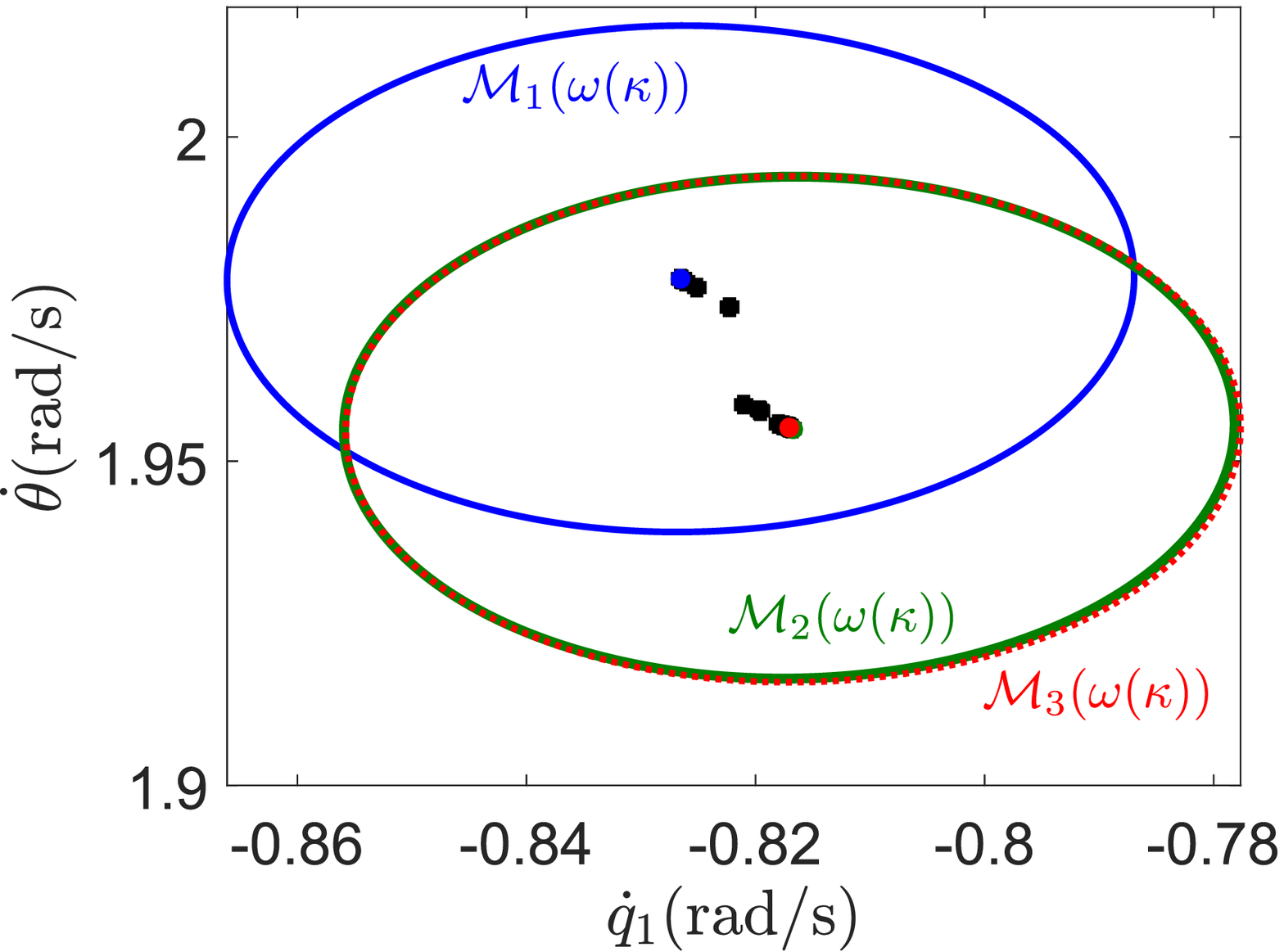} \label{fig:zero_dynamics_evolution}
}
\subfigure[]{\includegraphics[width=0.23\textwidth]{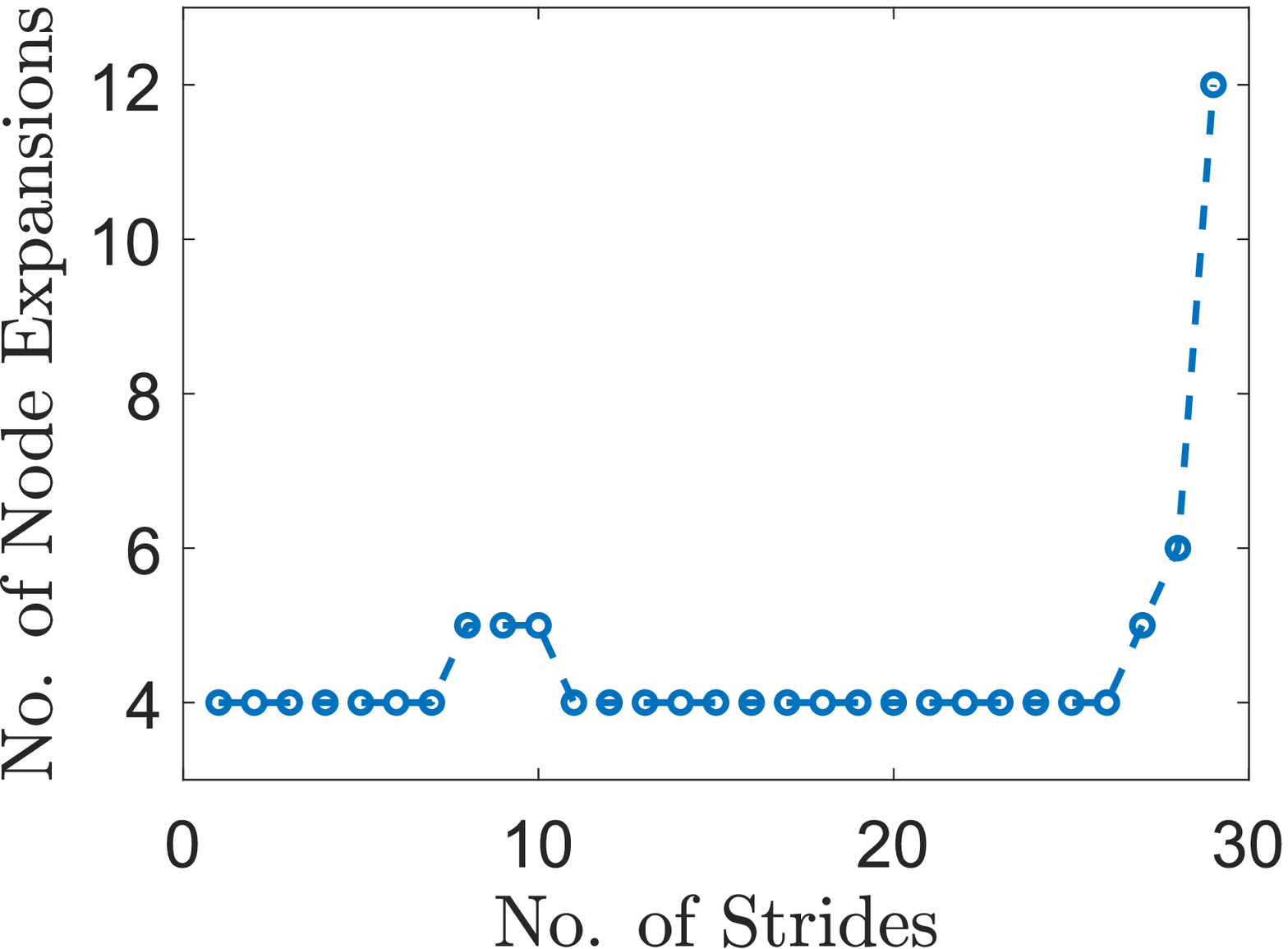} \label{fig:node_exp}
}
\caption{(a) Evolution of the gait switching system \eqref{eq:switch-system-zd} (black dots) within the set $\Omega$ (union of the ellipses). (b) Node expansions per stride.}
\label{fig:node_expansion_zero_dynamics}
\vskip -10pt
\end{figure}

\subsubsection{Moving obstacle avoidance}
This example illustrates how moving obstacles are treated by the sequential \ac{MPC}. In what follows, we set the \ac{MPC} horizon at $N=4$ and use a gait library composed of $|\mathcal{P}|=3$ primitives, which are the same as the ones computed in Section~\ref{subsec:example-actions}. Figure~\ref{fig:mpc} depicts a $10\mathrm{m} \times 10\mathrm{m}$ workspace with $10$ static obstacles, along with a \ac{SWC} of sequentially overlapping polytopes and the corresponding waypoints. A circular obstacle of radius $0.4\mathrm{m}$ is moving in the free space along a straight line with a constant speed $0.32\mathrm{m/s}$ and its motion is assumed to be known. As shown in Fig.~\ref{fig:mpc}, the robot (represented as a red circle) is able to avoid the moving obstacle while staying within the \ac{SWC}, owing to the added constraint \eqref{eq:MPC_obst_dynamic}. Note that high-level online planning is combined with practically-stable low-level evolution of the gait switching system \eqref{eq:switch-system-zd}, as predicted by Theorem~\ref{thm:dwell-time} and indicated by Fig.~\ref{fig:zero_dynamics_evolution}.

Finally, it is of interest to discuss here how the number of node expansions required by the \ac{MPC} is affected by the moving obstacle and by the proximity of the goal point to the boundary of the polytope that contains it. Note first that the \ac{MPC} is solved once per stride. As shown in Fig.~\ref{fig:node_exp}, when the moving obstacle is far from the robot, the total number of node expansions is $4$, which is equal to the horizon $N=4$. This indicates that only one node is expanded per stage of the \ac{MPC}, as expected by the best first search algorithm. However, when the obstacle is in the vicinity of the robot, we observe a slight increase in the number of node expansions, implying that the algorithm utilizes backtracking to suggest a suitable primitive. 
It can also be seen in Fig.~\ref{fig:node_exp} that the node expansion increases abruptly when the robot approaches the goal point. This is because when the robot approaches the goal, the nodes in the primitive tree that are closest to the goal extend outside the polytope, particularly if the goal is close to the polytope boundary. As a result, these nodes are pruned and the algorithm backtracks more frequently to compute a viable primitive sequence. This shortcoming is related to the design of the gait library, which, in this example, uses just three walking gaits with similar stride lengths. This issue can be alleviated easily by enriching the library to include gaits with smaller stride lengths for a more refined reachability tree.

\section{Conclusion}
\label{sec:conclusion}

This paper presented a rigorous approach towards stable gait composition for reactive planning with dynamic (limit-cycle) walking motions amidst static and moving obstacles. Our approach relies on extracting a library of \ac{DMP}s in the form of stride maps and their fixed points. These \ac{DMP}s capture the local dynamics of walking around exponentially stable limit-cycle gaits at the stride level, and the corresponding displacements provide actions available for planning. We identified a general class of 3D bipedal models that enable the decomposition of the problem to  high-level planning and low-level gait generation modules. To capture stability aspects, we formulated gait composition as a \ac{SSME}, and we offered provable stability guarantees under explicit constraints on the frequency of switching. Finally, we proposed a primitive-based sequential \ac{MPC} to safely drive the robot to the goal while bounding its stride evolution within a \ac{SWC} and reacting to moving obstacles. The end result is a method that combines collision-free navigation with practically stable locomotion guarantees. 

The paper also addressed various implementation aspects towards reducing the computational cost of the method. At the core of our approach lies the idea of dimensional reduction through the enforcement of suitably designed (virtual) holonomic constraints using feedback. This allows the restriction of the planning system on lower-dimensional surfaces and drastically simplifies computations. However, this reduction comes at the cost of guaranteeing stability for the emerging reduced-order planning system. Extending these guarantees to the full-order system requires careful consideration of the effect of unexpected disturbances, model mismatch, and noisy measurements, which violate the assumptions associated with the reduction and will be present in experimental implementations of the method. Addressing these challenges is at the focus of our current work, which examines practical stability in the context of perturbed switched systems; a preliminary result in that direction can be found in~\cite[Section 6.1.4]{Veer2018PhDThesis}. Finally, it is worth mentioning that the tools developed in this paper are relevant to other robots which---like legged robots---move through their environment via periodic interactions. 
 
\begin{appendices}

\section{Proof of Propositions \ref{prop:symmetry} and~\ref{prop:symmetry_P}}
\label{app:sec:model}

The properties listed in Propositions~\ref{prop:symmetry} and~\ref{prop:symmetry_P} are consequences of the symmetries discovered in~\cite{spong2005controlled} with the modification of including the position of the robot in $\{ \mathcal{I} \}$. 

\subsection{Proof of Proposition~\ref{prop:symmetry}}

For part~(\ref{P_f}), it was observed in~\cite{spong2005controlled} that the kinetic energy $\mathcal{K}$ is invariant under the group of rotations of the inertia frame, and thus invariant under yaw rotations $q_1$; i.e., rotations about the $Z$-axis. Combining this with the observation that yaw rotations also leave the potential energy $\mathcal{V}$ invariant---for, the $Z$-axis of $\{ \mathcal{I} \}$ is aligned with gravity---leads to the conclusion that the Lagrangian $\mathcal{L}$ in \eqref{eq:lagrangian} is invariant under yaw rotations $q_1$. Since  $\mathcal{L}$ does not depend on $\mathrm{p}$ and $\dot{\mathrm{p}}$, we arrive at the conclusion that the vector field $\hat{f}$ in \eqref{eq:state_space} and the corresponding zero-input solutions are equivariant under the action \eqref{eq:Psi}, as required.    
 
Part~(\ref{P_S}) is a consequence of the flat ground assumption, which implies that the vertical distance $h^{\rm v}_{\rm sw}$ of the tip of the swing leg from the ground, which defines $\hat{\mathcal{S}}$ by \eqref{eq:impact_surface}, does not depend on translations along the $(X, Y)$-axes and rotations about the $Z$-axis of the inertia frame.

Finally, for part~(\ref{P_Delta}), note that 
\begin{equation}\nonumber
\hat{\Delta}\left( \hat{x}^- \right) = 
\begin{bmatrix}
\mathrm{p}^- + h^\mathrm{h}_\mathrm{sw}(q^-) \\
\Delta_q(q^-) \\
\Delta_{\dot{q}}(q^-, \dot{q}^-)
\end{bmatrix}
\end{equation}
where $\mathrm{p}^- $ is the position at the end of the step of the stance foot (where the body frame $\{\mathcal{B}\}$ is attached) which is the same as its position at the beginning of the step, and $q^-$ is the model's angular configuration at the end of the step;  $h^\mathrm{h}_\mathrm{sw}$ is a map that provides the $(X,Y)$ coordinates of the swing foot. Next, we show that $\hat{\Delta}$ is equivariant under the action $\Psi_\mathsf{g}$. First, note that based on observations in~\cite[Section IV-B]{spong2005controlled}, the forward kinematics map that associates to each configuration $q \in \mathcal{Q}$ the position and orientation of the tip of the swing foot is equivariant with respect to $\mathsf{SO}(3)$. Since the map $h^\mathrm{h}_\mathrm{sw}$ that participates in the first two components of $\hat{\Delta}$ is the $(X,Y)$ part of the forward kinematics map, we deduce that these components are equivariant under yaw rotations. Now, being the identity map, $\Delta_q$ is trivially equivariant with respect to yaw rotations. Finally, by~\cite[Lemma]{spong2005controlled} the map $\Delta_{\dot{q}}$ is also equivariant with respect to $\mathsf{SO}(3)$, and thus equivariant under yaw rotations. Thus, $\hat{\Delta}$ is equivariant under yaw rotations. Combining this with the fact that only the first two components of $\hat{\Delta}$ depend on $\mathrm{p}$ and are equivariant under translations along the $(X, Y)$-axes concludes the proof. 

\subsection{Proof of Proposition~\ref{prop:symmetry_P}}

Proposition~\ref{prop:symmetry_P} is a direct consequence of Proposition~\ref{prop:symmetry} and property \eqref{eq:sol_equivariance_cl}. To show that $\hat{P}$ is equivariant under the action $\Psi_\mathsf{g}$ defined by \eqref{eq:Psi}, we will show that its components $\hat{P}_\mathrm{LR}$ and $\hat{P}_\mathrm{RL}$ are equivariant under $\Psi_\mathsf{g}$. Indeed, if this holds  
\begin{align}
\nonumber
  \Psi_\mathsf{g} \circ \hat{P} &= \Psi_\mathsf{g} \circ  \hat{P}_\mathrm{LR} \circ \hat{P}_\mathrm{RL}
  \\
  \nonumber
  & = \hat{P}_\mathrm{LR} \circ \Psi_\mathsf{g}  \circ \hat{P}_\mathrm{RL} 
  \\
  \nonumber
  & = \hat{P}_\mathrm{LR} \circ \hat{P}_\mathrm{RL} \circ \Psi_\mathsf{g} 
   = \hat{P} \circ \Psi_\mathsf{g} 
\end{align}
where the second and third equations follow from the equivariance of $\hat{P}_\mathrm{LR}$ and $\hat{P}_\mathrm{RL}$, respectively. Let us focus on showing equivariance of $\hat{P}_\mathrm{LR} : \hat{\mathcal{S}}_\mathrm{R} \to \hat{\mathcal{S}}_\mathrm{L}$, defined by 
\begin{equation}\nonumber
\hat{P}_\mathrm{LR}(\hat{x}^-_\mathrm{R}) = \hat{\varphi}^\mathrm{cl}_\mathrm{L}(\hat{T}_\mathrm{L}(\hat{\Delta}_\mathrm{R}(\hat{x}^-_\mathrm{R})), \hat{\Delta}_\mathrm{R}(\hat{x}^-_\mathrm{R}))
\end{equation}
where $\hat{T}_\mathrm{L} : \mathcal{S}_\mathrm{L} \to \mathbb{R}_+$
\begin{equation}\nonumber
\hat{T}_\mathrm{L}(\hat{x}^+_\mathrm{L}) = \inf\{ t >0 ~|~ \hat{\varphi}^\mathrm{cl}_\mathrm{L}(t, \hat{x}^+_\mathrm{L}) \in \hat{\mathcal{S}}_\mathrm{L} \}
\end{equation}
is the time-to-impact map and $\hat{x}^+_\mathrm{L} = \hat{\Delta}_\mathrm{R}(\hat{x}^-_\mathrm{R})$. The arguments for the equivariance of $\hat{P}_\mathrm{RL}$ are entirely analogous. Note first that by the proof of Proposition~\ref{prop:symmetry}, $\hat{\mathcal{S}}_\mathrm{R}$ and $\hat{\mathcal{S}}_\mathrm{L}$ are invariant and $\hat{\Delta}_\mathrm{R}$ is equivariant under $\Psi_\mathsf{g}$. Furthermore, by property \eqref{eq:sol_equivariance_cl} the closed-loop flow $\hat{\varphi}^\mathrm{cl}_\mathrm{L}$ is equivariant under $\Psi_\mathsf{g}$. Thus, we only need to show that the time-to-impact map is well defined and equivariant under $\Psi_\mathsf{g}$. This effectively follows by the fact that $\hat{T}_\mathrm{L}$ is independent of the coordinates $(X,Y)$ and yaw angle, $q_1$, since the height of the swing foot $h^{\rm v}_{\rm sw}$ that determines the end of the swing phase is independent of these quantities; i.e.,  
\begin{equation} \nonumber
    \hat{\varphi}^\mathrm{cl}_\mathrm{L}(t, \hat{x}^+_\mathrm{L}) \in \hat{\mathcal{S}}_\mathrm{L} 
    ~\Rightarrow~
    h^{\rm v}_{\rm sw} (\hat{\varphi}^\mathrm{cl}_\mathrm{L}(t, \hat{x}^+_\mathrm{L}))=0
\end{equation}
The proof that $\hat{T}_\mathrm{L}$ is well defined follows from the implicit function theorem applied on  $\hat{\chi}(t,x) = h^{\rm v}_{\rm sw} \circ \hat{\varphi}^\mathrm{cl}_\mathrm{L}(t, x)$.

\section{Theorem~\ref{thm:dwell-time}: 
Discussion, Proof and Application}
\label{sec:proof}

\subsection{Discussion: Practical stability and Theorem~\ref{thm:dwell-time}}

The notion of practical stability was introduced by LaSalle and Lefschetz in~\cite[p. 121]{la1961stability}. Practical stability neither implies nor is implied by Lyapunov stability; the two concepts are distinct. While the latter provides a qualitative characterization of system behavior, practical stability aims at capturing quantitative aspects of it, such as estimates of solution bounds. Practical stability is essentially a uniform boundedness property of solutions relative to the set $\Omega_0$. However, it is not merely that  a bound exists; the bound is explicitly specified by the set $\Omega$ that ensures safe operation. To explain the meaning of these bounds, notice that the ratios in \eqref{eq:fixed-dwell} and \eqref{eq:avg-dwell} effectively capture two competing behaviors in the switching system. One is the expansive behavior related to the ``jumps'' caused by switching, and is captured by the parameter $\mu \geq 1$; the other, is the compressive behavior associated with the exponential decay between switchings, and is captured by the rate $\lambda \in (0,1)$. Focusing first on the effect of $\mu$, note that by \eqref{eq:mu-bound-V} a large $\mu$ implies that large ``spikes'' are possible on switching between subsystems. On the other hand, by \eqref{eq:fixed-dwell} and \eqref{eq:avg-dwell}, a large $\mu$ also increases the values of $\overline{N}_\mathrm{d}$ and $\overline{N}_\mathrm{a}$, which, by Definition~\ref{def:dwell-time}, leads to slower switching. Thus, longer intervals between switchings are enforced, allowing the compressing effect of exponential decay to dominate and keep the state bounded. Similarly when $\lambda$ is close to one; i.e., when convergence between switchings is slow.

\subsection{Proof of Theorem~\ref{thm:dwell-time}}

For convenience, the dependence of $\omega(\kappa)$ in \eqref{eq:omega-def} and $\mu(\kappa)$ in \eqref{eq:mu-def} on $\kappa$ will be dropped. Without loss of generality, assume that the system starts at $k=0$ and let $x_0 \in \Omega_0$. By the definition of $\Omega_0$, this implies that $x_0 \in {\cal M}_p(\omega)$ for all $p \in {\cal P}$ so that $x_0 \in {\cal M}_{\sigma(0)}(\omega)$ no matter what the index $\sigma(0) \in \mathcal{P}$ of the active system at $k=0$ is. Thus,
\begin{equation}\label{eq:v0}
V_{\sigma(0)} (x_0) \leq \omega \enspace.
\end{equation}
Combining \eqref{eq:V_2} with \eqref{eq:v0}, we have 
\begin{equation}\label{eq:v1}
V_{\sigma(0)} (x_k) \leq \lambda^k V_{\sigma(0)} (x_0) \leq \lambda^k \omega ~~\text{for~all}~~k \in [0, k_1] \enspace.
\end{equation}
Next, to treat switching, we address parts (i) and (ii) separately.

\subsubsection{Proof of Theorem~\ref{thm:dwell-time}(i)}
Consider any arbitrary (but fixed) switching signal $\sigma : \mathbb{Z}_+ \to {\cal P}$ with switching times $\{k_1, k_2, \ldots \}$ satisfying Definition~\ref{def:dwell-time}(\ref{def:dwell-time-fixed}) with (fixed) dwell time constraint \eqref{eq:fixed-dwell}. Recall that $\lambda \in (0,1)$ and so \eqref{eq:v1} implies  $V_{\sigma(0)} (x_k) \leq \omega \Leftrightarrow x_{k} \in \mathcal{M}_{\sigma(0)}(\omega)$ over $[0, k_1]$. Since $\mathcal{M}_{\sigma(0)}(\omega) \subset \Omega$ by the definition of $\Omega$, we have  
\begin{equation}\label{eq:xk_first_interval}
x_{k} \in  \Omega ~~\text{for~all}~~k\in[0, k_1] \enspace.
\end{equation}
As a result, $x_{k_1} \in \Omega$ and we distinguish the following cases:

\emph{Case (a):} $x_{k_1} \in \Omega_0$. 
\\
This case implies that $x_{k_1} \in {\cal M}_p(\omega)$ for all $p \in {\cal P}$ so that $x_{k_1} \in {\cal M}_{\sigma(k_1)}(\omega)$ no matter what the index $\sigma(k_1) \in \mathcal{P}$ is. Thus, $V_{\sigma(k_1)} (x_{k_1}) \leq \omega$, and arguing as we did to get \eqref{eq:xk_first_interval} we have $x_{k} \in \mathcal{M}_{\sigma(k_1)}(\omega) \subset \Omega$ for all $k \in [k_1, k_2]$.

\emph{Case (b):} $x_{k_1} \in \Omega \setminus  \Omega_0$. 
\\
We show that this case is not possible due to the dwell time constraint \eqref{eq:fixed-dwell} with\footnote{The dwell-time bound \eqref{eq:fixed-dwell} is designed so that \emph{Case (b)} does not occur. To see this, the arguments in the proof imply that if $x_{k_1} \in \Omega \setminus  \Omega_0$ were true, then $V_p(x_{k_1}) \leq \mu \lambda^{k_1} V_{\sigma(0)}(x_0)$ for any $p \in \mathcal{P}$ that is switched in at $k_1$. But then, due to the fact that $V_{\sigma(0)}(x_0) \leq \omega$, if $k_1$ is large enough so that $\mu \lambda^{k_1} \leq 1$, we have that $V_p(x_{k_1}) \leq \omega$ for any $p \in \mathcal{P}$, which implies that $x_{k_1} \in \Omega_0$ contradicting the original assumption. By Definition \ref{def:dwell-time}(\ref{def:dwell-time-fixed}), a switching signal $\sigma$ with dwell time $N_\mathrm{d}$ satisfies $k_{i+1}-k_i \geq N_\mathrm{d}$ for every pair of switching instants, which, since $k_0=0$, implies $k_1 \geq N_\mathrm{d}$. This fact, combined with $\lambda \in (0,1)$, imply that a switching signal with dwell time $N_\mathrm{d}$ that satisfies $\mu \lambda^{N_\mathrm{d}} \leq 1$ also satisfies $\mu \lambda^{k_1} \leq 1$, thus leading to the desired contradiction. The dwell time bound \eqref{eq:fixed-dwell} then results from $\mu \lambda^{N_\mathrm{d}} \leq 1$ by taking logarithms.} $\overline{N}_\mathrm{d}=\ln{\mu} / \ln{(1/\lambda)}$. By \eqref{eq:N_sub_Munder}, $\mathcal{M}(\kappa) \subset \Omega_0$ and thus the fact that  $x_{k_1} \notin \Omega_0$ implies that $x_{k_1} \notin \mathcal{M}(\kappa)$. Then, \eqref{eq:mu-bound-V} can be used to obtain $V_{p}(x_{k_1}) \leq \mu V_{\sigma(0)}(x_{k_1})$ for all $p \in \mathcal{P}$, which by \eqref{eq:v1} results in $V_{p}(x_{k_1})  \leq \mu \lambda^{k_1} V_{\sigma(0)}(x_0)$ for all $p \in \mathcal{P}$. Then, since $k_1 \geq \overline{N}_{\rm d}$ by the definition of the fixed dwell time constraint, we obtain  
\begin{equation}\label{eq:vk1}
V_{p}(x_{k_1})  \leq \mu \lambda^{\overline{N}_{\rm d}} V_{\sigma(0)}(x_0) ~~~ \text{for~all} ~~~ p \in \mathcal{P} \enspace .
\end{equation}
In view of \eqref{eq:fixed-dwell}, we have $\mu \lambda^{\overline{N}_{\rm d}} \leq 1$, and so \eqref{eq:v0} and \eqref{eq:vk1} result in $V_{p}(x_{k_1})  \leq  \omega$ for all $p \in \mathcal{P}$. This implies that for any $p \in {\cal P}$ that is ``switched in'' at $k_1$, $x_{k_1} \in \mathcal{M}_p (\omega)$. Thus, $x_{k_1} \in \Omega_0$, which contradicts our assumption that $x_{k_1} \in \Omega \setminus \Omega_0$, essentially guaranteeing that \emph{Case (b)} does not occur.

To summarize, for any $x_0 \in \Omega_0$, we have shown in \eqref{eq:xk_first_interval} that $x_{k} \in \Omega$ over the interval $[0, k_1]$. Then, the constraint \eqref{eq:fixed-dwell} on the dwell time ensures that $x_{k_1} \in \Omega_0$ so that $x_k \in \Omega$ over the interval $[k_1 , k_2]$. Propagating this construction to future time steps proves the result.  

\subsubsection{Proof of Theorem~\ref{thm:dwell-time}(ii)}
Our argument for the average dwell time differs from the proof of part (i). This is because we can no longer ensure that every successive switching times are separated by at least a fixed number of steps. Consider again any arbitrary (but fixed) switching signal $\sigma : \mathbb{Z}_+ \to {\cal P}$ with switching times $\{k_1, k_2, \ldots \}$ satisfying Definition~\ref{def:dwell-time}(\ref{def:dwell-time-avg}) with average dwell time constraint \eqref{eq:avg-dwell}.
Let $\underline{n} = \inf \{n \in \mathbb{Z}_+ \cup \{\infty\} ~|~ x_{k_n} \in \accentset{\circ}{\mathcal{M}}_{\sigma(k_n-1)}(\kappa)\}$ be the index of the first switching time $k_{\underline{n}}$ for which \eqref{eq:mu-bound-V} cannot be applied to bound the value at $x_{k_{\underline{n}}}$ of the Lyapunov function of the system that is ``switched in.'' Then, the following claim is true
\begin{equation}\nonumber
\text{\emph{Claim:}}~
V_{\sigma(k)}(x_k) \leq \mu^{N_0} V_{\sigma(0)}(x_0)~~\text{for~all}~~ k \in [0, k_{\underline{n}}) \enspace.
\end{equation}
The proof of this claim is postponed until the end of the main argument, which considers the following cases:

\emph{Case (a):} $\underline{n} = \infty$.
\\
In this case, the claim applies for all $k \geq 0$ and using \eqref{eq:v0} we have that $V_{\sigma(k)}(x_k) \leq \mu^{N_0} \omega \Leftrightarrow x_k \in \Omega$ for all $k \geq 0$.

\emph{Case (b):} $\underline{n} < \infty$
\\
The claim holds over the interval $[0, k_{\underline{n}})$. By the definition of $\underline{n}$ we have that $x_{k_{\underline{n}}} \in \accentset{\circ}{\mathcal{M}}_{\sigma(k_{\underline{n}-1)}}(\kappa) \subset \mathcal{M}(\kappa) \subset \Omega_0$ by \eqref{eq:N_sub_Munder} and the definition of $\Omega_0$. Hence,
\begin{equation}\label{eq:Vsigma_b}
    V_{\sigma(k_{\underline{n}})}(x_{k_{\underline{n}}}) \leq \omega        
\end{equation}
Since $\mu \geq 1$ and $N_0 \geq 1$, \eqref{eq:Vsigma_b} also implies $V_{\sigma(k_{\underline{n}})}(x_{k_{\underline{n}}}) \leq \mu^{N_0} \omega$, and thus we can extend the claim above over the closed interval $[0, k_{\underline{n}}]$ to get $x_k \in \Omega$ for all $k \in [0, k_{\underline{n}}]$. To complete the proof, treat $k = k_{\underline{n}}$ as a new initial time and define $k' = k - k_{\underline{n}}$. Since \eqref{eq:Vsigma_b} corresponds to \eqref{eq:v0} for the translated time axis, we can re-use the claim exactly as we did above but for the translated time axis to either show $x_{k'} \in \Omega$ for all $k' \geq 0$ or to show that the claim can be extended over $[0, k'_{\underline{n}}]$. These arguments can be propagated indefinitely to show that the solution never leaves $\Omega$, thus concluding the proof.

\emph{Proof of the claim:} 
We first show the claim for $\underline{n}=1$. By \eqref{eq:v1} we have $V_{\sigma(0)}(x_{k}) \leq \lambda^{k} V_{\sigma(0)}(x_0)$ for all $k \in [0, k_1]$. Noting that $\sigma(k)=\sigma(0)$ over $[0,k_1)$, we can write $V_{\sigma(k)}(x_{k}) \leq \lambda^{k} V_{\sigma(0)}(x_0)$ for all $k \in [0, k_1)$. Since $\lambda \in (0,1)$, $\mu \geq 1$ and $N_0 \geq 0$, this inequality implies $V_{\sigma(k)}(x_{k}) \leq V_{\sigma(0)}(x_0) \leq \mu^{N_0} V_{\sigma(0)}(x_0)$ over $[0, k_1)$. Thus, 
\begin{equation}\label{eq:claim_0_k1}
    V_{\sigma(k)}(x_{k}) \leq \mu^{N_0} V_{\sigma(0)}(x_0)
    ~~\text{for~all}~~k \in [0, k_1)
\end{equation}
which confirms the claim for $\underline{n}=1$.    

If, on the other hand, $\underline{n}>1$ so that $x_{k_1} \notin \accentset{\circ}{\mathcal{M}}_{\sigma(k_1-1)}$ we can apply \eqref{eq:mu-bound-V} to get $V_p(x_{k_1}) \leq \mu \lambda^{k_1} V_{\sigma(0)}(x_0)$ for all $p \in \mathcal{P}$. Thus, no matter what the index $\sigma(k_1) \in \mathcal{P}$ is, we arrive at $V_{\sigma(k_1)}(x_{k_1}) \leq \mu \lambda^{k_1} V_{\sigma(0)}(x_0)$. Noting that $\sigma(k) = \sigma(k_1)$ for $k \in [k_1, k_2)$ and using \eqref{eq:V_2} we have
\begin{equation}\nonumber
    V_{\sigma(k)}(x_{k}) \leq \mu \lambda^{k} V_{\sigma(0)}(x_0)
    ~~\text{for~all}~~k \in [k_1, k_2) \enspace.
\end{equation}
This argument can be propagated forward in time as follows. Let $N_\sigma$ be the number of switches over $[0,k_{\underline{n}})$; note that when $\underline{n}>1$, we have $1 \leq N_\sigma < \underline{n}$. Then, arguing as above we get
\begin{equation}\nonumber
    V_{\sigma(k)}(x_{k}) \leq \mu^{N_\sigma} \lambda^{k} V_{\sigma(0)}(x_0)
    ~~\text{for~all}~~k \in [k_{N_\sigma}, k_{N_\sigma+1}) \enspace.
\end{equation}
By the definition \eqref{eq:avg-dwell-time-def} of the average dwell time we have $N_\sigma \leq N_0 + k / N_\mathrm{a}$, and substituting we get 
\begin{equation}\nonumber
    V_{\sigma(k)}(x_{k}) \leq \mu^{N_0} ( \mu^{1/N_\mathrm{a}} \lambda)^{k} V_{\sigma(0)}(x_0)
\end{equation}
for all $k \in [k_{N_\sigma}, k_{N_\sigma+1})$. Now, recall that $\mu \geq 1$ and notice that the average dwell time constraint \eqref{eq:avg-dwell} requires $N_\mathrm{a} \geq \overline{N}_\mathrm{a}$ with $\overline{N}_{\rm a}= \ln{(\mu)} / \ln{(1/\lambda)}$ so that $\mu^{1/N_\mathrm{a}} \lambda \leq 1$. As a result, we obtain the following bound
\begin{equation}\nonumber
    V_{\sigma(k)}(x_{k}) \leq \mu^{N_0} V_{\sigma(0)}(x_0)
    ~~\text{for~all}~~k \in [k_{N_\sigma}, k_{N_\sigma+1}) \enspace.
\end{equation}
Note that this bound holds for all $k \in [k_{N_\sigma}, k_{N_\sigma+1})$ and for \emph{any} $1 \leq N_\sigma < \underline{n}$; i.e., it holds for all $k \in [k_1, k_{\underline{n}})$. Combining this with \eqref{eq:claim_0_k1} completes the proof of the claim.

\subsection{Application of Theorem~\ref{thm:dwell-time}: Relevant Computations}
\label{app:sec:implementation}

\subsubsection{Estimation of the BoA} 
It is beneficial to work with quadratic Lyapunov functions computed via linearization and certified via \ac{SOS} programming. For convenience, we define the coordinates $\bar{x} = x - x^*_p$, with respect to which we have
\begin{equation}\label{eq:P-translated}
\bar{x}_{k+1} = \rho_p(\bar{x}_k)
\end{equation}
where $\rho_p(\bar{x}) = P_p(\bar{x} + x^*_p)-x^*_p$ and $\rho_p(0)=0$. Since $x^*_p$ is a locally exponentially stable fixed point for $P_p$, so is $0$ for $\rho_p$, and thus the eigenvalues of the linearization $A_p = \partial{\rho_p}(\bar{x}) / \partial{\bar{x}}|_{\bar{x}=0}$ are all within the unit disc centered at the origin. Then, a Lyapunov function for \eqref{eq:P-translated} over a neighborhood of the origin can be computed by   
\begin{equation} \nonumber 
V_p(\bar{x})= \bar{x}^\mathsf{T} S_p \bar{x}
\end{equation}
where $S_p$ is the unique positive definite solution of the discrete Lyapunov equation $A^\mathsf{T}_p S_p A_p - S_p + Q_p = 0$ with $Q_p$ positive definite; here, $Q_p=I$, the identity matrix, for all $p \in \mathcal{P}$.

To compute an inner approximation of the  \ac{BOA} of the fixed point $\bar{x}^*=0$ of \eqref{eq:P-translated}, we employ \ac{SOS} programming. As in~\cite{tedrake2010}, we first approximate $\rho_p$ by  its Taylor series up to second order, denoted by $\bar{\rho}_p$. Then, a compact inner approximation of the \ac{BOA} can be computed by finding the largest $\overline{\kappa}_p>0$ for which the implication
\begin{equation}\label{eq:BoA-condition-approx}
V_p(\bar{x}) \leq \overline{\kappa}_p  ~\Longrightarrow~ \lambda_p V_p (\bar{x}) - V_p \left( \bar{\rho}_p (\bar{x}) \right) \geq 0 
\end{equation}
can be verified. This is done by formulating the problem as a sequence of \ac{SOS} feasibility programs~\cite{tedrake2010}, which result in the maximum value of $\overline{\kappa}_p$.

\subsubsection{Feasibility and dwell-time constraints}
\label{subsubsec:computations-feasibility}

Feasibility of switching among fixed points requires the verification of condition \eqref{eq:feasibility-2}, which can be done simply by checking 
\begin{equation}\nonumber
(x^*_q - x^*_p)^\mathsf{T} S_p (x^*_q - x^*_p) \leq \overline{\kappa}_p ~~\text{for all pairs}~~ p, q \in \mathcal{P}  \enspace.
\end{equation}

To apply Theorem~\ref{thm:dwell-time}, a suitable value for the parameter $\kappa > 0$ must be determined. This involves the computation of $\omega(\kappa)$ and $\mu(\kappa)$, which, owing to the quadratic form of the Lyapunov functions $V_p$  can be done in closed form, utilizing the following bounds~\cite[Proposition~1]{Veer2020TAC} 
\begin{align}
\label{eq:omega-bound}
\omega(\kappa) & \!\leq \max_{p,q\in\mathcal{P}} \Bigg[ \lambda_{\max}(S_p) \left( \sqrt{\frac{\kappa}{\lambda_{\min}(S_q)}} + D_{p,q} \right)^2 \Bigg]  \\
\label{eq:mu-bound}
\mu(\kappa) & \!\leq \max_{p,q\in\mathcal{P}} \Bigg[ \frac{\lambda_{\max}(S_q)}{\lambda_{\min}(S_p)}\left( 1 + \sqrt{\frac{\lambda_{\max}(S_p)}{\kappa}} D_{p,q} \right)^2 \Bigg]
\end{align}
where $D_{p,q}=\|x_p^*-x_q^*\|$ and $\lambda_{\min}$ and $\lambda_{\max}$ denote the minimum and maximum eigenvalues of the corresponding matrices, respectively.

With these bounds Theorem~\ref{thm:dwell-time} is applied as follows. For the (fixed) dwell  time case, pick a $\kappa > 0$, compute the bound on $\omega(\kappa)$ and check if \eqref{eq:feasible-0-inp-fixed} is satisfied. If it is, compute $\overline{N}_\mathrm{d}$ using the bound on $\mu(\kappa)$ and \eqref{eq:fixed-dwell}. For the average dwell time, again start with selecting a $\kappa>0$ and compute the bounds on $\omega(\kappa)$ and $\mu(\kappa)$ as above. If the inclusion \eqref{eq:feasible-0-inp} holds for some $\overline{N}_0 \geq 1$, then provide the planner with the numbers $(\overline{N}_0, \overline{N}_{\rm a})$ where $\overline{N}_{\rm a}$ is obtained by \eqref{eq:avg-dwell}. In both cases, if the set inclusions \eqref{eq:feasible-0-inp-fixed} or \eqref{eq:feasible-0-inp} cannot be verified, choose a new $\kappa$ and repeat the procedure. Finally, note that for higher-dimensional spaces, checking \eqref{eq:feasible-0-inp-fixed} or \eqref{eq:feasible-0-inp} can be done using convex optimization tools as in~\cite[Section 8.4]{Boyd2004book}.

\subsection{Discussion: On fixed versus average dwell time}
\label{app:dwell_time}

\begin{figure}[b!]
\centering
\vskip -5pt
\subfigure[]{\includegraphics[width=0.23\textwidth]{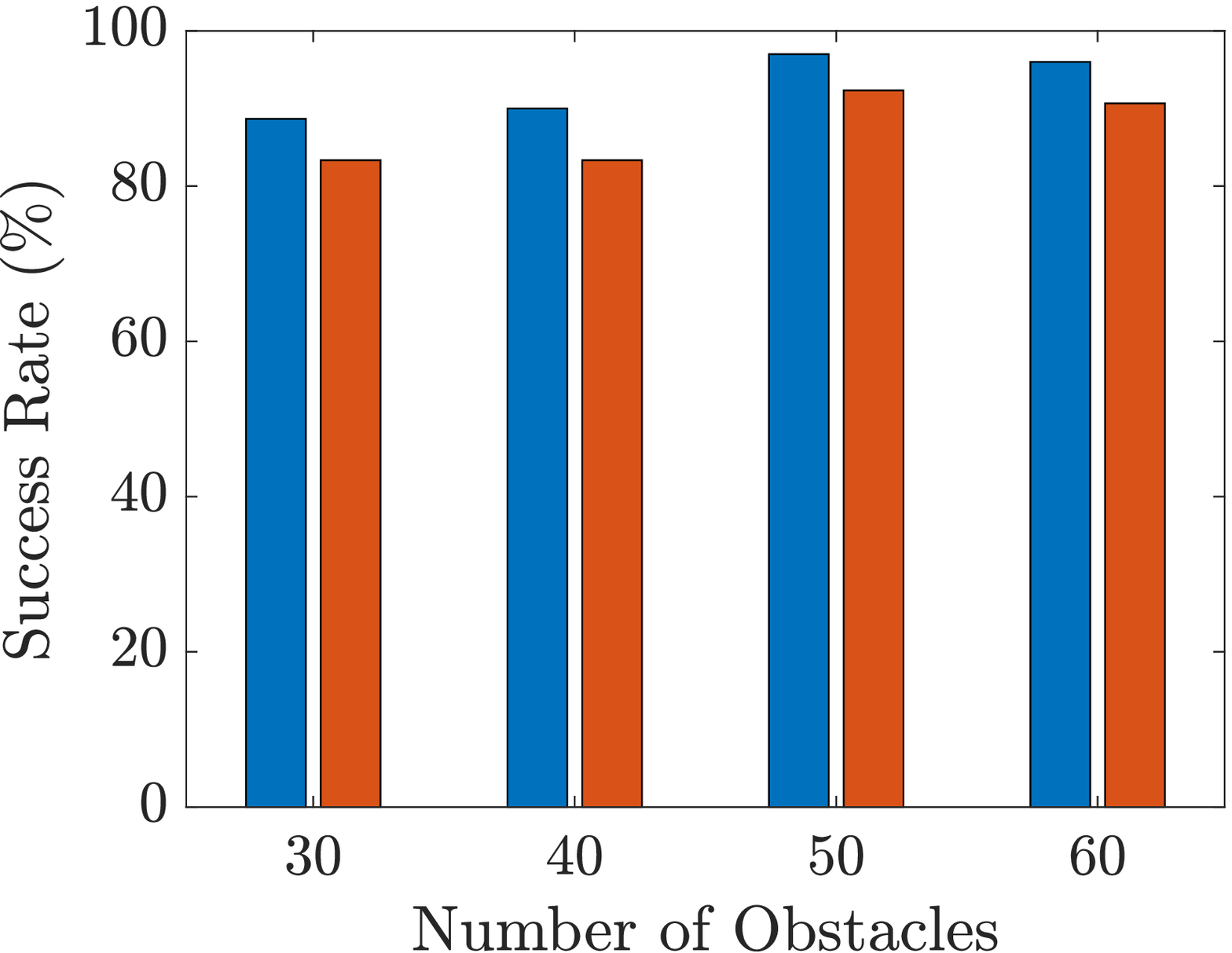} \label{fig:fixed_average_success_rates}
}
\subfigure[]{\includegraphics[width=0.23\textwidth]{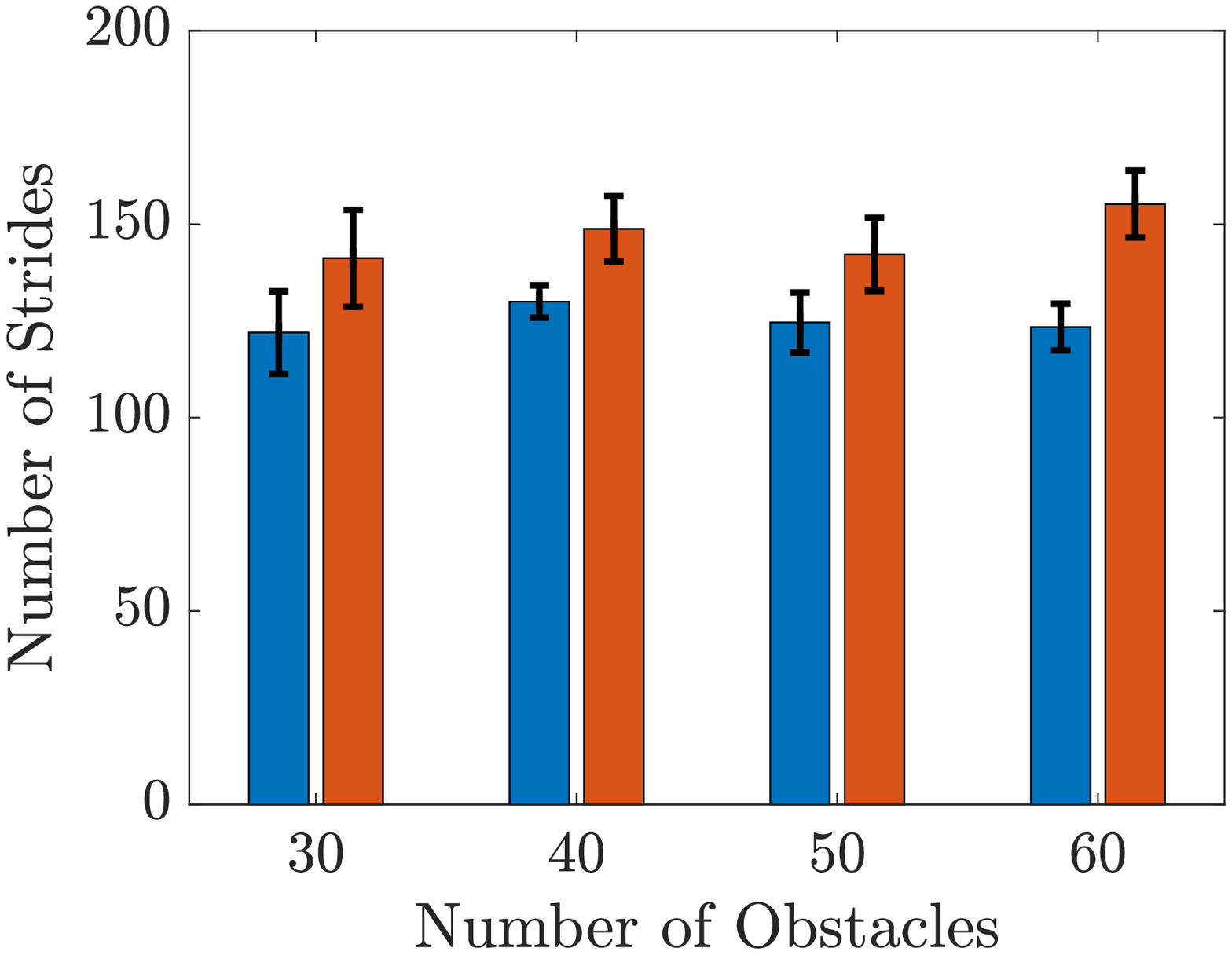} \label{fig:fixed_average_strides}
}
\caption{(a) Success rates over 600 randomly generated environments for $\mathcal{C}_{\rm d}[\overline{N}_{\rm d}=2]$ (orange) and $\mathcal{C}_{\rm a}[\overline{N}_0=2,\overline{N}_{\rm a}=2]$ (blue). (b) Average number of strides required to reach the goal.}
\label{fig:fixed_average_dwell_time}
\vskip -10pt
\end{figure}

In general, insisting that the system can switch to a new primitive only after a \emph{fixed} number $N_\mathrm{d}$ of strides are completed may be overly restrictive when $\overline{N}_\mathrm{d} >1$ in \eqref{eq:fixed-dwell}. Particularly for path planning in tight environments, a (fixed) dwell time constraint can result in longer paths or even in failure of the planning algorithm to compute a feasible plan. Indeed, let us compare two classes of switching signals, one with a fixed dwell time constraint $\overline{N}_{\rm d}=2$ denoted by $\mathcal{C}_{\rm d}[\overline{N}_{\rm d}=2]$ and another with an average dwell time constraint with $\overline{N}_0=2$ and $\overline{N}_{\rm a}=2$ denoted by $\mathcal{C}_{\rm a}[\overline{N}_0=2,\overline{N}_{\rm a}=2]$; note that $\mathcal{C}_{\rm d}[\overline{N}_{\rm d}=2] \subset \mathcal{C}_{\rm a}[\overline{N}_0=2,\overline{N}_{\rm a}=2]$. For the comparison, we use the \ac{RRT} algorithm from~\cite{motahar2016composing} with the library of primitives computed in Section~\ref{subsec:example-actions} over the $600$ randomly generated environments of Section~\ref{subsec:MPC}. The algorithm terminates if a path is found that brings the robot within a disc of radius $1\mathrm{m}$ around the goal, or if the number of iterations exceeds 10000. Figure~\ref{fig:fixed_average_dwell_time} summarizes the results. It can be seen in Fig.~\ref{fig:fixed_average_success_rates} that the average dwell time constraint results in higher success rates compared to the fixed dwell time one. Furthermore, Fig.~\ref{fig:fixed_average_dwell_time} shows that former also results in shorted paths, requiring a smaller number of strides on average. These results provide an indication of the potentially enhanced flexibility offered by the average  over the fixed dwell time constraint when $\overline{N}_\mathrm{d} >1$.

\section{Virtual Constraint Design}
\label{app:HZD-controller}

In each single support phase, the output \eqref{eq:output_function} has the form
\begin{equation}\label{eq:leg_output_cor_steer}
y =  q_{\rm a} - h_{\rm d} (\theta) 
 - h_{\rm c}(\theta,y_{\rm i},\dot{y}_{\rm i})
 - h_{\rm s}(\theta,\beta)
\end{equation}
where $q_{\rm a} = \begin{bmatrix} q_3 & q^\mathsf{T}_\mathrm{r} \end{bmatrix}^\mathsf{T}$ are the actuated variables and $\theta$ is a monotonic quantity  chosen as in~\cite{3D-TRO2009} to be the angle of the line connecting the point of contact of the support leg with the corresponding hip joint; i.e., $\theta = - q_2 -  q_4 /2$. Note that the outputs \eqref{eq:leg_output_cor_steer} depend only on $q_2$, $q_3$ and $q_\mathrm{r}$, as required by \eqref{eq:output_function}. Thus, the corresponding output-zeroing controller \eqref{eq:fdb-general} depends only on the $x$-part of \eqref{eq:x_hat} and the equivariance property \eqref{eq:sol_equivariance_cl} holds for the closed-loop solutions. 

To specify the outputs \eqref{eq:leg_output_cor_steer}, we begin with the function $h_{\rm d}$, which corresponds to third-degree Bezier polynomials designed to generate straight-line walking motions, as in~\cite[Section IV]{3D-TRO2009}. Following the design of $h_\mathrm{d}$, the correction term $h_\mathrm{c}$ is introduced in \eqref{eq:leg_output_cor_steer} with the purpose of satisfying condition~C.\ref{O1} of Section~\ref{subsec:reduction-feedback}. This is done by selecting the functions $h_\mathrm{c}$ so that the post-impact error of the ``uncorrected'' part $q_{\rm a} - h_{\rm d} (\theta)$ of the output is smoothly driven to zero by the middle of the step; the details can be found in~\cite[Section V]{3D-TRO2009}. The end result is a straight-line, limit-cycle walking gait with local dynamics represented by the reduced-order gait primitive $\mathcal{G}_1|_{\mathcal{S} \cap \mathcal{Z}} = \{P_1|_{\mathcal{S} \cap \mathcal{Z}},~z^*_1\}$.

Having designed $h_\mathrm{d}$ and $h_\mathrm{c}$, we turn our attention to the term $h_\mathrm{s}$, the purpose of which is to obtain turning gaits that satisfy condition~C.\ref{O2} of Section~\ref{subsec:reduction-feedback} with $\mathcal{S} \cap \mathcal{Z}$ being the same surface as in $\mathcal{G}_1|_{\mathcal{S} \cap \mathcal{Z}}$. Let $\theta^+$ and $\theta^-$  be the values of $\theta$ at the beginning (post-impact) and the end (pre-impact) of a---right or left---step. First, we require
\begin{equation}\nonumber
h_{\rm s}(\theta,0)=0~~~\text{for all}~\theta^+\leq\theta\leq\theta^-
\end{equation}
so that when $\beta=0$ the term $h_\mathrm{s}$ does not affect the terms $h_\mathrm{d}$ and $h_\mathrm{c}$ designed above to obtain the straight-line walking gait $\mathcal{G}_1$; thus, $\mathcal{G}_1$ corresponds to $\beta_1=0$. Next, we define $\theta_{\rm s} = \theta^+ + 0.9 (\theta^- - \theta^+)$ and impose the conditions
\begin{eqnarray}\label{eq:hs-conditions}
	\begin{cases}
		\begin{aligned}
		& h_{\rm s}(\theta^+,\beta) = 0,~\frac{\partial h_{\rm s}}{\partial \theta}(\theta^+,\beta) = 0\\
		& h_{\rm s}(\theta_{\rm s},\beta) = 0,~\frac{\partial^i h_{\rm s}}{\partial \theta^i}(\theta_{\rm s},\beta) = 0,~i = 1,2\\
		& h_{\rm s}(\theta,\beta) = 0,~~\mathrm{for}~~\theta_s \leq \theta \leq \theta^-
		\end{aligned}
	\end{cases}
\end{eqnarray}
so that $h_{\rm s}(\theta,\beta)$ vanishes at the beginning of the step (when $\theta = \theta^+$) and after 90\% of the step is completed (when $\theta \in [\theta_{\rm s}, \theta^-)$). Thus, right after one impact and before the next the surfaces $\mathcal{Z}_p$ agree with the surface $\mathcal{Z}$ of the straight-line gait. As a result, all $\mathcal{Z}_p$'s have a common intersection with  $\mathcal{S}$ and condition~C.\ref{O2} of Section~\ref{subsec:reduction-feedback} is satisfied; see also Fig.~\ref{fig:multi-surface}.

Now, we are ready to use these constructions to compute limit cycles that correspond to different values $\beta_p$ of the parameters $\beta$, and result in turning walking gaits with different changes in orientation. The challenge here is to ensure that the corresponding fixed points $z^*_p$ are all in the vicinity of $z^*_1$. This  facilitates provably safe planning by increasing the overlap among the \ac{BOA}s of the fixed points, and by reducing the dwell time constraint for practical stability of gait switching \eqref{eq:switch-system-zd}. A simple way to achieve this is as follows. Given a straight-line gait primitive $\mathcal{G}_1|_{\mathcal{S} \cap \mathcal{Z}} = \{P_1|_{\mathcal{S} \cap \mathcal{Z}},~z^*_1\}$ and the corresponding maps $h_\mathrm{d}$ and $h_\mathrm{c}$ introduce the term $h_\mathrm{s}$ so that the conditions \eqref{eq:hs-conditions} are satisfied. The resulting stride map depends on $\beta$ as  
\begin{equation} \label{eq:decomp-poincare-beta}
\begin{bmatrix}
(q_1)_{k+1} \\ z_{k+1} 
\end{bmatrix} = 
\begin{bmatrix}
(q_1)_k + H^{(q_1)} |_{\mathcal{S} \cap \mathcal{Z}} (z_k,\beta) \\ 
P|_{\mathcal{S} \cap \mathcal{Z}} (z_k,\beta)
\end{bmatrix}
\end{equation}
where we kept only the orientation component $H^{(q_1)}|_{\mathcal{S} \cap \mathcal{Z}}$ of the displacement map $H|_{\mathcal{S} \cap \mathcal{Z}}$. Note that $P|_{\mathcal{S} \cap \mathcal{Z}} (z,0) = P_1|_{\mathcal{S} \cap \mathcal{Z}} (z)$ and $H |_{\mathcal{S} \cap \mathcal{Z}} (z,0) = H_1|_{\mathcal{S} \cap \mathcal{Z}} (z)$, thus these maps are well defined on the reduced-order switching surface $\mathcal{S} \cap \mathcal{Z}$ associated with primitive $\mathcal{G}_1$. 

Linearizing \eqref{eq:decomp-poincare-beta} about $(z^*_1,0)$ results in
\begin{equation} \label{eq:linearization}
\begin{bmatrix}
(\delta q_1)_{k+1} \\ \delta z_{k+1} \end{bmatrix} 
= 
\begin{bmatrix}
A_{q_1} \delta z_k + G_{q_1} \beta \\ A_z \delta z_k + G_z \beta
\end{bmatrix}
\end{equation}
where $A_{q_1}, A_z$ and $G_{q_1}, G_z$ are suitable Jacobians. To find the value of $\beta$ that corresponds to turning by a desired angle $\psi$ without drastically affecting the state part $z$, we substitute $\delta q_1 = \psi$  and $\delta z=0$ in \eqref{eq:linearization} and solve for $\beta$ to get $\beta = G^{-\rm R}\begin{bmatrix} \psi & 0 \end{bmatrix}^\mathsf{T}$, where $G^{-\rm R}$ is the right pseudoinverse of $G= \begin{bmatrix} G^\mathsf{T}_{q_1} & G^\mathsf{T}_{z} \end{bmatrix}^\mathsf{T}$. Then, using the $h_\mathrm{d}$ and $h_\mathrm{c}$ that correspond to the straight-line  gait primitive $\mathcal{G}_1|_{\mathcal{S} \cap \mathcal{Z}} = \{P_1|_{\mathcal{S} \cap \mathcal{Z}},~z^*_1\}$ and $h_\mathrm{s}$ with $\beta$ computed as above we simulate the system until it converges to a new fixed point $z^*_\beta$ which is in the vicinity of $z^*_1$ and causes turning approximately by $\psi$. Repeating this procedure for different desired values of turning angle $\psi_p$, we obtain different values of $\beta_p$. This way, different turning fixed points $z^*_p$ can be generated, which lie in the vicinity of the straight-line fixed point $z^*_1$.

\section{Library of Planning Actions in Section \ref{subsec:example-actions}}
\label{app:Polynomial_approximations}

\begin{figure*}[b!]
\vskip +5pt 
\centering
\includegraphics[width =0.8\linewidth]{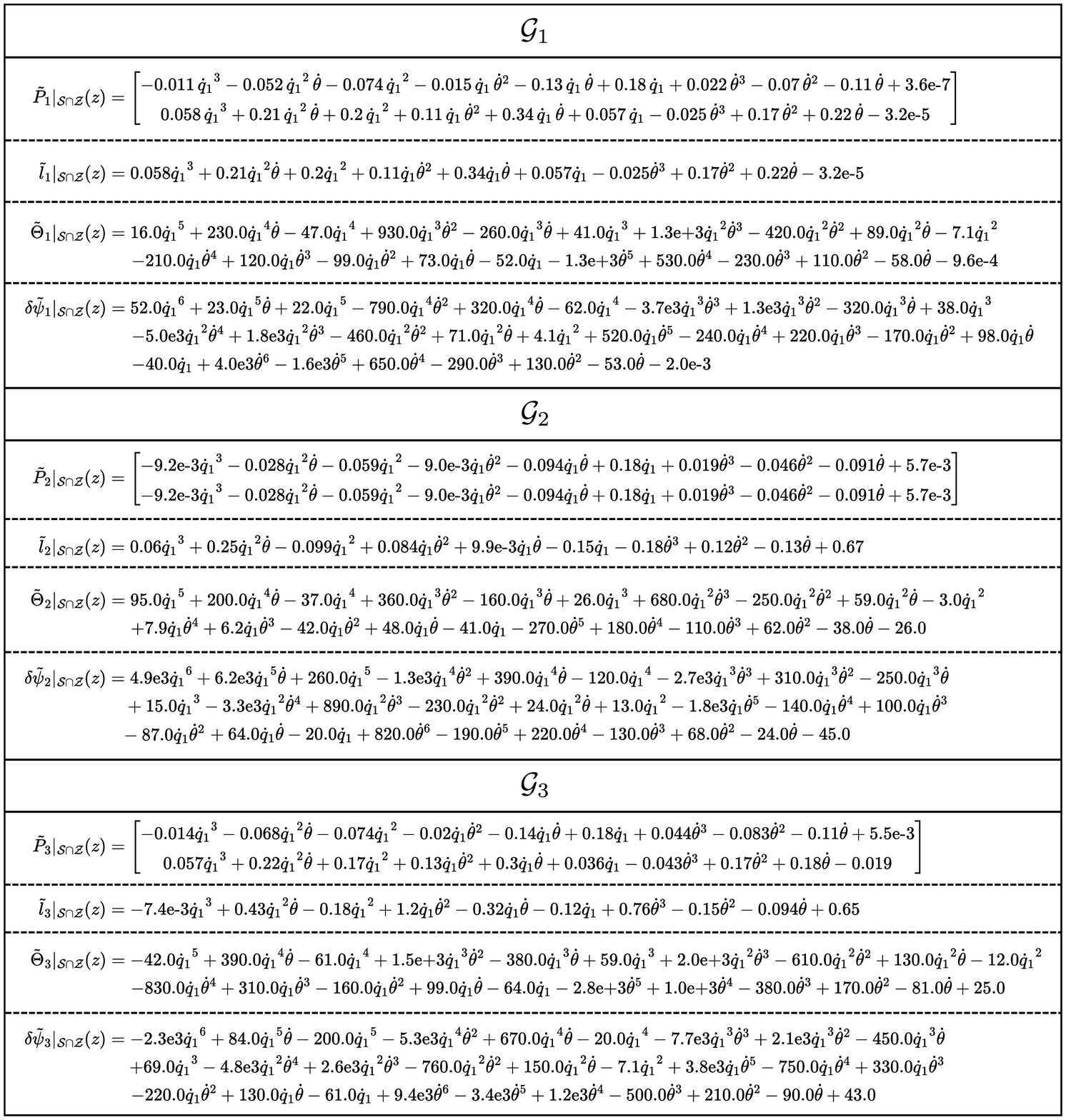}
\caption{ Polynomial approximations of the primitives $\mathcal{G}_1$, $\mathcal{G}_2$ and $\mathcal{G}_3$ from Section~\ref{subsec:example-actions} where $z = (\dot{q}_1,\dot{\theta})$.} 
\label{fig:polynomials}
\vskip -5pt
\end{figure*}

In polar form, the displacement map $\tilde{H}_p|_{\mathcal{S} \cap \mathcal{Z}}(z)$ from \eqref{eq:polys} consists of $\tilde{l}_p|_{\mathcal{S} \cap \mathcal{Z}}(z)$, $\delta\tilde{\psi}_p|_{\mathcal{S} \cap \mathcal{Z}}(z)$ and $\tilde{\Theta}_p|_{\mathcal{S} \cap \mathcal{Z}}(z)$ (see \cite{Veer2017IROS}). If $\mathsf{g}_k = [\mathrm{p}^\mathsf{T}_k ~ \psi_k]^\mathsf{T}$ is the pose of the robot at stride $k$ then the predicted output at the end of the stride is
\begin{equation*}
    \mathsf{g}_{k+1} = \mathsf{g}_{k} + \begin{bmatrix}
    \tilde{l}_p|_{\mathcal{S} \cap \mathcal{Z}}(z_k)\cos(\psi_k + \tilde{\Theta}_p|_{\mathcal{S} \cap \mathcal{Z}}(z_k)) \\
    \tilde{l}_p|_{\mathcal{S} \cap \mathcal{Z}}(z_k)\sin(\psi_k + \tilde{\Theta}_p|_{\mathcal{S} \cap \mathcal{Z}}(z_k)) \\
    \delta\tilde{\psi}_p|_{\mathcal{S} \cap \mathcal{Z}}(z_k)
\end{bmatrix} \enspace .
\end{equation*}
For $\tilde{P}_p|_{\mathcal{S} \cap \mathcal{Z}}(z)$, $M_P=3$ and for $\tilde{l}_p|_{\mathcal{S} \cap \mathcal{Z}}(z)$, $\delta\tilde{\psi}_p|_{\mathcal{S} \cap \mathcal{Z}}(z)$, and $\tilde{\Theta}_p|_{\mathcal{S} \cap \mathcal{Z}}(z))$,  $M_H$ is 3, 6, and 5 respectively. The polynomial approximations for $\tilde{P}_p|_{\mathcal{S} \cap \mathcal{Z}}(z)$, $\tilde{l}_p|_{\mathcal{S} \cap \mathcal{Z}}(z)$, $\delta\tilde{\psi}_p|_{\mathcal{S} \cap \mathcal{Z}}(z)$, and $\tilde{\Theta}_p|_{\mathcal{S} \cap \mathcal{Z}}(z)$ are shown in Fig.~\ref{fig:polynomials}.

\end{appendices}

\bibliographystyle{IEEEtran}
\bibliography{legged_3D}

\begin{thebibliography}{10}
\providecommand{\url}[1]{#1}
\csname url@samestyle\endcsname
\providecommand{\newblock}{\relax}
\providecommand{\bibinfo}[2]{#2}
\providecommand{\BIBentrySTDinterwordspacing}{\spaceskip=0pt\relax}
\providecommand{\BIBentryALTinterwordstretchfactor}{4}
\providecommand{\BIBentryALTinterwordspacing}{\spaceskip=\fontdimen2\font plus
\BIBentryALTinterwordstretchfactor\fontdimen3\font minus
  \fontdimen4\font\relax}
\providecommand{\BIBforeignlanguage}[2]{{%
\expandafter\ifx\csname l@#1\endcsname\relax
\typeout{** WARNING: IEEEtran.bst: No hyphenation pattern has been}%
\typeout{** loaded for the language `#1'. Using the pattern for}%
\typeout{** the default language instead.}%
\else
\language=\csname l@#1\endcsname
\fi
#2}}
\providecommand{\BIBdecl}{\relax}
\BIBdecl

\bibitem{mcgeer1990passive}
T.~McGeer, ``Passive dynamic walking,'' \emph{Int. J. Robot. Res.}, vol.~9,
  no.~2, pp. 62--82, 1990.

\bibitem{Collins2005Science}
S.~Collins, A.~Ruina, R.~Tedrake, and M.~Wisse, ``Efficient bipedal robots
  based on passive-dynamic walkers,'' \emph{Science}, vol. 302, pp. 1082--1085,
  2005.

\bibitem{Hobbelen2007}
D.~Hobbelen and M.~Wisse, ``Limit cycle walking,'' in \emph{Humanoid Robots:
  Human-like Machines}, M.~Hackel, Ed.\hskip 1em plus 0.5em minus 0.4em\relax
  Itech, 2007, pp. 277--294.

\bibitem{Ramezani2017Science}
A.~Ramezani, S.-J. Chung, and S.~Hutchinson, ``A biomimetic robotic platform to
  study flight specializations of bats,'' \emph{Science Robotics}, vol.~2,
  no.~3, 2017.

\bibitem{Georgiades2009OE}
C.~Georgiades, M.~Nahon, and M.~Buehler, ``Simulation of an underwater hexapod
  robot,'' \emph{Ocean Engineering}, vol.~36, pp. 39--47, 2009.

\bibitem{ijspeert2013dynamical}
A.~J. Ijspeert, J.~Nakanishi, H.~Hoffmann, P.~Pastor, and S.~Schaal,
  ``Dynamical movement primitives: Learning attractor models for motor
  behaviors,'' \emph{Neural computation}, vol.~25, no.~2, pp. 328--373, 2013.

\bibitem{motahar2016composing}
M.~S. Motahar, S.~Veer, and I.~Poulakakis, ``Composing limit cycles for motion
  planning of {3D} bipedal walkers,'' in \emph{Proc. IEEE Conf. Decis.
  Control}, Las Vegas, December 2016, pp. 6368--6374.

\bibitem{Veer2017IROS}
S.~Veer, M.~S. Motahar, and I.~Poulakakis, ``Almost driftless navigation of
  {3D} limit-cycle walking bipeds,'' in \emph{Proc. IEEE/RSJ Int. Conf. Intell.
  Robots Syst.}, 2017, pp. 5025--5030.

\bibitem{la1961stability}
J.~LaSalle and S.~Lefschetz, \emph{Stability by Liapunov's Direct Method with
  Applications}.\hskip 1em plus 0.5em minus 0.4em\relax Academic Press, New
  York, 1961.

\bibitem{Veer2020TAC}
S.~Veer and I.~Poulakakis, ``Switched systems with multiple equilibria under
  disturbances: Boundedness and practical stability,'' \emph{IEEE Trans. Autom.
  Control}, vol.~65, no.~6, pp. 2371--2386, 2020.

\bibitem{westervelt2007feedback}
E.~R. Westervelt, J.~W. Grizzle, C.~Chevallereau, J.~H. Choi, and B.~Morris,
  \emph{{F}eedback {C}ontrol of {D}ynamic {B}ipedal {R}obot
  {L}ocomotion}.\hskip 1em plus 0.5em minus 0.4em\relax Boca Raton, FL: CRC
  Press, 2007.

\bibitem{Sharbafi2017book}
M.~A. Sharbafi and A.~Seyfarth, Eds., \emph{Bioinspired Legged
  Locomotion}.\hskip 1em plus 0.5em minus 0.4em\relax Butterworth-Heinemann,
  2017.

\bibitem{Poulakakis2009ICRA}
I.~Poulakakis and J.~W. Grizzle, ``Modeling and control of the monopedal robot
  {Thumper},'' in \emph{Proc. IEEE Int. Conf. Robot. Autom.}, 2009, pp.
  3327--3334.

\bibitem{ames2014rapidly}
A.~Ames, K.~Galloway, J.~Grizzle, and K.~Sreenath, ``Rapidly exponentially
  stabilizing control lyapunov functions and hybrid zero dynamics,'' \emph{IEEE
  Trans. Autom. Control}, vol.~59, no.~4, pp. 876--891, 2014.

\bibitem{hamed2014event}
K.~A. Hamed and J.~W. Grizzle, ``Event-based stabilization of periodic orbits
  for underactuated 3-{D} bipedal robots with left-right symmetry,'' \emph{IEEE
  Trans. Robot.}, vol.~30, no.~2, pp. 365--381, 2014.

\bibitem{Sadeghian2017ICRA}
H.~Sadeghian, C.~Ott, G.~Garofalo, and G.~Cheng, ``Passivity-based control of
  underactuated biped robots within hybrid zero dynamics approach,'' in
  \emph{Proc. IEEE Int. Conf. Robot. Autom.}, 2017, pp. 4096--4101.

\bibitem{Hereid2020ICRA}
G.~A. Castillo, B.~Weng, W.~Zhang, and A.~Hereid, ``Hybrid zero dynamics
  inspired feedback control policy design for {3D} bipedal locomotion using
  reinforcement learning,'' in \emph{Proc. IEEE Int. Conf. Robot. Autom.},
  2020, pp. 8746--8752.

\bibitem{AmesPoulakakis2017BLL}
A.~D. Ames and I.~Poulakakis, ``Hybrid zero dynamics control of legged
  robots,'' in \emph{Bioinspired Legged Locomotion: Models, Concepts, Control
  and Applications}, M.~A. Sharbafi and A.~Seyfarth, Eds.\hskip 1em plus 0.5em
  minus 0.4em\relax Butterworth-Heinemann, 2017, pp. 292--331.

\bibitem{manchester2014ICRA}
I.~R. Manchester and J.~Umenberger, ``Real-time planning with primitives for
  dynamic walking over uneven terrain,'' in \emph{Proc. IEEE Int. Conf. Robot.
  Autom.}, 2014, pp. 4639--4646.

\bibitem{nguyen2015safety}
Q.~Nguyen and K.~Sreenath, ``Safety-critical control for dynamical bipedal
  walking with precise footstep placement,'' \emph{IFAC-PapersOnLine}, vol.~48,
  no.~27, pp. 147--154, 2015.

\bibitem{Nguyen2020Algo}
Q.~Nguyen, X.~Da, J.~W. Grizzle, and K.~Sreenath, ``Dynamic walking on stepping
  stones with gait library and control barrier functions,'' in
  \emph{Algorithmic Foundations of Robotics XII}, K.~Goldberg, P.~Abbeel,
  K.~Bekris, and L.~Miller, Eds.\hskip 1em plus 0.5em minus 0.4em\relax
  Springer, 2020, pp. 384--399.

\bibitem{Gregg2010ICRA}
R.~D. Gregg, T.~Bretl, and M.~Spong, ``Asymptotically stable gait primitives
  for planning dynamic bipedal locomotion in three dimensions,'' in \emph{Proc.
  IEEE Int. Conf. Robot. Autom.}, 2010, pp. 1695--1702.

\bibitem{gregg-planning2012}
R.~D. Gregg, A.~K. Tilton, S.~Candido, T.~Bretl, and M.~W. Spong, ``Control and
  planning of 3-{D} dynamic walking with asymptotically stable gait
  primitives,'' \emph{IEEE Trans. Robot.}, vol.~28, no.~6, pp. 1415--1423,
  2012.

\bibitem{Hubicki2015ICRA}
C.~Hubicki, M.~Jones, M.~Daley, and J.~Hurst, ``Do limit cycles matter in the
  long run? stable orbits and sliding-mass dynamics emerge in task-optimal
  locomotion,'' in \emph{Proc. IEEE Int. Conf. Robot. Autom.}, 2015, pp.
  5113--5120.

\bibitem{hespanha1999stability}
J.~P. Hespanha and A.~S. Morse, ``Stability of switched systems with average
  dwell-time,'' in \emph{Proc. IEEE Conf. Decis. Control}, 1999, pp.
  2655--2660.

\bibitem{liberzon2003switching}
D.~Liberzon, \emph{Switching in {S}ystems and {C}ontrol}.\hskip 1em plus 0.5em
  minus 0.4em\relax Boston, MA: Birkh\"{a}user, 2003.

\bibitem{basar2010}
T.~Alpcan and T.~Basar, ``A stability result for switched systems with multiple
  equilibria,'' \emph{J. Dyn. Continuous, Discrete Impulsive Syst. Ser A, Math.
  Anal.}, vol.~17, pp. 949--958, 2010.

\bibitem{da2016first}
X.~Da, R.~Hartley, and J.~W. Grizzle, ``First steps toward supervised learning
  for underactuated bipedal robot locomotion, with outdoor experiments on the
  wave field,'' in \emph{Proc. of IEEE Int. Conf. on Robotics and Automation},
  2017, pp. 3476--3483.

\bibitem{Gong2019ACC}
Y.~Gong, R.~Hartley, X.~Da, A.~Hereid, O.~Harib, J.-K. Huang, and J.~Grizzle,
  ``Feedback control of a cassie bipedal robot: Walking, standing, and riding a
  segway,'' in \emph{Proc. Amer. Control Conf.}, 2019, pp. 4559--4566.

\bibitem{Nguyen2018IJRR}
Q.~Nguyen, A.~Agrawal, W.~Martin, H.~Geyer, and K.~Sreenath, ``Dynamic bipedal
  locomotion over stochastic discrete terrain,'' \emph{Int. J. Robot. Res.},
  vol.~37, no. 13-14, pp. 1537 -- 1553, 2018.

\bibitem{Sreenath2021Arxiv}
Z.~Li, J.~Zeng, S.~Chen, and K.~Sreenath, ``Vision-aided autonomous navigation
  of underactuated bipedal robots in height-constrained environments,''
  \emph{arXiv:2109.05714}, 2021.

\bibitem{Saglam2014RSS}
C.~O. Saglam and K.~Byl, ``Robust policies via meshing for metastable rough
  terrain walking,'' in \emph{Proc. Robot. Sci. Syst.}, 2014.

\bibitem{Park2013TRO}
H.-W. Park, A.~Ramezani, and J.~W. Grizzle, ``A finite-state machine for
  accommodating unexpected large ground-height variations in bipedal robot
  walking,'' \emph{IEEE Trans. Robot.}, vol.~29, no.~2, pp. 331--345, 2013.

\bibitem{Buss2016ICRA}
S.~Apostolopoulos, M.~Leibold, and M.~Buss, ``Online motion planning over
  uneven terrain with walking primitives and regression,'' in \emph{Proc. IEEE
  Int. Conf. Robot. Autom.}, 2016, pp. 3799--3805.

\bibitem{veer2017supervisory}
S.~Veer, M.~S. Motahar, and I.~Poulakakis, ``Adaptation of limit-cycle walkers
  for collaborative tasks: {A} supervisory switching control approach,'' in
  \emph{Proc. IEEE/RSJ Int. Conf. Intell. Robots Syst.}, 2017, pp. 5840--5845.

\bibitem{Veer2019ICRA}
S.~Veer and I.~Poulakakis, ``Safe adaptive switching among dynamical movement
  primitives: Application to {3D} limit-cycle walkers,'' in \emph{Proc. IEEE
  Int. Conf. Robot. Autom.}, 2019, pp. 3719--3725.

\bibitem{Chand2022RAL}
P.~Chand, S.~Veer, and I.~Poulakakis, ``Interactive dynamic walking: Learning
  gait switching policies with generalization guarantees,'' \emph{IEEE Robot.
  Automat. Lett.}, vol.~7, no.~2, pp. 4149--4156, 2022.

\bibitem{Chand2020ICRA}
------, ``An adaptive supervisory control approach to dynamic locomotion under
  parametric uncertainty,'' in \emph{Proc. IEEE Int. Conf. Robot. Autom.},
  2020, pp. 2443--2449.

\bibitem{Bhounsule2018ACC}
P.~A. Bhounsule, A.~Zamani, and J.~Pusey, ``Switching between limit cycles in a
  model of running using exponentially stabilizing discrete control {L}yapunov
  function,'' in \emph{Proc. Amer. Control Conf.}, 2018, pp. 3714--3719.

\bibitem{Bhounsule2019ICRA}
A.~Zamani, J.~D. Galloway, and P.~A. Bhounsule, ``Feedback motion planning of
  legged robots by composing orbital lyapunov functions using rapidly-exploring
  random trees,'' in \emph{Proc. IEEE Int. Conf. Robot. Autom.}, 2019, pp.
  1410--1416.

\bibitem{QuCao2015}
Q.~Cao, A.~T. {van Rijn}, and I.~Poulakakis, ``On the control of gait
  transitions in quadrupedal running,'' in \emph{Proc. IEEE/RSJ Int. Conf.
  Intell. Robots Syst.}, Sep. 2015, pp. 5136 -- 5141.

\bibitem{Cao2016ALR}
Q.~Cao and I.~Poulakakis, ``Quadrupedal running with a flexible torso:
  {C}ontrol and speed transitions with sums-of-squares verification,''
  \emph{Artificial Life and Robotics}, vol.~21, no.~4, pp. 384--392, 2016.

\bibitem{Ames2021IROS}
W.~Ubellacker, N.~Csomay-Shanklin, T.~G. Molnar, and A.~D. Ames, ``Verifying
  safe transitions between dynamic motion primitives on legged robots,'' in
  \emph{Proc. IEEE/RSJ Int. Conf. Intell. Robots Syst.}, 2021, pp. 8477--8484.

\bibitem{Veer2019ACC}
S.~Veer and I.~Poulakakis, ``Practical stability of switched systems with
  multiple equilibria under disturbances,'' in \emph{Proc. Amer. Control
  Conf.}, 2019.

\bibitem{Dorothy2016SCL}
M.~Dorothy and S.-J. Chung, ``Switched systems with multiple invariant sets,''
  \emph{Syst. Control Lett.}, vol.~96, pp. 103--109, 2016.

\bibitem{haddad2008nonlinear}
W.~M. Haddad and V.~Chellaboina, \emph{Nonlinear {D}ynamical {S}ystems and
  {C}ontrol: {A} {L}yapunov-{B}ased {A}pproach}.\hskip 1em plus 0.5em minus
  0.4em\relax Princeton, NJ: Princeton University Press, 2008.

\bibitem{burridge1999sequential}
R.~R. Burridge, A.~A. Rizzi, and D.~E. Koditschek, ``Sequential composition of
  dynamically dexterous robot behaviors,'' \emph{Int. J. Robot. Res.}, vol.~18,
  no.~6, pp. 534--555, 1999.

\bibitem{Mason1985ICRA}
M.~Mason, ``The mechanics of manipulation,'' in \emph{Proc. IEEE Int. Conf.
  Robot. Autom.}, 1985, pp. 544--548.

\bibitem{Lozano1984IJRR}
T.~Lozano-P{\`e}rez, M.~T. Mason, and R.~H. Taylor, ``Automatic synthesis of
  fine-motion strategies for robots,'' \emph{Int. J. Robot. Res.}, vol.~3,
  no.~1, pp. 3--23, 1984.

\bibitem{parrilo2000structured}
P.~A. Parrilo, ``Structured semidefinite programs and semialgebraic geometry
  methods in robustness and optimization,'' Ph.D. dissertation, California
  Institute of Technology, 2000.

\bibitem{lavalle-RRT}
S.~M. LaValle and J.~J. Kuffner, ``Randomized kinodynamic planning,''
  \emph{Int. J. Robot. Res.}, vol.~20, no.~5, pp. 378--400, 2001.

\bibitem{tedrake2010}
R.~Tedrake, I.~R. Manchester, M.~Tobenkin, and J.~W. Roberts, ``{LQR}-trees:
  Feedback motion planning via sums-of-squares verification,'' \emph{Int. J.
  Robot. Res.}, vol.~29, no.~8, pp. 1038--1052, 2010.

\bibitem{majumdar2016funnel}
A.~Majumdar and R.~Tedrake, ``Funnel libraries for real-time robust feedback
  motion planning,'' \emph{Int. J. Robot. Res.}, vol.~36, no.~8, pp. 947--982,
  2017.

\bibitem{DiCairano2016CDC}
C.~Danielson, A.~Weiss, K.~Berntorp, and S.~Di~Cairano, ``Path planning using
  positive invariant sets,'' in \emph{Proc. IEEE Conf. Decis. Control}, 2016.

\bibitem{DiCairano2017CCTA}
A.~Weiss, C.~Danielson, K.~Berntorp, I.~Kolmanovsky, and S.~Di~Cairano,
  ``Motion planning with invariant set trees,'' in \emph{Proc. Conf. Control
  Technol. Appl.}, 2017, pp. 1625--1630.

\bibitem{Barbosa2020CDC}
F.~S. Barbosa, L.~Lindemann, D.~V. Dimarogonas, and J.~Tumova, ``Provably safe
  control of {L}agrangian systems in obstacle-scattered environments,'' in
  \emph{Proc. IEEE Conf. Decis. Control}, 2020, pp. 2056--2061.

\bibitem{Shen2021ICRA}
S.~Li, D.~Park, Y.~Sung, J.~A. Shah, and N.~Roy, ``Reactive task and motion
  planning under temporal logic specifications,'' in \emph{Proc. IEEE Int.
  Conf. Robot. Autom.}, 2021, pp. 12\,618--12\,624.

\bibitem{Vasilopoulos2022ICRA}
V.~Vasilopoulos, S.~Castro, W.~Vega-Brown, D.~E. Koditschck, and N.~Roy, ``A
  hierarchical deliberative-reactive system architecture for task and motion
  planning in partially known environments,'' in \emph{Proc. IEEE Int. Conf.
  Robot. Autom.}, 2022, pp. 7342--7348.

\bibitem{Zhao2021arxiv}
A.~Shamsah, Z.~Gu, J.~Warnke, S.~Hutchinson, and Y.~Zhao, ``Integrated task and
  motion planning for safe legged navigation in partially observable
  environments,'' \emph{arXiv:2110.12097}, 2021.

\bibitem{Wieber2016ModelingAC}
P.-B. Wieber, R.~Tedrake, and S.~Kuindersma, ``Modeling and control of legged
  robots,'' in \emph{Springer Handbook of Robotics}, B.~Sicialiano and
  O.~Khatib, Eds.\hskip 1em plus 0.5em minus 0.4em\relax Springer, 2016, pp.
  1203--1234.

\bibitem{Wensing2017BLL}
P.~M. Wensing and S.~Revzen, ``Template models for control,'' in
  \emph{Bioinspired Legged Locomotion: Models, Concepts, Control and
  Applications}, M.~A. Sharbafi and A.~Seyfarth, Eds.\hskip 1em plus 0.5em
  minus 0.4em\relax Butterworth-Heinemann, 2017, pp. 240--263.

\bibitem{Kajita2001IROS}
S.~Kajita, F.~Kanehiro, K.~Kaneko, K.~Yokoi, and H.~Hirukawa, ``The {3D} linear
  inverted pendulum mode: A simple modeling for a biped walking pattern
  generation,'' in \emph{Proc. IEEE/RSJ Int. Conf. Intell. Robots Syst.}, 2001,
  pp. 239--246.

\bibitem{poulakakis2009spring}
I.~Poulakakis and J.~W. Grizzle, ``The spring loaded inverted pendulum as the
  hybrid zero dynamics of an asymmetric hopper,'' \emph{IEEE Trans. Autom.
  Control}, vol.~54, no.~8, pp. 1779--1793, 2009.

\bibitem{Poulakakis2010ICRA}
I.~Poulakakis, ``Spring loaded inverted pendulum embedding: {E}xtensions toward
  the control of compliant running robots,'' in \emph{Proc. IEEE Int. Conf.
  Robot. Autom.}, 2010, pp. 5219--5224.

\bibitem{Wensing2021RAL}
V.~Kurtz, P.~M. Wensing, and H.~Lin, ``Approximate simulation for
  template-based whole-body control,'' \emph{IEEE Robot. Automat. Lett.},
  vol.~6, no.~2, pp. 558--565, 2021.

\bibitem{veer2019poincare}
S.~Veer, Rakesh, and I.~Poulakakis, ``Input-to-state stability of periodic
  orbits of systems with impulse effects via {P}oincar\'e analysis,''
  \emph{IEEE Trans. Autom. Control}, vol.~64, no.~11, pp. 4583--4598, 2019.

\bibitem{Veer2019CDC}
S.~Veer and I.~Poulakakis, ``Robustness of periodic orbits of impulsive systems
  {\`a} la {P}oincar{\'e},'' in \emph{Proc. IEEE Int. Conf. on Dec. Control},
  2019, pp. 3966--3971.

\bibitem{Isidori}
A.~Isidori, \emph{Nonlinear {C}ontrol {S}ystems}, 3rd~ed.\hskip 1em plus 0.5em
  minus 0.4em\relax New York: Springer, 1995.

\bibitem{spong2005controlled}
M.~W. Spong and F.~Bullo, ``Controlled symmetries and passive walking,''
  \emph{IEEE Trans. Autom. Control}, vol.~50, no.~7, pp. 1025--1031, 2005.

\bibitem{3D-robotica2012}
C.-L. Shih, J.~Grizzle, and C.~Chevallereau, ``From stable walking to steering
  of a {3D} bipedal robot with passive point feet,'' \emph{Robotica}, vol.~30,
  no.~07, pp. 1119--1130, 2012.

\bibitem{Zhai2003CDC}
G.~Zhai and A.~N. Michel, ``Generalized practical stability analysis of
  discontinuous dynamical systems,'' in \emph{Proc. IEEE Conf. Decis. Control},
  2003, pp. 1663--1668.

\bibitem{majumdar2013robust}
A.~Majumdar and R.~Tedrake, ``Robust online motion planning with regions of
  finite time invariance,'' in \emph{Algorithmic Foundations of Robotics X},
  E.~Frazzoli, T.~Lozano-Perez, N.~Roy, and D.~Rus, Eds.\hskip 1em plus 0.5em
  minus 0.4em\relax Springer, 2013, pp. 543--558.

\bibitem{Chevallereau2010IROS}
C.~Chevallereau, J.~Grizzle, and C.-L. Shih, ``Steering of a {3D} bipedal robot
  with an underactuated ankle,'' in \emph{Proc. IEEE/RSJ Int. Conf. Intell.
  Robots Syst.}, 2010.

\bibitem{gregg2009reduction}
R.~D. Gregg and M.~W. Spong, ``Reduction-based control of three-dimensional
  bipedal walking robots,'' \emph{Int. J. Robot. Res.}, vol.~29, no.~6, pp.
  680--702, 2009.

\bibitem{sinnet20093d}
R.~W. Sinnet and A.~D. Ames, ``3{D} bipedal walking with knees and feet: {A}
  hybrid geometric approach,'' in \emph{Proc. IEEE Conf. Decis. Control},
  Shanghai, P.R. China, December 2009, pp. 3208--3213.

\bibitem{Spong2007RAM}
M.~W. Spong, J.~K. Holm, and D.~Lee, ``Passivity- based control of bipedal
  locomotion,'' \emph{IEEE Robot. Automat. Mag.}, vol.~12, no.~2, pp. 30--40,
  2007.

\bibitem{Nguyen2015RSS}
Q.~Nguyen and K.~Sreenath, ``Optimal robust control for bipedal robots through
  control lyapunov function based quadratic programs,'' in \emph{Proc. Robot.
  Sci. Syst.}, 2015.

\bibitem{grizzle2014models}
J.~W. Grizzle, C.~Chevallereau, R.~W. Sinnet, and A.~D. Ames, ``Models,
  feedback control, and open problems of 3d bipedal robotic walking,''
  \emph{Automatica}, vol.~50, no.~8, pp. 1955--1988, 2014.

\bibitem{3D-TRO2009}
C.~Chevallereau, J.~W. Grizzle, and C.-L. Shih, ``Asymptotically stable walking
  of a five-link underactuated {3-D} bipedal robot,'' \emph{IEEE Trans.
  Robot.}, vol.~25, no.~1, pp. 37--50, 2009.

\bibitem{videolink}
\BIBentryALTinterwordspacing
Simulation results. [Online]. Available: \url{https://youtu.be/FYRBuWOqKww}
\BIBentrySTDinterwordspacing

\bibitem{Nark2022RAL}
K.~S. Narkhede, A.~M. Kulkarni, D.~A. Thanki, and I.~Poulakakis, ``A sequential
  {MPC} approach to reactive planning for bipedal robots using safe corridors
  in highly cluttered environments,'' \emph{IEEE Robot. Automat. Lett.},
  vol.~7, no.~4, pp. 11\,831--11\,838, 2022.

\bibitem{Mora_2019_RAL}
B.~{Brito}, B.~{Floor}, L.~{Ferranti}, and J.~{Alonso-Mora}, ``Model predictive
  contouring control for collision avoidance in unstructured dynamic
  environments,'' \emph{IEEE Robot. Automat. Lett.}, vol.~4, no.~4, pp.
  4459--4466, 2019.

\bibitem{Veer2018PhDThesis}
S.~Veer, ``Composing motion primitives under disturbances: A switched systems
  approach,'' Ph.D. dissertation, University of Delaware, 2018.

\bibitem{Boyd2004book}
S.~Boyd and L.~Vandenberghe, \emph{Convex Optimization}.\hskip 1em plus 0.5em
  minus 0.4em\relax Cambridge University Press, 2004.

\end{thebibliography}

\end{document}